\documentclass[dvipsnames]{article}
\usepackage[eandd,nonatbib,preprint]{neurips_2026}

\usepackage{fancyhdr}
\usepackage[numbers]{natbib}
\setcitestyle{numbers}

\usepackage{iftex}
\ifPDFTeX
  \usepackage[utf8]{inputenc}
  \usepackage[T1]{fontenc}
\else
  \usepackage{fontspec}
  \usepackage{xeCJK}
\fi
\usepackage{url}
\usepackage{booktabs}
\usepackage{amsfonts}
\usepackage{nicefrac}
\usepackage{microtype}
\usepackage{amsmath}
\usepackage{amssymb}
\usepackage{mathtools}
\usepackage{amsthm}
\usepackage{multirow}
\usepackage{hyperref}
\usepackage[table,dvipsnames]{xcolor}
\usepackage{subcaption}
\usepackage{graphicx}
\usepackage{siunitx}
\usepackage{caption}

\hypersetup{
  colorlinks = true,
  urlcolor = RoyalBlue,
  linkcolor = RoyalBlue,
  citecolor = RoyalBlue
}

\definecolor{light-light-gray}{gray}{0.92}

\usepackage{rotating}
\usepackage{placeins}
\usepackage{enumitem}
\usepackage{wrapfig}
\usepackage{float}

\usepackage{tikz}
\usepackage{pgfplots}
\pgfplotsset{compat=1.18}
\usetikzlibrary{arrows.meta,positioning,calc,shapes.geometric,
                matrix,backgrounds,decorations.pathreplacing}

\usepackage{colortbl}
\usepackage{cellspace}
\newcolumntype{w}{>{\columncolor{white}}c}
\usepackage[capitalize,noabbrev]{cleveref}

\usepackage{xpatch}
\makeatletter
\renewcommand{\subsection}{%
  \@startsection{subsection}{2}{\z@}%
                {-1.4ex \@plus -0.4ex \@minus -0.2ex}%
                { 0.45ex \@plus  0.15ex}%
                {\normalsize\bfseries\raggedright}%
}
\renewcommand\paragraph{\@startsection{paragraph}{4}{\z@}
{1.35ex \@plus1ex \@minus.2ex}
{-.5em}
{\normalfont\normalsize\bfseries}}
\makeatother

\usepackage{xspace}
\newcommand{\cosfly}{\textsc{CosFly}\xspace}

\usepackage{pifont}
\newcommand{\cmark}{\ding{51}}%
\newcommand{\xmark}{\ding{55}}%

\title{CosFly: Plan in the Matrix, Fly in the World}

\author{
  Hanxuan Chen$^{1}$ \quad
  Xiangyue Wang$^{1}$\thanks{Co-second authors.} \quad
  Songsheng Cheng$^{1}$\footnotemark[1] \quad
  Ruilong Ren$^{1}$\footnotemark[1] \quad
  Jie Zheng$^{2}$\footnotemark[1] \\
  \bf Shuai Yuan$^{3}$ \quad
  Tianle Zeng$^{4}$ \quad
  Hanzhong Guo$^{5}$ \quad
  Binbo Li$^{3}$ \quad
  Kangli Wang$^{1}$\thanks{Corresponding authors.} \quad
  Ji Pei$^{1}$\footnotemark[2] \\
  $^{1}$Autel Robotics \quad
  $^{2}$Nanjing University \quad
  $^{3}$Peking University \\
  $^{4}$Southern University of Science and Technology \quad
  $^{5}$University of Hong Kong \\
  \texttt{peiji@autelrobotics.com}
}

\begin{document}

\maketitle

\begin{center}
\centering
\includegraphics[width=0.95\textwidth]{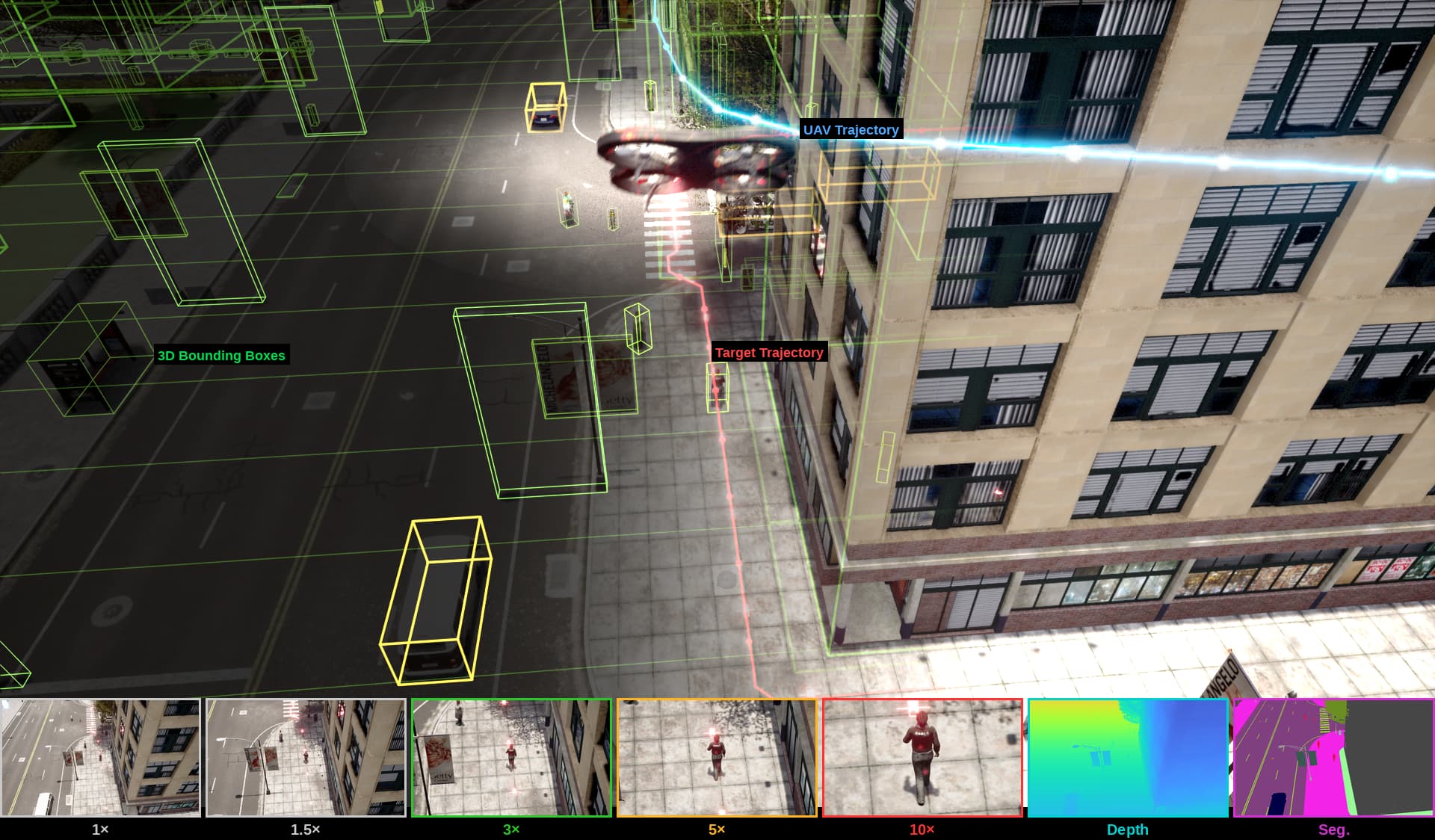}
\captionof{figure}{Derived from the cultural metaphor of \textit{The Matrix}, the core view---``We do not transform reality, but transform the Matrix''---summarizes our paradigm: we build editable, controllable virtual worlds to bypass the limitations of physical reality. The left half illustrates trajectory planning within a structured 3D ``Matrix''; the right half shows the photorealistic world for UAV flight execution; the bottom row presents multi-modal outputs at multiple zoom levels.}
\label{fig:teaser}
\end{center}

\begin{abstract}
We present CosFly, a box-structured planning and multimodal simulation pipeline for aerial tracking, together with CosFly-Track, a large-scale UAV dataset for dynamic target tracking across diverse environments including urban centers, highways, rural landscapes, forests, and coastal towns. In our current implementation on CARLA, \cosfly provides a modular 7-step construction pipeline that converts complex 3D worlds into structured obstacle representations for planning, then projects the resulting trajectories back into multi-modal sensor data---including RGB images, high-precision depth maps, and semantic segmentation masks---paired with natural language navigation instructions. A key feature is the support for configurable fixed-FOV zoom levels (one FOV setting drawn per trajectory and held constant throughout), enabling simulation of various focal lengths through camera-intrinsic adjustments. The pipeline covers the complete workflow from 3D map export through grid simplification, pedestrian and drone trajectory planning, multi-modal rendering with 6-DOF pose annotations, quality inspection, and teacher-student caption generation. We analyze two trajectory-planning paradigms for aerial target tracking: a conventional two-stage pipeline with frontend candidate generation and backend refinement, and a direct gradient-based formulation that optimizes multiple tracking constraints in a single objective. The public CosFly-Track release contains 250 validated trajectories and approximately 100,000 rendered images with complete 6-DOF drone pose annotations (position $x, y, z$ and orientation yaw, pitch, roll). Together, the pipeline and dataset establish a scalable foundation for aerial-ground collaborative research, supporting dynamic target tracking, UAV navigation, and multi-modal perception across diverse environments.
\end{abstract}

\section{Introduction}
\label{sec:introduction}

Vision-Language Models (VLMs) have emerged as a powerful paradigm for robotic navigation and autonomous decision-making, combining visual perception with natural language understanding to enable instruction-following agents~\citep{radford2021clip,liu2023llava,openai2023gpt4}. Recent advances in multi-modal learning, including contrastive vision-language alignment~\citep{radford2021clip}, instruction-tuned multimodal assistants~\citep{liu2023llava,li2023blip2}, and large-scale vision-language models~\citep{openai2023gpt4}, have demonstrated remarkable zero-shot and few-shot capabilities across a wide range of visual understanding tasks. These breakthroughs suggest a promising direction for aerial target tracking and drone navigation, where an autonomous unmanned aerial vehicle (UAV) must continuously observe and follow a ground-level target while understanding high-level navigation instructions.

However, training and benchmarking VLM-based drone tracking models requires large-scale, diverse, and richly annotated multi-modal datasets that pair natural language instructions with multi-sensor observations. Existing aerial tracking datasets such as UAV123~\citep{mueller2016uav123} and VisDrone~\citep{zhu2021visdrone} have driven progress in aerial object detection and tracking, but they suffer from critical limitations. First, real-world drone data collection is expensive, time-consuming, and poses safety risks~\citep{zhu2021visdrone}, resulting in datasets that typically contain only hundreds of short video sequences with limited scene diversity. Second, these datasets provide only RGB video with simple bounding box annotations, lacking the depth maps, instance segmentation masks, and natural language navigation instructions that are essential for training multi-modal VLMs~\citep{gu2022vlnsurvey}. Third, the annotation cost for dense natural language instructions is prohibitively expensive, preventing manual labeling at the scale required for modern VLM training.

Simulation-based data generation offers a scalable alternative to real-world collection. Platforms such as CARLA~\citep{dosovitskiy2017carla} provide diverse, high-fidelity driving environments with multi-modal sensor simulation, while indoor simulators like AI2-THOR~\citep{kolve2017ai2thor} and Habitat~\citep{savva2019habitat} have enabled rapid progress in embodied AI research through synthetic data generation. Recently, several datasets have emerged to explore the aerial perspective, such as OpenFly~\citep{gao2025openfly} and AerialVLN~\citep{liu2023aerialvln}, which provide large-scale environments for UAV navigation. However, these existing aerial datasets primarily focus on simplified point-to-point vision-language navigation tasks in unconstrained 3D spaces without ground-level physical restrictions. Consequently, complex and highly dynamic aerial tasks, particularly drone tracking in realistic outdoor environments, remain largely unexplored.

To address these gaps, we introduce \cosfly, a simulation-based pipeline and dataset for multi-modal aerial tracking. Built on CARLA, \cosfly automatically generates diverse aerial tracking trajectories and multi-modal sensor data (RGB images, depth maps, and semantic segmentation) paired with natural language navigation instructions. Our work makes four key contributions:

\begin{enumerate}
\item \textbf{The CosFly-Track Dataset:} A large-scale multi-modal aerial tracking dataset with RGB, depth, semantic segmentation, and natural language navigation instructions, generated from realistic simulation. The first public release contains 250 validated trajectories, yielding approximately 100,000 rendered images with roughly 400 images per trajectory captured at 2\,Hz.

\item \textbf{A Generalizable 7-Step Pipeline:} A modular, reproducible construction pipeline that can be readily adapted to diverse scenarios and simulation backends. The pipeline covers the complete workflow from 3D map export through trajectory planning, multi-modal rendering, quality inspection, and caption generation, enabling researchers to construct custom aerial tracking datasets for new maps and scenarios.

\item \textbf{Trajectory Planning Paradigm Analysis:} A comparison between a conventional two-stage planner (TA*+Smooth: visibility-aware Track A* frontend plus post-smoothing backend) and a direct one-shot multi-constraint gradient optimizer (MuCO). The first public release is generated with MuCO-optimized trajectories; TA*+Smooth is evaluated as a comparison baseline and may support future releases.

\item \textbf{Baseline Experiments:} We release detailed experimental data for each stage of the rendering and data construction pipeline, including real measurements of planning speed, rendering efficiency, and stage-wise processing cost, providing references for downstream deployment planning and system design.
\end{enumerate}

Figure~\ref{fig:pipeline_overview} provides an overview of the seven-step \cosfly pipeline. The remainder of this paper is organized as follows: Section~\ref{sec:related} reviews related work; Section~\ref{sec:pipeline} describes the pipeline in detail; Section~\ref{sec:dataset} presents dataset statistics; Section~\ref{sec:experiments} reports baseline experiments; and Section~\ref{sec:conclusion} discusses limitations and concludes.

\section{Related Work}
\label{sec:related}

We organize related work into four areas: aerial tracking datasets, synthetic data for visual navigation, vision-language navigation, and trajectory planning for UAVs.

\subsection{Aerial Tracking Datasets}
\label{sec:rw_tracking}

Aerial tracking has been a longstanding research topic in computer vision, with several benchmark datasets establishing standardized evaluation protocols. Early benchmarks such as UAV123~\citep{mueller2016uav123} provide 123 full-HD video sequences captured by drones at various altitudes and viewing angles, annotated with bounding boxes for single-object tracking. VisDrone~\citep{zhu2021visdrone} extends this paradigm to more complex scenarios with dense object annotations for detection, tracking, and counting in drone-captured imagery. DroneCrowd~\citep{wen2021dronecrowd} further pushes the scale by providing crowd-level annotations for detection, tracking, and counting in densely populated aerial scenes. UAVDT~\citep{du2018uavdt} contributes approximately 80,000 frames with rich attribute annotations for object detection and tracking tasks.

Recent years have seen continued progress in UAV dataset development, particularly with the introduction of multi-modal and large-scale benchmarks. WebUAV-3M~\citep{webuav3m2023} introduces a million-scale benchmark for deep UAV tracking, significantly expanding the scale of available training data. M3OT~\citep{m3ot2025} presents the first multi-drone multi-modality dataset combining RGB and infrared thermal imagery for multi-object tracking. MUST~\citep{qin2025must} provides the first large-scale multispectral UAV single object tracking dataset with 250 video sequences. For multi-object tracking in wild environments, BuckTales~\citep{naik2024bucktales} offers a large-scale multi-UAV dataset for tracking and re-identification of wild antelopes. D-PTUAC~\citep{dptuac2023} addresses the challenging scenario of tracking individuals in uniform appearance crowds from drone perspectives. The Anti-UAV Challenge~\citep{antiuav2023} has established benchmarks for detecting and tracking UAVs themselves, addressing security applications. Furthermore, UAVScenes~\citep{wang2025uavscenes} introduces a comprehensive multi-modal dataset with frame-wise semantic annotations for both camera images and LiDAR point clouds, supporting high-level scene understanding tasks. The accurate fusion of such multi-modal sensor data relies on precise extrinsic calibration between LiDAR and camera systems, as addressed by recent calibration methods such as Yoco~\citep{zeng2025yoco}.

Despite these contributions, existing datasets share fundamental limitations. Their scale remains constrained by the cost and risk of real-world drone flights, typically yielding only hundreds to thousands of sequences. More importantly, they provide exclusively RGB video (or RGB-IR pairs) with geometric annotations (bounding boxes, trajectories), lacking depth maps, semantic segmentation masks, and natural language descriptions that are essential for multi-modal VLM training. The expense and danger of collecting large-scale aerial footage with multiple sensor modalities makes it impractical to extend these datasets through real-world data collection alone.

In contrast, \cosfly leverages simulation to generate large-scale multi-modal data at low cost. Each frame includes synchronized RGB, depth, and semantic segmentation outputs alongside natural language navigation instructions---modalities that real-world aerial datasets do not provide.

\subsection{Synthetic Data for Visual Navigation}
\label{sec:rw_synthetic}

Synthetic data generation has become a cornerstone of embodied AI research, with several simulation platforms providing realistic environments for visual navigation tasks. AI2-THOR~\citep{kolve2017ai2thor} offers interactive indoor environments with physics simulation and multi-modal rendering for embodied agents. Habitat~\citep{savva2019habitat} enables efficient navigation in scanned real-world 3D environments, supporting RGB, depth, and semantic sensors. The Gibson Env~\citep{xia2018gibson} provides real-world 3D reconstructions for visual navigation with sim-to-real transfer capabilities.

For outdoor and vehicle-scale simulation, CARLA~\citep{dosovitskiy2017carla} provides a high-fidelity open-source driving simulator supporting multi-modal sensor suites including cameras, LiDAR, and semantic segmentation. AirSim~\citep{shah2018airsim} offers physics-accurate multirotor flight and aerial sensing, but lacks realistic ground traffic and pedestrian interactions. To overcome the synchronization overhead and spatial-temporal inconsistency inherent in bridge-based co-simulation, CARLA-Air~\citep{zeng2026carla} unifies CARLA and AirSim within a single Unreal Engine process, delivering a shared physics tick, a unified rendering pipeline, and full preservation of both native APIs. These platforms have enabled large-scale dataset generation for autonomous driving~\citep{dosovitskiy2017carla} and aerial robotics research.

While recent advancements have yielded large-scale aerial datasets such as VisDrone~\citep{zhu2021detectiontrackingmeetdrones} and Griffin~\citep{wang2025griffinaerialgroundcooperativedetection}, these benchmarks predominantly focus on passive perception tasks. To advance active aerial autonomy, platforms like OpenFly~\citep{gao2026openflycomprehensiveplatformaerial} and AerialVLN~\citep{liu2023aerialvlnvisionandlanguagenavigationuavs} have introduced large-scale benchmarks for drone-centric vision-language navigation, while UAV-Track~\citep{zhang2026uavtrackvlaembodiedaerial} targets embodied visual tracking of dynamic vehicles and pedestrians in urban environments. However, these datasets either focus on navigating toward static landmarks or lack obstacle-aware trajectory planning during dynamic target pursuit. \cosfly addresses this gap by building on CARLA to generate drone-centric multi-modal data specifically tailored for active aerial tracking, combining elevated viewpoints with dynamic pedestrian targets and obstacle-aware trajectory planning.

\subsection{Vision-Language Navigation}
\label{sec:rw_vln}

Vision-Language Navigation (VLN) combines visual perception with natural language understanding for instruction-following tasks~\citep{gu2022vlnsurvey}. The foundational R2R dataset~\citep{anderson2018r2r} introduced the task of following natural language navigation instructions in indoor environments, pairing step-by-step instructions with panoramic visual observations. Subsequent datasets and benchmarks have expanded VLN to outdoor street-level navigation and 3D object grounding.

The emergence of large-scale VLMs has significantly advanced VLN capabilities. CLIP~\citep{radford2021clip} established a foundation for vision-language alignment through contrastive learning on 400 million image-text pairs, enabling zero-shot visual recognition through natural language prompts. LLaVA~\citep{liu2023llava} demonstrated that instruction tuning on image-text data produces powerful multimodal assistants capable of complex visual reasoning. BLIP-2~\citep{li2023blip2} efficiently bridged frozen image encoders with large language models using a lightweight querying transformer, achieving state-of-the-art performance on vision-language tasks with reduced computational cost.

Despite these architectural advances, VLN benchmarks have historically remained concentrated in indoor and street-level settings. Recently, there has been a growing interest in aerial VLN. A recent comprehensive survey~\citep{chen2026vision} systematically reviews the progress and challenges of VLN for UAVs, identifying the lack of large-scale multi-modal aerial datasets as a critical bottleneck. CityNav~\citep{wang2024citynav} introduces a large-scale dataset for real-world aerial navigation over cities, providing human demonstration trajectories paired with natural language descriptions. TravelUAV~\citep{wang2024traveluav} proposes a realistic UAV simulation platform and an assistant-guided UAV object search benchmark for vision-language navigation. Additionally, UAV-MM3D~\citep{zou2025uavmm3d} offers a large-scale synthetic benchmark for 3D perception of UAVs with multi-modal data, including RGB, IR, LiDAR, Radar, and DVS. In the broader context of outdoor robot navigation, EZREAL~\citep{zeng2025ezreal} demonstrates zero-shot navigation toward distant targets under varying visibility conditions, highlighting the importance of robust perception in unstructured outdoor environments. While these recent works have begun to explore aerial navigation and perception, there remains a notable absence of aerial datasets that specifically pair natural language instructions with dense multi-modal sensor data (including depth and semantic annotations) tailored for dynamic aerial tracking scenarios. \cosfly fills this gap by providing paired navigation instructions with RGB, depth, and semantic segmentation data specifically designed for aerial tracking, enabling training and evaluation of VLM-based drone navigation models.

\subsection{Trajectory Planning for UAVs}
\label{sec:rw_planning}

Trajectory planning for UAVs encompasses both classical heuristic methods and modern optimization-based approaches. The A* algorithm~\citep{hart1968astar} remains a foundational graph search method for finding minimum-cost paths, widely used in robotics and autonomous navigation. More recently, Track A* (TA*)~\citep{chen2026trackastar} extends A*-style search to active target tracking by planning visibility-aware trajectories on a discretized four-dimensional spatio-temporal grid, thereby addressing the need for scalable offline reference trajectories that jointly consider target visibility, obstacle clearance, and temporal feasibility. Sampling-based planners such as RRT~\citep{lavalle1998rrt} provide probabilistic completeness for high-dimensional configuration spaces, enabling motion planning in complex environments with dynamic obstacles.

For trajectory optimization, Model Predictive Control (MPC)~\citep{kouvaritakis2016mpc} formulates path planning as a receding-horizon optimization problem, iteratively computing optimal control inputs while respecting dynamic constraints. Nonlinear Programming (NLP) approaches enable direct trajectory optimization with smoothness, collision avoidance, and tracking quality objectives. However, a persistent challenge in UAV trajectory planning is the trade-off between computational efficiency and tracking quality: fast heuristic methods produce feasible but potentially suboptimal paths, while high-quality optimization methods require significant computation.

Path simplification and smoothing techniques play an important role in trajectory post-processing. The Douglas-Peucker algorithm~\citep{douglas1973dp} reduces path complexity by removing redundant waypoints, while Catmull-Rom splines~\citep{catmull1974spline} provide smooth interpolation through control points. These techniques are complementary to planning algorithms and essential for generating executable trajectories.

\cosfly uses aerial target tracking as a benchmark setting for comparing two planning paradigms along this efficiency-quality spectrum: conventional two-stage planning, which separates candidate generation from refinement, and direct multi-constraint gradient planning, which jointly optimizes tracking, smoothness, visibility, and collision-avoidance objectives. This paradigm-level comparison on realistic obstacle densities provides a practical reference for constructing scalable aerial tracking datasets.

\section{The \cosfly Pipeline}
\label{sec:pipeline}

\begin{figure}[t]
\centering
\includegraphics[width=0.88\textwidth]{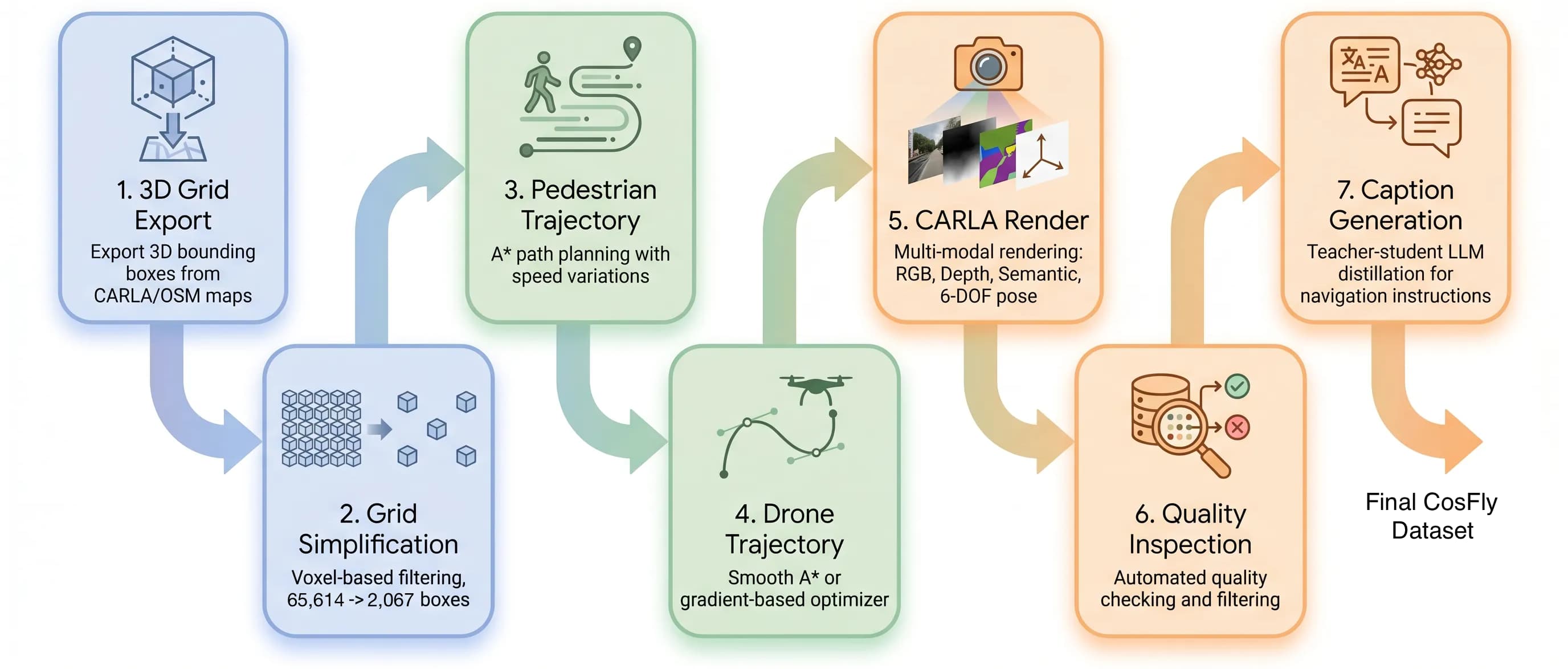}
\captionsetup{font=small}
\caption{Overview of the seven-step \cosfly construction pipeline. Step~4 produces the current CosFly-Track release via the MuCO planner; TA*+Smooth-rendered data may be supported in a future release.}
\label{fig:pipeline_overview}
\end{figure}

\cosfly is a modular seven-step pipeline for constructing multi-modal aerial tracking datasets. Each step has well-defined inputs and outputs and is independently runnable. \Cref{fig:pipeline_overview} shows the architecture. This section is primarily a method description: small in-section validation statistics motivate the simplification (\cref{sec:step1,sec:step2}), the single-scenario planner comparison (\cref{sec:step4}), and the release-level quality envelope (\cref{sec:step6}); all other experimental results are deferred to \cref{sec:dataset,sec:experiments}. We compare two planners under shared interfaces: \emph{TA*+Smooth} (a two-stage pipeline with a visibility-aware Track A*~\citep{chen2026trackastar} frontend on a 4D voxel grid plus a post-smoothing backend) and \emph{MuCO} (a one-shot multi-constraint gradient optimizer). For the current CosFly-Track release we render MuCO-optimized drone trajectories; TA*+Smooth rendering may be supported in a future release, and TA*+Smooth is therefore used in this paper only for the paradigm-level comparison in \cref{sec:step4,sec:exp_traj}. Full algorithmic specifications are given in \cref{app:planner_details}.

\subsection{Step 1: Map Offline to 3D Grid}
\label{sec:step1}

The pipeline begins by exporting 3D bounding boxes from CARLA's semantic annotation system. The export script connects to a CARLA $0.9.16$ server with Town10HD\_Opt loaded, iterates over every \texttt{carla.CityObjectLabel} category, and calls \texttt{get\_environment\_objects(label)} to collect the per-object \texttt{bounding\_box}. We pick Town10HD\_Opt as the primary map because it has the highest geometric density and visual fidelity among CARLA's shipped Town* maps; generalization to other maps is illustrated qualitatively on Town07\_Opt in \cref{sec:appendix_g_mask_editor}.

Each box record stores the \texttt{CityObjectLabel} type and a stable integer id, the centre and half-extents in metres, the Euler rotation in degrees, and pre-computed AABB min/max corners; the full eight-field schema is reproduced in Appendix~\ref{sec:appendix_g_mask_editor} (\cref{tab:boxes_schema}). All numeric fields are in metres / degrees in a right-handed Cartesian frame ($+x$ East, $+y$ North, $+z$ Up) whose origin coincides with the CARLA world origin. The \emph{coordinate correction} re-expresses CARLA's native left-handed Unreal poses into this right-handed frame and writes the resulting min/max once, so all downstream stages share the same coordinates.

\begin{figure}[!tb]
\centering
\includegraphics[width=0.62\columnwidth]{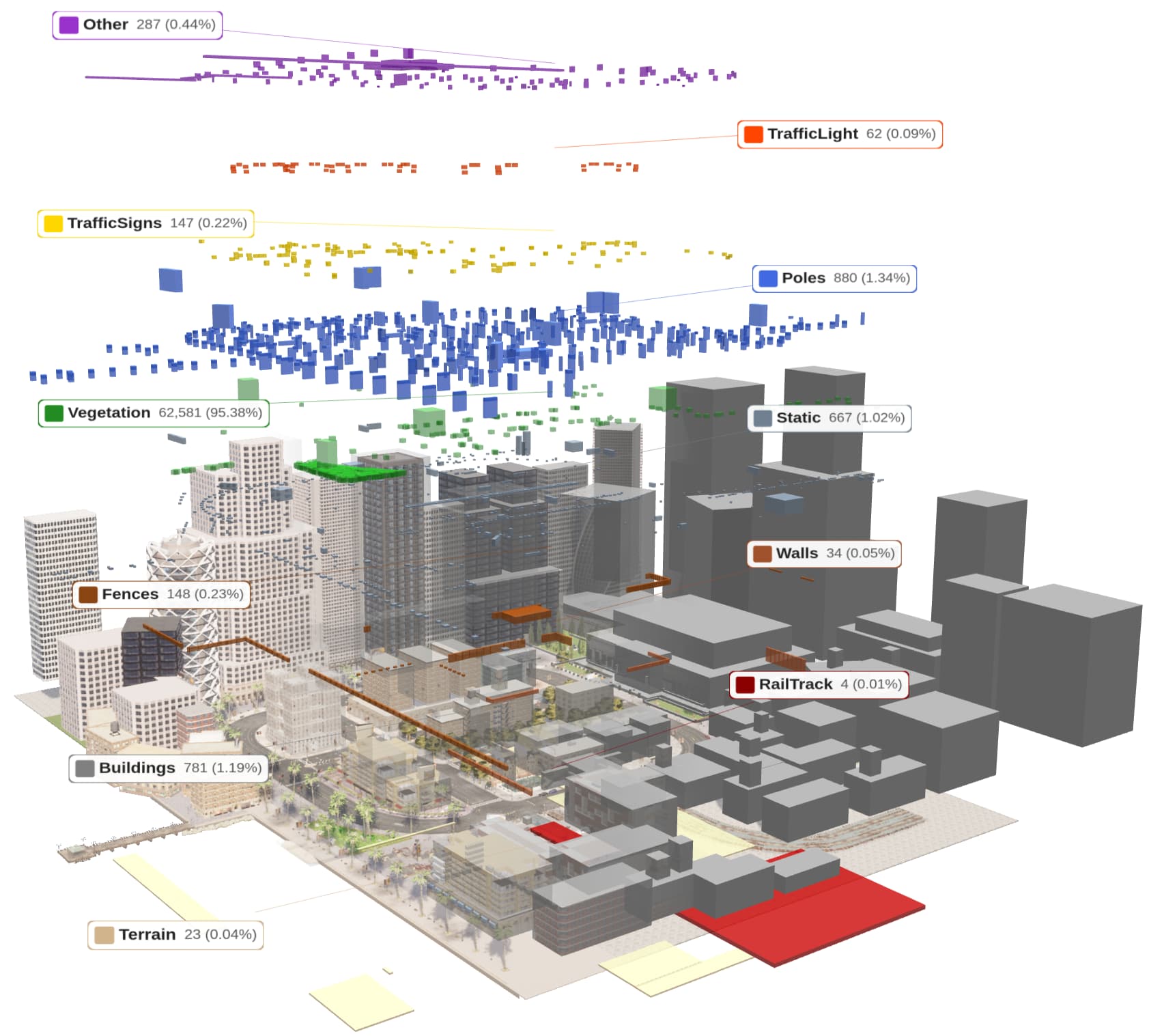}
\captionsetup{font=small}
\caption{3D bounding boxes in Town10HD\_Opt by semantic category (Vegetation dominates; long tail of structural categories visible at the edges). Numeric counts in Appendix~\ref{sec:appendix_g_mask_editor} (\cref{tab:box_categories}).}
\label{fig:box_categories}
\end{figure}

The export yields $65{,}614$ axis-aligned boxes for Town10HD\_Opt. Vegetation alone accounts for $62{,}581$ of them ($95.38\%$); the remaining $\sim$$3{,}033$ boxes form a long tail dominated by Poles, Buildings, Static props, Fences, and traffic infrastructure (per-category counts in Appendix~\ref{sec:appendix_g_mask_editor}, \cref{tab:box_categories}). Because Vegetation also dominates the per-cell collision-check budget of the downstream 2D occupancy grid, we use a dedicated merge and crop operation for Vegetation in \cref{sec:step2} rather than down-sampling all categories uniformly. The category layout that motivates this decision is visualised in \cref{fig:box_categories}.

\subsection{Step 2: 3D Grid Cleaning and Simplification}
\label{sec:step2}

The raw 3D grid from Step~1 contains significant redundancy that would impede efficient trajectory planning. We apply a sequence of simplification operations to reduce the box count while maintaining the main obstacle layout required for downstream planning.

\paragraph{Merging adjacent boxes.}
For each semantic category we cluster boxes by an \emph{AABB-gap} adjacency rule evaluated in 3D: two boxes are deemed adjacent if the per-axis gap between their axis-aligned bounding boxes is at most a category-specific threshold on \emph{every} axis (a gap of zero corresponds to touching boxes and a negative gap to overlap, both of which count as adjacent). Each cluster is then replaced by its enclosing AABB. We use a 2\,m threshold for Vegetation, matching the typical sub-meter spacing of CARLA's tree-canopy sub-boxes, and a 5\,m threshold for Buildings, matching typical wall-segment alignment gaps. The chosen values err on the conservative (slightly larger) side to keep planning collision-safe.

\paragraph{Crop tree operation.}
For every Vegetation box we split the original AABB into a \emph{trunk} sub-box and a \emph{canopy} sub-box at a fixed cut height $h_{\mathrm{cut}}=2.0$\,m above ground. The trunk sub-box keeps the original $xy$ extent and spans $[0, h_{\mathrm{cut}}]$ in $z$; the canopy sub-box keeps the original $xy$ extent and spans the original $z$-range above $h_{\mathrm{cut}}$, additionally lifted upward by a small clearance $\Delta_{\mathrm{lift}} = 0.5$\,m so that the gap between trunk and canopy is large enough for the 2\,m pedestrian height interval used in \cref{sec:step3}. This decomposition is necessary because the original CARLA tree boxes often extend as a single cuboid from ground level upward, which would incorrectly block pedestrian motion beneath the tree and make such trajectories infeasible during planning. After the split, the trunk remains a local hard obstacle near the ground, while the canopy represents overhead occlusion. The resulting gap is geometrically conservative for collision checking but, by construction, does not represent a literal physical opening in the original asset; we revisit this assumption in \cref{sec:limitations}.

\paragraph{Below-ground removal and nested box pruning.}
We remove a box only if its \emph{top face} lies below ground ($z_{\mathrm{center}} + e_z < 0$), rather than using a pure centroid criterion, so partially-buried boxes whose top still rises above the ground plane are preserved. Nested boxes are pruned only when their AABB is fully contained inside another AABB of the \emph{same} semantic category, with a $1\,\mathrm{cm}$ numerical tolerance on every axis; cross-category containment (for example a traffic-light box inside a larger building box) is left untouched to avoid silently discarding semantic labels.

\begin{figure}[!tb]
\centering
\includegraphics[width=0.66\columnwidth]{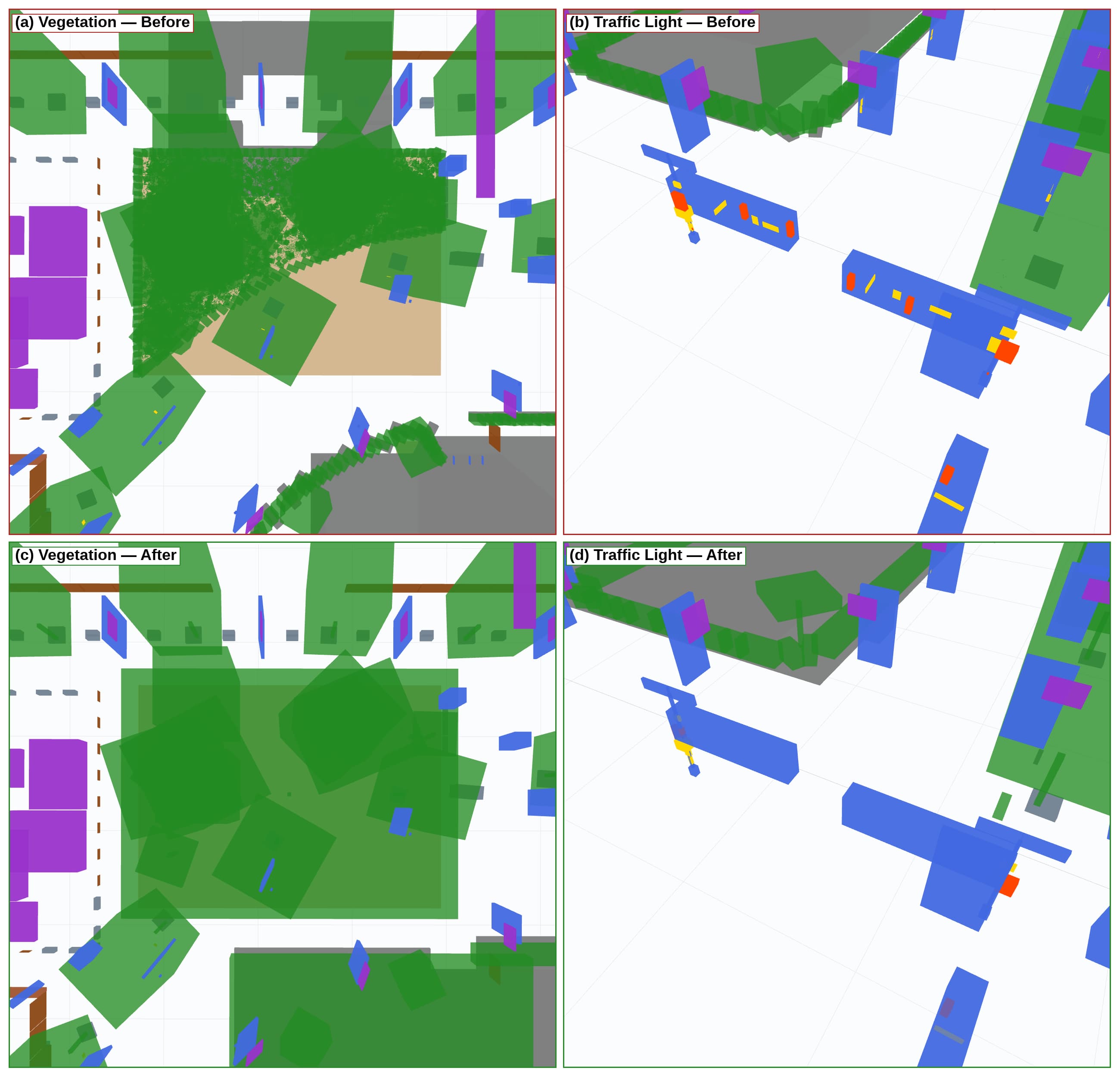}
\captionsetup{font=small}
\caption{Grid simplification: (a/c) Vegetation before/after merging; (b/d) traffic lights and poles before/after pruning.}
\label{fig:grid_simplification}
\end{figure}

As shown in \cref{fig:grid_simplification}, these operations reduce the total box count from 65,614 to 2,067 (about $32\times$). Vegetation drops from 62,581 to 381 and Buildings from 781 to 218; the remaining categories decrease from 2,252 to 1,468. Cross-validating the simplified map against the original $65{,}614$-box export on a $1239 \times 1014$ pedestrian-height occupancy grid yields $96.5\%$ cell-level agreement and an intersection-over-union of $0.92$ on the occupied cells, with discrepancies concentrated inside merged tree canopies and building blocks rather than in walkable corridors. The per-cell breakdown that produces these numbers is shipped with the reproduction script in the release repository.

\FloatBarrier

\subsection{Step 3: Batch Pedestrian Trajectory Generation}
\label{sec:step3}

Pedestrian trajectories define the ground-level paths that the aerial tracking target follows. We generate these trajectories using a grid-based approach with A* path planning~\citep{hart1968astar}.

\paragraph{2D grid discretization.}
The ground plane is discretized into a 2D occupancy grid at $0.5$\,m resolution. For Town10HD\_Opt this yields a $1238 \times 1013$ cell grid in the world frame, with the world-to-grid map $g_x = \lfloor (x-x_{\min})/0.5 \rfloor$, $g_y = \lfloor (y-y_{\min})/0.5 \rfloor$ (full derivation in \cref{sec:appendix_g_mask_editor}). A cell is occupied iff any 3D box's vertical extent has non-empty intersection with the human height interval $[0, 2.0]$\,m. All occupied cells are dilated by a $0.5$\,m safety margin matching the lateral half-width of a walking adult.

\paragraph{Connected component analysis.}
We label connected components on the free cells with 8-neighbour connectivity and discard components below $4\,000$ cells ($\approx 1\,000\,\mathrm{m}^2$). Start/end points are then drawn from the same retained component, guaranteeing an A* solution.

\begin{figure}[!htb]
\centering
\includegraphics[width=0.83\columnwidth]{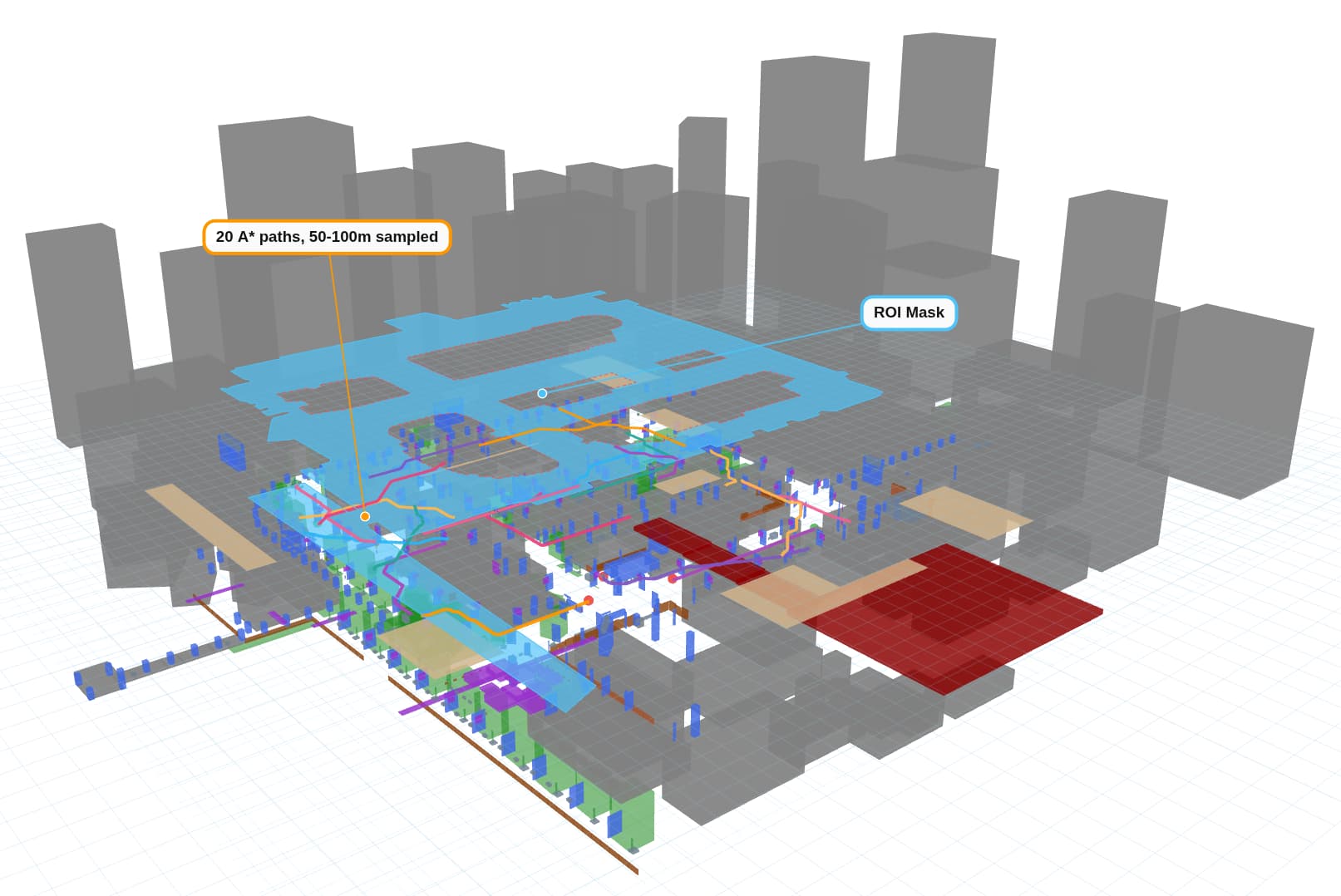}
\captionsetup{font=small}
\caption{Pedestrian trajectories in Town10HD\_Opt: 20 A* paths sampled in the ROI-masked walkable region (light blue) under a 50--100\,m start-to-end constraint.}
\label{fig:pedestrian_trajectory_generation}
\end{figure}

\paragraph{Start/end point sampling and path planning.}
Start and end points are sampled uniformly at random under a \emph{Euclidean} distance constraint of 50--100\,m, which matches our short-range UAV-escort deployment scenarios. A*~\citep{hart1968astar} is then run on the masked, inflated grid, with up to $5$ retries per requested trajectory. \Cref{fig:pedestrian_trajectory_generation} shows a representative batch of 20 such trajectories.

\paragraph{ROI polygon annotation.}
Region of Interest (ROI) polygons are used to restrict walkable areas, effectively excluding non-walkable regions such as water bodies and map edges from the sampling space. The ROI is authored once per map using a dedicated map-registered polygon editor that operates in the same coordinate frame as the planning grid; the tool, its export format, and its integration with the inflated free-space grid are detailed in \cref{sec:appendix_g_mask_editor}. \Cref{tab:ped_params} summarizes the key parameters.

\begin{table}[t]
\centering
\caption{Pedestrian trajectory generation parameters.}
\label{tab:ped_params}
\begin{tabular}{lll}
\toprule
Parameter & Value & Description \\
\midrule
Grid resolution & 0.5\,m & 2D grid discretization \\
Human height & 2.0\,m & Height for obstacle overlap check \\
Safety radius & 0.5\,m & Inflation radius for obstacles \\
Min point distance & 50\,m & Minimum start-end distance \\
Max point distance & 100\,m & Maximum start-end distance \\
Ground $z$ & 0.0\,m & Ground plane height \\
\bottomrule
\end{tabular}
\end{table}

\paragraph{Variable speed modeling.}
We approximate pedestrian dynamics by resampling the A* polyline with a curvature-dependent speed. The Menger curvature at interior waypoint $i$ from $(p_{i-1}, p_i, p_{i+1})$ is
\begin{equation}
\label{eq:menger}
\kappa_i = \frac{2 \, \lvert (p_{i+1} - p_i) \times (p_{i-1} - p_i) \rvert}{\max\!\bigl(\|p_{i+1}-p_i\| \cdot \|p_{i-1}-p_i\| \cdot \|p_{i+1}-p_{i-1}\|, \varepsilon_\kappa\bigr)},
\end{equation}
with $\varepsilon_\kappa = 10^{-6}\,\mathrm{m}^3$ for numerical safety (we additionally set $\kappa_i = 0$ when the unguarded denominator falls below $\varepsilon_\kappa$). The instantaneous speed is
\begin{equation}
\label{eq:speed}
v_i = \mathrm{clip}\!\left(\tfrac{v_{\mathrm{cruise}}}{1 + \alpha \kappa_i} (1 + \beta\, \mathcal{U}(-1,1)),\; [v_{\min}, v_{\max}]\right),
\end{equation}
with default $v_{\mathrm{cruise}} = 1.2\,\mathrm{m/s}$, $\alpha = 5.0\,\mathrm{m}$, $\beta = 0.15$ and clip range $[0, 1.6]\,\mathrm{m/s}$. We convert the A* polyline into a time-stamped trajectory using $\Delta t_i = \|p_{i+1}-p_i\| / \tfrac{1}{2}(v_i+v_{i+1})$ and resample at the pipeline-wide time step $dt$ shared with \cref{sec:step4}, so every pedestrian time stamp has a matching drone waypoint.

\paragraph{ROI authoring.}
The ROI polygons are authored manually once per map using the editor in \cref{sec:appendix_g_mask_editor} (a few minutes per map, reused across all Step~3 runs). The trade-off introduced by this human-in-the-loop step is discussed in \cref{sec:limitations}.

\subsection{Step 4: Batch Drone Trajectory Computation}
\label{sec:step4}

Given pedestrian trajectories from Step~3, we compute drone trajectories that maintain aerial tracking of the ground-level target. As stated in the section opening, the current public release is produced by MuCO and TA*+Smooth is retained for the paradigm-level comparison only; both planners share the interface in \cref{tab:drone_params}. The full algorithmic specifications (post-smoothing routine, projection details, building-circling mitigation, and complete loss/weight tables) are given in \cref{app:planner_details}.

\paragraph{TA*+Smooth (two-stage).}
The frontend is the visibility-aware Track A*~\citep{chen2026trackastar} search on a 4D spatio-temporal voxel grid (default voxel size $4{\times}4{\times}4$\,m, beam width $2048$, five-ray visibility test, corridor margin $45$\,m), which directly returns a discrete drone trajectory that respects target visibility and obstacle clearance. The backend then applies a \emph{post-smoothing} stage consisting of (i) a shortcut pass that attempts straight-line replacements over spans of up to $12$ frames, and (ii) up to $30$ iterations of an elastic-band relaxation with step $\alpha = 0.35$; each candidate update is accepted only if the per-frame visibility drop is within $5\,\mathrm{pp}$, the average visibility is preserved within $5\,\mathrm{pp}$, and the minimum obstacle distance remains $\geq$ the safety distance.

\paragraph{MuCO (one-shot multi-constraint optimization).}
MuCO jointly adjusts every interior waypoint by finite-difference gradient descent ($\varepsilon = 0.5$\,m, learning rate $0.05$, per-iter step cap $0.5$\,m, up to $1500$ outer iterations, convergence at $|\Delta L| < 10^{-5}$). The loss decomposes into seven weighted terms---\emph{tracking}, \emph{smoothness}, \emph{jerk}, \emph{safety}, \emph{visibility}, \emph{view angle}, and \emph{path length}---plus fixed-coefficient altitude regularisers. The weight values reported in \cref{tab:drone_params} were obtained through a small number of manual tuning rounds (not a systematic grid search) and remain conservative defaults; the precise loss formulae are given in \cref{subsec:appendix_planner_muco}. The optimizer is offline (single batch optimization with no rolling horizon and no closed-loop feedback), but the per-term cost structure mirrors that of classical receding-horizon control~\citep{kouvaritakis2016mpc}.

\begin{table}[!b]
\centering
\small
\setlength{\tabcolsep}{8pt}
\caption{Shared interface parameters of the two planners. Internal hyper-parameters of each planner (TA* frontend, post-smoothing backend, and MuCO optimizer) are listed in \cref{app:planner_details}.}
\label{tab:drone_params}
\begin{tabular}{@{}lcc@{}}
\toprule
Parameter & TA*+Smooth & MuCO \\
\midrule
Pipeline $dt$ (rendered trajectory sampling rate) & 0.5\,s (2\,Hz) & 0.5\,s (2\,Hz) \\
Safety distance (nominal / relaxed) & 3.0 / 2.5\,m & 3.0 / 2.5\,m \\
Min / preferred / max altitude & 20 / 20 / 100\,m & 20 / 20 / 100\,m \\
Behind distance from target & 20.0\,m & 20.0\,m \\
Max drone velocity & 10.0\,m/s & 10.0\,m/s \\
Tracking target distance $d_{\mathrm{opt}}$ & --- & 28.0\,m \\
\bottomrule
\end{tabular}
\end{table}

\paragraph{Safety distance and obstacle projection.}
Both planners use a nominal safety distance of $3.0$\,m and a relaxed floor of $2.5$\,m. In MuCO, the relaxed value is engaged only when projection cannot reach the nominal floor within the iteration budget; it never disables hard-collision rejection or the per-iteration push-out. The full obstacle-projection logic is documented in \cref{subsec:appendix_planner_muco}.

\begin{figure}[!tb]
\centering
\includegraphics[width=0.82\columnwidth]{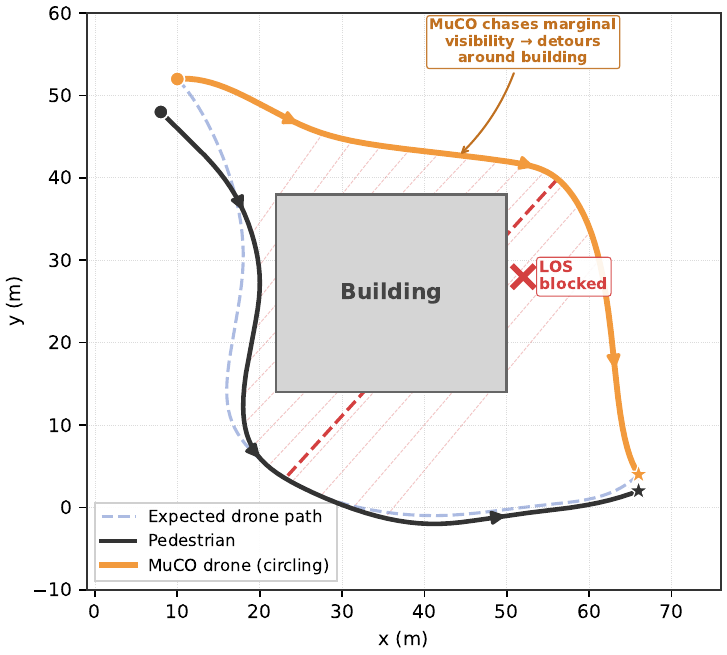}
\captionsetup{font=small}
\caption{Building-circling failure mode (top-down schematic). When a building blocks line-of-sight for an extended segment, MuCO's visibility gradient can pull the drone around the far side of the obstacle (orange) instead of following the pedestrian (black). The dashed blue curve shows the expected path. Red dashed lines mark blocked LOS rays through the building.}
\label{fig:building_circling}
\end{figure}

\paragraph{Building-circling mitigation.}
We observed that, on long line-of-sight blockages, MuCO can chase marginal visibility gains and produce loops around buildings (\cref{fig:building_circling}). The mitigation is implemented inside the optimizer rather than as ad-hoc post-processing: persistent low-visibility runs trigger a temporary zero-out of the visibility and view-angle gradients while the remaining terms continue to drive the trajectory; the rule and its parameters are given in \cref{subsec:appendix_planner_muco}.

\begin{figure}[!tb]
\centering
\includegraphics[width=0.92\columnwidth]{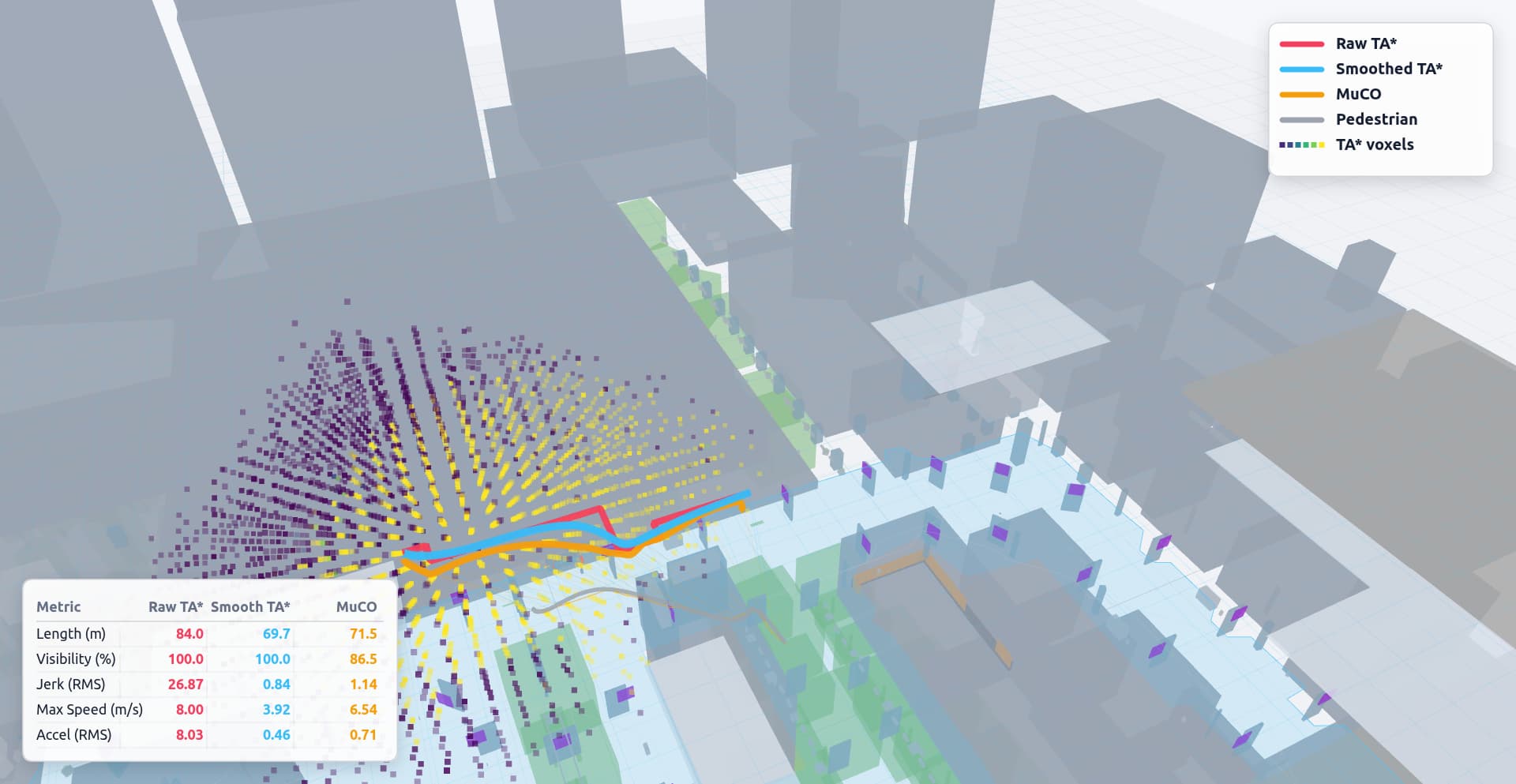}
\captionsetup{font=small}
\caption{TA*+Smooth (red = raw TA*, blue = post-smoothed) vs.\ MuCO (orange) on a Town10HD\_Opt sample (Path~1). Grey is the pedestrian trajectory; coloured voxels expose the TA* search frontier. Numerical comparison is on Path~0 in \cref{tab:algo_comparison}.}
\label{fig:trajectory_comparison}
\end{figure}

\begin{table}[!b]
\centering
\small
\setlength{\tabcolsep}{6pt}
\caption{Planner comparison on Town10HD\_Opt / Path~0 ($186$ waypoints, $2{,}067$ obstacles), measured from the released planner outputs. Path length for TA*+Smooth is the smoothed-trajectory length (raw TA* path length was $108.0$\,m). Avg.\ visibility is the planner-internal 5-ray line-of-sight metric and is distinct from the depth-buffer-based visibility used by the Step~6 dataset filter. Cells with ``---'' are inapplicable.}
\label{tab:algo_comparison}
\begin{tabular}{lrr}
\toprule
Metric & TA*+Smooth & MuCO \\
\midrule
Path length (m) & $92.5$ & $102.4$ \\
Planning time (ms) & $906$ & $958$ \\
$\quad$ A* search / post-smooth (ms) & $451$ / $455$ & --- \\
Outer iterations & --- & $12$ \\
Accepted shortcuts / elastic updates & $287$ / $1{,}095$ & --- \\
Avg.\ target distance (m) & $27.5$ & $29.2$ \\
Avg.\ visibility (planner-internal, 5-ray) & $0.999$ & $0.952$ \\
Smoothed jerk RMS / accel RMS & $0.81$ / $0.35$ & --- \\
Hard-collision waypoints (clearance $<0$\,m) & $0$ & $0$ \\
\bottomrule
\end{tabular}
\end{table}

\subsection{Step 5: Simulator Rendering and Data Collection}
\label{sec:step5}

The drone trajectories are replayed in CARLA in synchronous mode at $\Delta t_{\mathrm{sim}} = 0.05$\,s. A single rendering tick teleports the drone, ticks the simulator, waits for all three sensor callbacks with the same frame ID, and writes the triplet to disk, guaranteeing one shared tick and one shared pose across RGB, depth, and segmentation. Sensors are co-located at the drone body origin with near/far clip $0.3$/$1000$\,m. To improve visual diversity and domain robustness, we inject weather and time-of-day (ToD) variations via a configurable augmentation module: 15 weather presets (clear, rain, fog, haze groups) $\times$ 4 ToD presets (morning, noon, dusk, night) yield 60 unique atmospheric configurations that are sampled per trajectory (see \cref{app:weather} for the full preset taxonomy, parameter schema, and selection modes). We use fixed-exposure manual mode with motion blur disabled. Default resolution is $1280{\times}720$. The per-frame outputs are
\begin{itemize}[nosep]
    \item \textbf{RGB} ($1280{\times}720$, 8-bit) from the drone viewpoint;
    \item \textbf{depth} as a float32 NumPy array storing camera-frame $z$-depth (perpendicular distance to the image plane, not Euclidean ray distance) in the $[0, 1000]$\,m range; sky pixels carry the sentinel value $1000.0$ and invalid pixels carry $0.0$;
    \item \textbf{semantic segmentation} using CARLA's built-in labels.
\end{itemize}

\paragraph{Configurable fixed-FOV zoom levels.}
\cosfly supports zoom by varying the horizontal field-of-view (FOV); CARLA exposes no millimetre-scale focal length. The equivalent pixel focal length is
\begin{equation}
\label{eq:focal}
f_{\text{pixels}} = \frac{W}{2 \cdot \tan(\text{FOV} \cdot \pi / 360)},
\end{equation}
where $W$ is the image width. The pipeline supports four fixed FOV levels from $30^\circ$ (telephoto) to $110^\circ$ (wide-angle), summarized in \cref{tab:zoom_config}. In every public trajectory the FOV is held \emph{constant for the entire trajectory} (one level drawn per trajectory), so the public release is a union of four FOV-disjoint sub-pools. We do not vary FOV \emph{within} a trajectory because CARLA requires destroying and recreating the camera actor to change FOV, which would break the per-tick sensor synchronization above. The intrinsic matrix corresponding to each FOV is written into the per-frame JSON annotation.

\begin{table}[!tb]
\centering
\caption{Zoom configuration options and equivalent focal lengths.}
\label{tab:zoom_config}
\begin{tabular}{lrrr}
\toprule
Zoom Level & FOV ($^\circ$) & Equiv.\ Focal Length (px) & Use Case \\
\midrule
Wide-angle & 110 & 485 & Overview shots \\
Standard & 90 & 640 & Default tracking \\
Narrow & 60 & 1109 & Close-up tracking \\
Telephoto & 30 & 2391 & Long-range surveillance \\
\bottomrule
\end{tabular}
\end{table}

Trajectory-level random perturbations (joint position + orientation; see \cref{app:dual_track} for the full parameter table) inject diversity without breaking temporal coherence: each frame falls back to the unperturbed pose whenever a perturbed pose would push the target out of the frustum or violate safety clearance.

\begin{figure}[!tb]
\centering
\includegraphics[width=0.66\columnwidth]{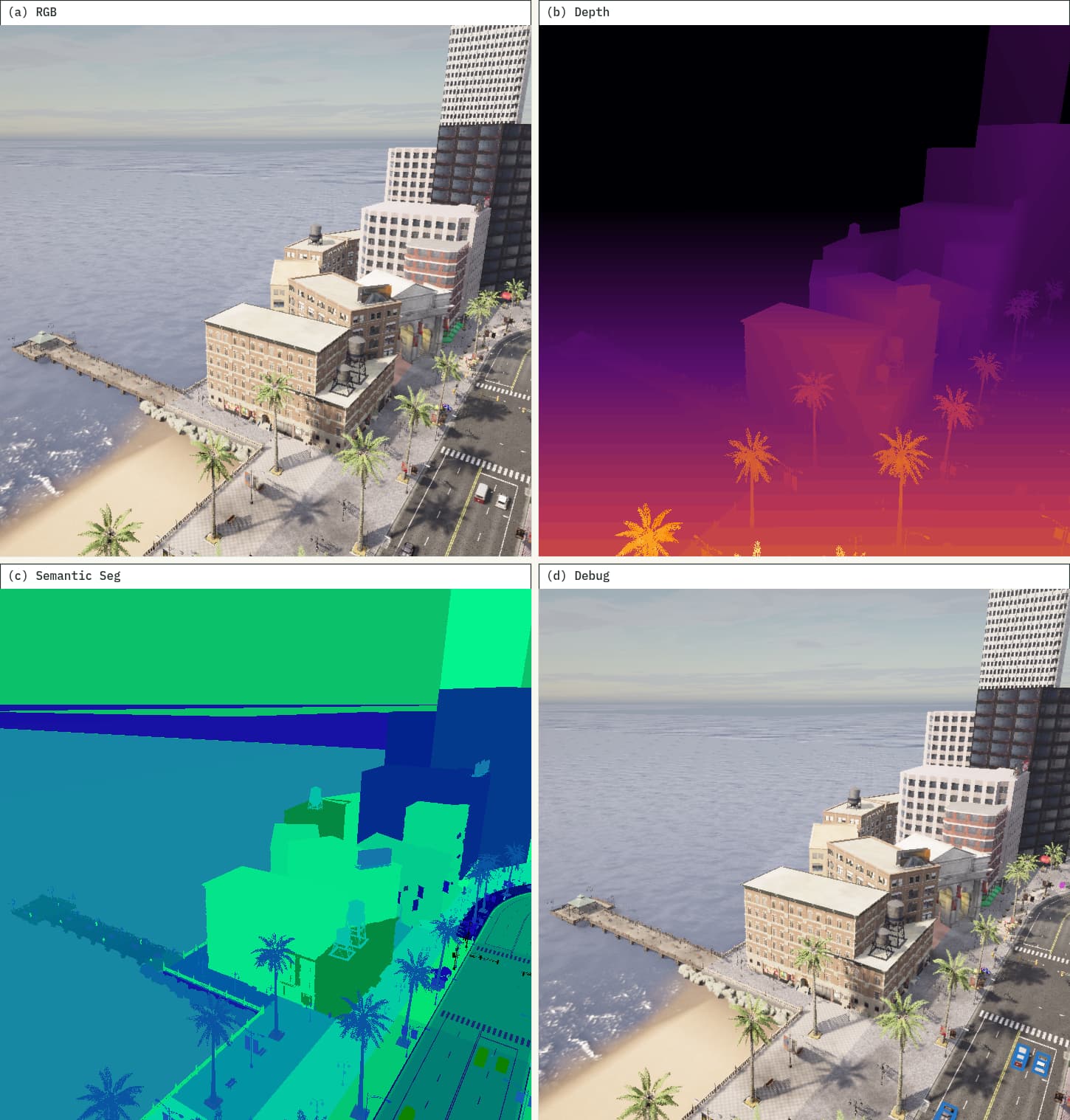}
\caption{One synchronized sample: (a) RGB, (b) depth (display rendering of float32 array), (c) segmentation, (d) 3D-box reprojection debug view. All four panels share the same simulator tick (\cref{sec:step5}).}
\label{fig:dataset_sample}
\end{figure}

\subsection{Step 6: Data Quality Inspection}
\label{sec:step6}

Step~6 is a semi-automated validation framework: per-trajectory smoothness, visibility, and rendering-artifact metrics are computed automatically, and trajectories falling outside the automated thresholds are routed to a human reviewer. In this paper we report these metrics on a 20-trajectory \emph{pilot subset} produced by the released MuCO planner on Town10HD\_Opt (\cref{tab:dist_stats,tab:quality_stats}); the same rules are applied to the full multi-map production batch that yields the $250$-trajectory first public release (\cref{sec:dataset_stats}).

\paragraph{Smoothness.}
For every drone trajectory we measure the RMS acceleration $a_{\mathrm{rms}}$ and the RMS jerk $j_{\mathrm{rms}}$ by discrete differentiation of the planned waypoints. We reject trajectories with $a_{\mathrm{rms}} > 5\,\mathrm{m/s}^2$ or $j_{\mathrm{rms}} > 10\,\mathrm{m/s}^3$ (conservative envelopes for small multirotors) and summarize the dynamics by a scalar smoothness score
\begin{equation}
\label{eq:smoothness_score}
S = \exp\!\Bigl(-\tfrac{1}{2}\bigl(\tfrac{a_{\mathrm{rms}}}{5\,\mathrm{m/s}^2} + \tfrac{j_{\mathrm{rms}}}{10\,\mathrm{m/s}^3}\bigr)\Bigr) \in (0,1],
\end{equation}
with $S < 0.5$ routed to manual review.

\paragraph{Visibility (two-stage funnel).}
\cosfly applies two distinct visibility metrics at two different stages of the funnel. \emph{Planner-internal 5-ray visibility} is computed at planning time by casting five rays from the drone to the target and counting the fraction unblocked by the obstacle grid; this quantity is aggregated per trajectory and used as the \emph{pre-render visibility prefilter}, rejecting any trajectory whose mean is below $40\%$. \emph{Rendered depth-buffer visibility} is computed only after rendering, as the per-frame fraction of the target's projected 3D bounding box that is unoccluded inside the camera frustum (depth-buffer test, cross-checked against pedestrian-labelled pixels in the segmentation mask); it is reported per FOV configuration in \cref{tab:zoom_eval} and is not used as a rejection rule in the pilot reported here. The $0.906$ pilot mean reported in \cref{tab:dist_stats,tab:quality_stats} is the planner-internal 5-ray quantity.

\paragraph{Data distribution.}
\Cref{tab:dist_stats} reports the per-axis statistics over the 20-trajectory pilot subset; depth-percentile and semantic-coverage statistics will accompany the full production dataset.

\begin{table}[H]
\centering
\small
\setlength{\tabcolsep}{6pt}
\caption{Per-axis statistics on the 20-trajectory pilot subset (MuCO outputs on Town10HD\_Opt Paths~0--19).}
\label{tab:dist_stats}
\begin{tabular}{lrrr}
\toprule
Axis & Mean & Std & Range \\
\midrule
Trajectory length (m) & $108.8$ & $30.3$ & $[71.5,\ 200.3]$ \\
Altitude (m) & $20.4$ & $1.2$ & $[20.0,\ 28.5]$ \\
Target distance (m) & $28.0$ & $1.0$ & $[25.9,\ 29.2]$ \\
Per-trajectory planner-internal 5-ray visibility & $0.906$ & --- & $[0.641,\ 1.000]$ \\
\bottomrule
\end{tabular}
\end{table}

\begin{figure}[!tb]
\centering
\includegraphics[width=0.72\columnwidth]{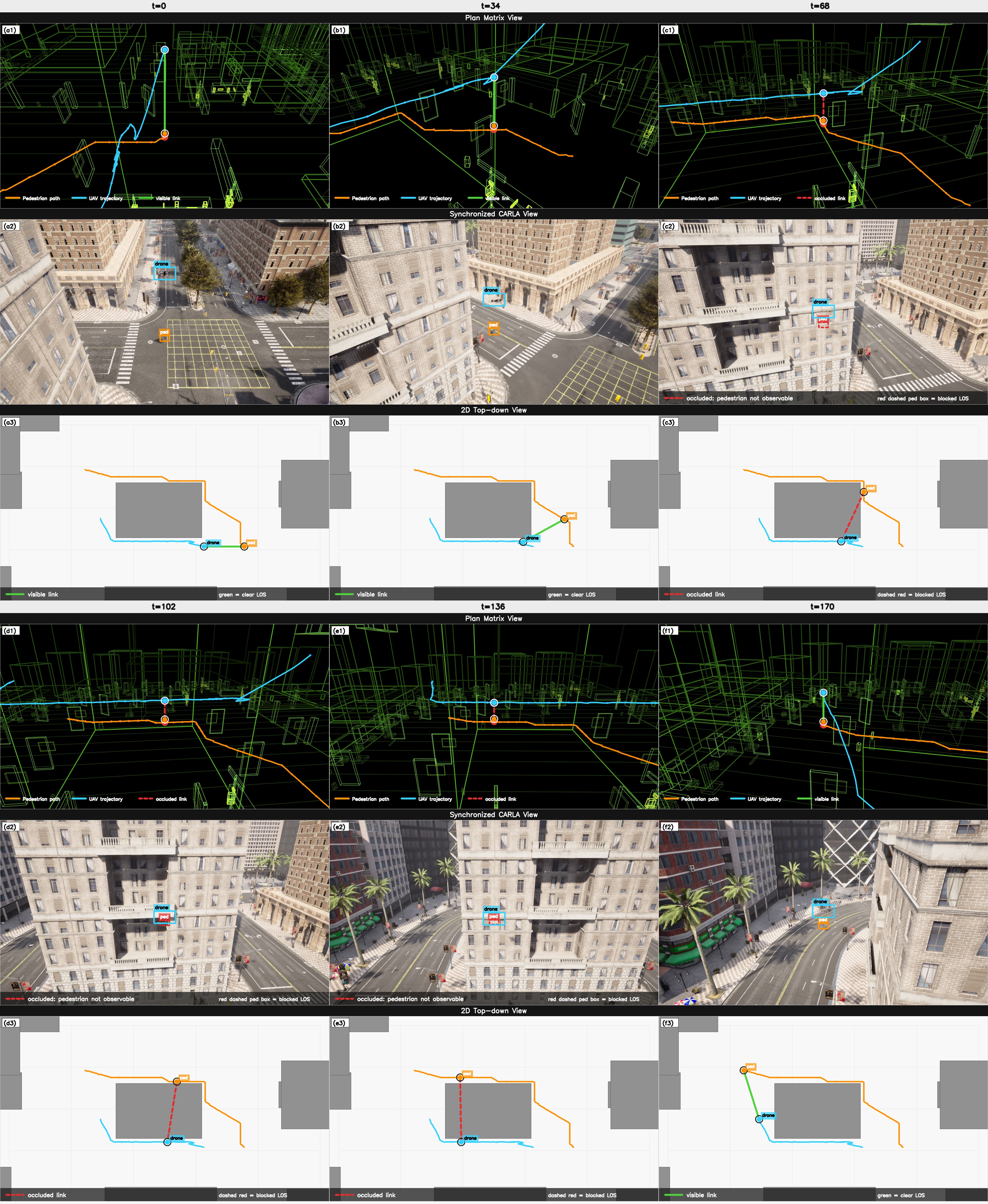}
\captionsetup{font=small}
\caption{A failure case excluded from the release (six synchronized timestamps, each shown as planning matrix / CARLA world view / 2D top-down view; green / red-dashed links denote clear / occluded LOS). The geometric inconsistency around the building disqualifies the sample.}
\label{fig:cosfly_matrix_world_failure}
\end{figure}

\paragraph{Rendering artifacts and filtering funnel.}
Automated checks identify missing textures, Z-fighting, and large temporal inconsistencies between consecutive frames; flagged frames are inspected by a human reviewer before any trajectory enters the public release. A representative failure case excluded from the release is shown in \cref{fig:cosfly_matrix_world_failure}; the filtering funnel applied to the 20-trajectory pilot subset (using the rules and thresholds defined above) is reported in \cref{tab:quality_stats}, and the full production funnel follows the same procedure and is summarised in \cref{sec:dataset_stats}.

\begin{table}[!tb]
\centering
\small
\caption{Planner-stage filtering funnel applied to the 20-trajectory MuCO pilot subset using the pre-render \cref{sec:step6} rules (planner-internal 5-ray visibility prefilter and smoothness envelope). Post-render rendered depth-buffer visibility is reported separately in \cref{tab:zoom_eval}; the rendering-artifact / frame-completeness QC step is run before public release but not aggregated into this pilot table. Aggregate statistics for the full production batch are in \cref{sec:dataset_stats}.}
\label{tab:quality_stats}
\begin{tabular}{lr}
\toprule
Metric & Value \\
\midrule
Pilot trajectories generated & $20$ \\
Rejected by planner-internal 5-ray visibility prefilter ($< 40\%$ per-trajectory) & $0$ \\
Rejected by smoothness envelope ($a_{\mathrm{rms}} > 5$, $j_{\mathrm{rms}} > 10$) & $0$ \\
Pilot trajectories passing planner-stage rules & $20$ \\
Mean per-trajectory planner-internal 5-ray visibility & $0.906$ \\
Mean RMS acceleration / RMS jerk (m/s$^{2}$, m/s$^{3}$) & $0.82$ / $1.85$ \\
Mean smoothness score $S$ (Eq.~\ref{eq:smoothness_score}, pilot) & $0.84$ \\
\bottomrule
\end{tabular}
\end{table}

\subsection{Step 7: Image Captioning}
\label{sec:step7}

\begin{figure}[t]
\centering
\includegraphics[width=0.86\columnwidth]{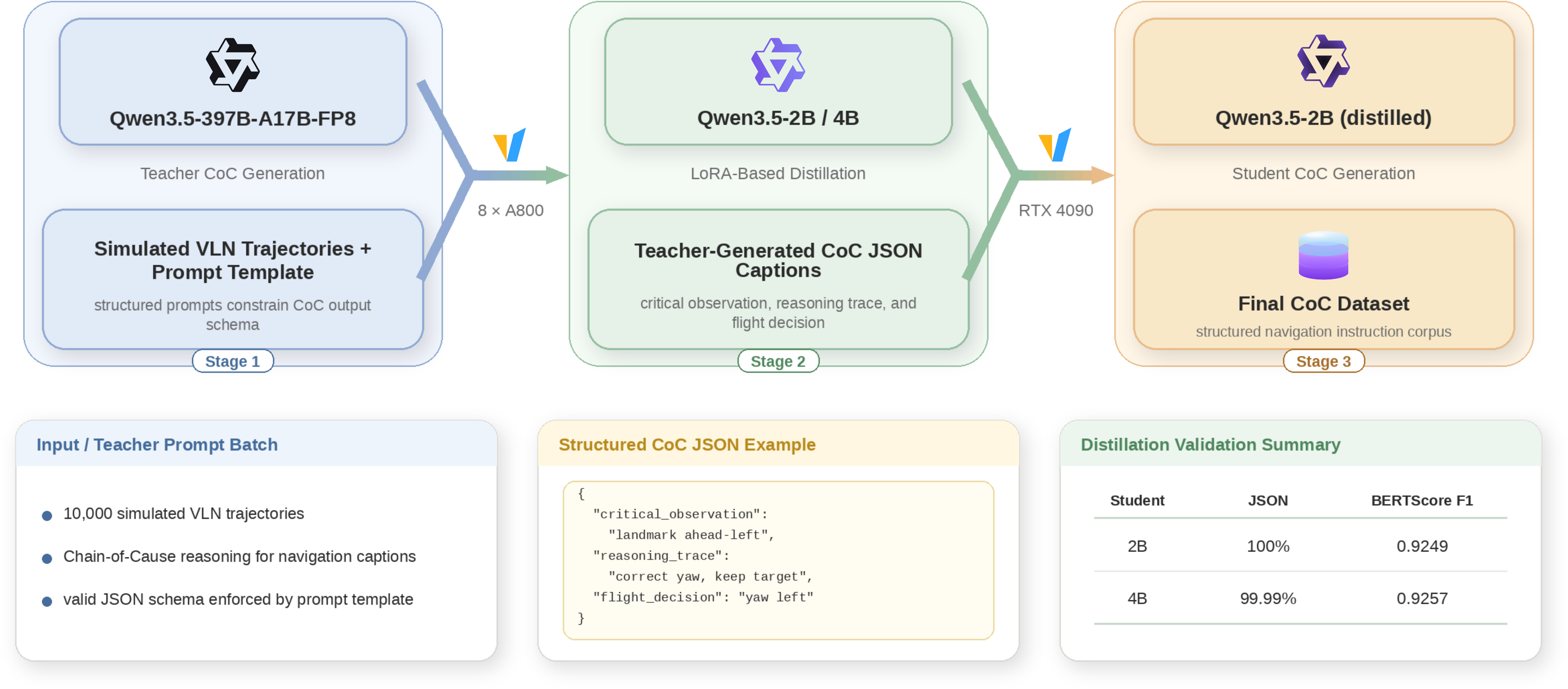}
\caption{Three-stage CoC generation pipeline. A Qwen3.5-397B-A17B-FP8 teacher generates structured CoC reasoning via distributed vLLM inference; LoRA distillation to Qwen3.5-2B/4B students achieves BERTScore F1 $\approx$ 0.925.}
\label{fig:caption_pipeline}
\end{figure}

We generate structured \emph{Chain-of-Cause} (CoC) annotations using a teacher--student distillation pipeline~\citep{hinton2015distilling} with LoRA-based finetuning~\citep{hu2021lora}, evaluated with BERTScore~\citep{zhang2020bertscore} as a lexical similarity proxy. The teacher is an internal Qwen3.5-397B-A17B-FP8 vision-language checkpoint used only inside our group for offline labeling; the deployment configuration (parameter footprint and quantization) is described in \cref{sec:appendix_e}. The students are the public Qwen3.5-2B and 4B base models. \Cref{fig:caption_pipeline} illustrates the three stages.

\paragraph{Stage 1: teacher labeling.}
For every target frame the teacher is conditioned on a 5-frame historical sliding window of RGB + depth pseudo-colour + segmentation with the pedestrian highlighted + 6-DOF drone pose + target world position + rendered depth-buffer visibility flag (\cref{sec:step6}). It returns one JSON object per frame with three fields: critical components (target state, obstacles, tracking geometry), a reasoning trace (one causal sentence), and a flight decision (one of $17$ tokens from the closed action set in \cref{tab:coc_actions}). A representative output and the verification logic (JSON validity, decision in closed set, geometric consistency vs.\ the planner's next $5$ frames, up to three constrained re-generations) are documented in \cref{sec:appendix_f_caption_distillation}. Outputs are bilingual (English first, then a fixed-glossary Chinese mirror); samples failing a per-sample bilingual consistency check are dropped before training.

\begin{table}[!tb]
\centering
\small
\caption{17-option closed action set used as the flight-decision value.}
\label{tab:coc_actions}
\setlength{\tabcolsep}{6pt}
\begin{tabular}{@{}lll@{}}
\toprule
Tracking (1) & Heading (4) & Vertical (2) \\
\midrule
\texttt{track\_straight} & \texttt{yaw\_left\_to\_follow} & \texttt{ascend} \\
 & \texttt{yaw\_right\_to\_follow} & \texttt{descend} \\
 & \texttt{circle\_left} & \\
 & \texttt{circle\_right} & \\
\midrule
Translation (4) & Speed (2) & Recovery / Modes (4) \\
\midrule
\texttt{move\_forward} & \texttt{slow\_down} & \texttt{search\_to\_reacquire\_target} \\
\texttt{move\_backward} & \texttt{speed\_up} & \texttt{hover} \\
\texttt{strafe\_left} & & \texttt{break\_off} \\
\texttt{strafe\_right} & & \texttt{return\_to\_home} \\
\bottomrule
\end{tabular}
\end{table}

\paragraph{Stage 2: student distillation.}
Qwen3.5-2B and 4B are LoRA-finetuned on the teacher labels; on 10\,k validation samples they reach BERTScore F1 $\approx 0.925$ and exact-match flight-decision accuracy of $68.7$--$70.1\%$ (full breakdown in \cref{sec:appendix_f_caption_distillation}). BERTScore is reported as a lexical-similarity proxy; causal correctness is carried by the geometric-consistency rule above and by the closed-set decision accuracy.

\paragraph{Stage 3: batch inference.}
We deploy the 2B student even though the 4B model is marginally better in BERTScore ($0.9257$ vs.\ $0.9249$) and decision accuracy ($70.07\%$ vs.\ $68.70\%$): the 2B model halves the parameter footprint, runs at $\sim 1.7\times$ throughput on the same inference batch, and achieves a $100\%$ JSON parse rate ($10\,000/10\,000$) versus $99.99\%$ for the 4B. The $100\%$ parse rate is a format-stability statement and is reported separately from the BERTScore-based semantic-quality statement. \Cref{tab:caption_pipeline} summarizes the three stages.

\begin{table}[t]
\centering
\small
\setlength{\tabcolsep}{6pt}
\caption{CoC generation pipeline. The teacher is an internal Qwen3.5-397B-A17B-FP8 checkpoint, not a publicly released model.}
\label{tab:caption_pipeline}
\begin{tabular}{@{}lll@{}}
\toprule
Stage & Model & Key metric \\
\midrule
Teacher labeling & Qwen3.5-397B-A17B-FP8 & Structured CoC JSON \\
Student distillation & Qwen3.5-2B/4B + LoRA & BERTScore F1 $\approx 0.925$ \\
Final inference & Qwen3.5-2B student & 100\% JSON parse rate \\
\bottomrule
\end{tabular}
\end{table}

\FloatBarrier

\section{Dataset Construction and Statistics}
\label{sec:dataset}

The public \cosfly dataset is constructed from multiple CARLA maps spanning diverse urban layouts. In this chapter we report performance benchmarks and ablations on the Town10HD\_Opt map---a representative dense-urban scene---to characterise each pipeline stage in isolation. Following the same modular convention as \cref{sec:pipeline}, we benchmark four distinct stages--pedestrian trajectory generation (Stage A), MuCO drone planning (Stage B), TA*+Smooth drone planning (Stage C), and CARLA rendering (Stage D)--each launched through a unified stage-level watchdog (\cref{sec:dataset_watchdog}) so that auto-restart counts are reported as a first-class reproducibility metric. The release artifact documents the benchmark protocol and raw logs needed to reproduce all results in this section.

\subsection{Hardware and Software Setup}
\label{sec:dataset_setup}

All measurements are collected on the same workstation: Intel Core i9-14900KF (32 logical cores), NVIDIA RTX 6000 Ada Generation GPU (48\,GiB VRAM, 49{,}140\,MiB reported by \texttt{nvidia-smi}), 62\,GiB RAM, NVMe SSD, Ubuntu 22.04 LTS, kernel 6.8, CUDA 12.4, NVIDIA driver 550-series. The rendering stack uses CARLA 0.9.16 on Unreal Engine 4.26 with the headless server reached via the official Python API. Python 3.10 hosts the orchestration and watchdog code; the two planner backends (MuCO and TA*+Smooth) are compiled Rust implementations invoked through the benchmark harness used for this report. The stage-level watchdog samples a heartbeat file every 2.0\,s and the resource probe samples CPU/GPU/RSS every 1.5\,s.

\subsection{Stage A: Pedestrian Trajectory Generation}
\label{sec:dataset_ped}

Stage A runs the pedestrian trajectory generation module in pipeline mode against the Town10HD\_Opt simplified map ($2{,}067$ boxes, $1{,}238\times1{,}013$ pedestrian-height grid at $0.5$\,m cell size as defined in \cref{sec:step3}). The pipeline executes seven deterministic sub-stages--grid construction, projection, inflation, connectivity analysis, sampling, A* search, and report generation--in a fixed order; sub-stage wall-times are captured by the orchestrator. \Cref{tab:stage_breakdown} reports a representative end-to-end run for $N = 20$ trajectories. The full pipeline completes in $3.2$\,s with a peak RSS of $72.7$\,MiB and no watchdog restart, dominated by the A* search ($0.90$\,s, 27.9\%), the connectivity / sample / report triplet ($\sim$58\%), and the obstacle grid build ($\sim$14\%). On this single-map, single-configuration pilot, pedestrian planning is negligible relative to drone planning or rendering ($<0.5\%$ of the total per-trajectory budget) and can be re-executed every release without budgeting concerns.

\begin{table}[t]
\centering
\caption{Stage A breakdown: pedestrian trajectory generation on Town10HD\_Opt for $N = 20$ trajectories ($1{,}238\times1{,}013$ grid at $0.5$\,m cell size). Sub-stage times are from one end-to-end run with the release configuration; peak RSS includes all child processes.}
\label{tab:stage_breakdown}
\begin{tabular}{lrr}
\toprule
Sub-stage & Wall-time (s) & Share (\%) \\
\midrule
grid & 0.126 & 3.9 \\
project & 0.154 & 4.8 \\
inflate & 0.174 & 5.4 \\
connectivity & 0.593 & 18.4 \\
sample & 0.661 & 20.6 \\
astar & 0.898 & 27.9 \\
report & 0.611 & 19.0 \\
\midrule
\textbf{total} & \textbf{3.216} & \textbf{100.0} \\
\midrule
\multicolumn{3}{l}{\emph{peak RSS (Python self+children): 72.7 MiB}} \\
\bottomrule
\end{tabular}
\end{table}

\begin{table}[t]
\centering
\caption{Stage B/C: drone trajectory planning over the same 20 pilot trajectories. Distribution statistics (mean, p50, p95, max) for per-trajectory wall-time and aggregate quality metrics. Visibility is the planner-internal 5-ray metric.}
\label{tab:planner_dist}
\resizebox{\columnwidth}{!}{%
\begin{tabular}{lrrrrrr}
\toprule
Planner & Mean (ms) & p50 (ms) & p95 (ms) & Max (ms) & Length (m) & Visibility \\
\midrule
MuCO & 218 & 230 & 274 & 282 & 108.8 & 0.906 \\
TA*+Smooth & 893 & 733 & 1660 & 1974 & 104.5 & 0.976 \\
\quad search only & 367 & 347 & 549 & 562 & -- & -- \\
\quad smooth only & 526 & 382 & 1342 & 1496 & -- & -- \\
\bottomrule
\end{tabular}%
}
\end{table}

\begin{table}[t]
\centering
\caption{Stage D: CARLA rendering throughput vs.\ worker count, measured in our benchmark run under the watchdog. Each worker renders one trajectory drawn from a common pilot pool into 1280$\times$720 RGB + depth + instance segmentation. ``Attempted'' includes partial output from failed workers; ``successful'' excludes them. ``Succ.\ FPS'' is the metric used for scaling analysis. ``Restarts'' counts CARLA server / client watchdog events.}
\label{tab:render_scaling}
\resizebox{\columnwidth}{!}{%
\begin{tabular}{lrrrrrrrrr}
\toprule
W & Wall (s) & Attempted & Successful & Succ.\ FPS & Speedup & GPU util.\ (\%) & GPU mem.\ (GiB) & Restarts & Failed \\
\midrule
1 & 717 & 770 & 770 & 1.07 & 1.00$\times$ & 46.3 & 7.5 & 0/0 & 0 \\
2 & 979 & 1404 & 1404 & 1.43 & 1.34$\times$ & 58.6 & 10.6 & 1/0 & 0 \\
4 & 1097 & 2792 & 2792 & 2.55 & 2.37$\times$ & 86.8 & 24.3 & 0/0 & 0 \\
6 & 1274 & 3650 & 3418 & 2.68 & 2.50$\times$ & 91.6 & 34.0 & 16/3 & 1 \\
\bottomrule
\end{tabular}%
}
\end{table}

\begin{table}[t]
\centering
\caption{Stage-level watchdog reliability. ``Attempts'' = initial launch + auto-restarts; ``restarts'' triggered by non-zero exit or heartbeat stall. Stages A--C time is the stage wall-clock; D outer is a dispatcher that delegates to workers (no own wall-time); inner-CARLA rows report the full render session wall-time.}
\label{tab:watchdog_reliability}
\small
\begin{tabular}{lrrrl}
\toprule
Stage & Attempts & Restarts & Time (s) & Failure modes \\
\midrule
A. Ped.\ planning & 1 & 0 & 3.2 & none \\
B. MuCO planning & 2 & 0 & 4.4 & none \\
C. TA*+Smooth & 1 & 0 & 17.9 & none \\
D. Render (dispatcher) & 1 & 0 & --- & none \\
D-inner: W=2 & 2 & 1 & 979.2 & CARLA health-check \\
D-inner: W=6 & 20 & 19 & 1274.3 & CARLA health-check, client retry failure \\
\bottomrule
\end{tabular}
\end{table}

\subsection{Stage B and Stage C: Drone Trajectory Planning on Two Optimizers}
\label{sec:dataset_planners}

\begin{figure}[t]
\centering
\includegraphics[width=0.95\columnwidth]{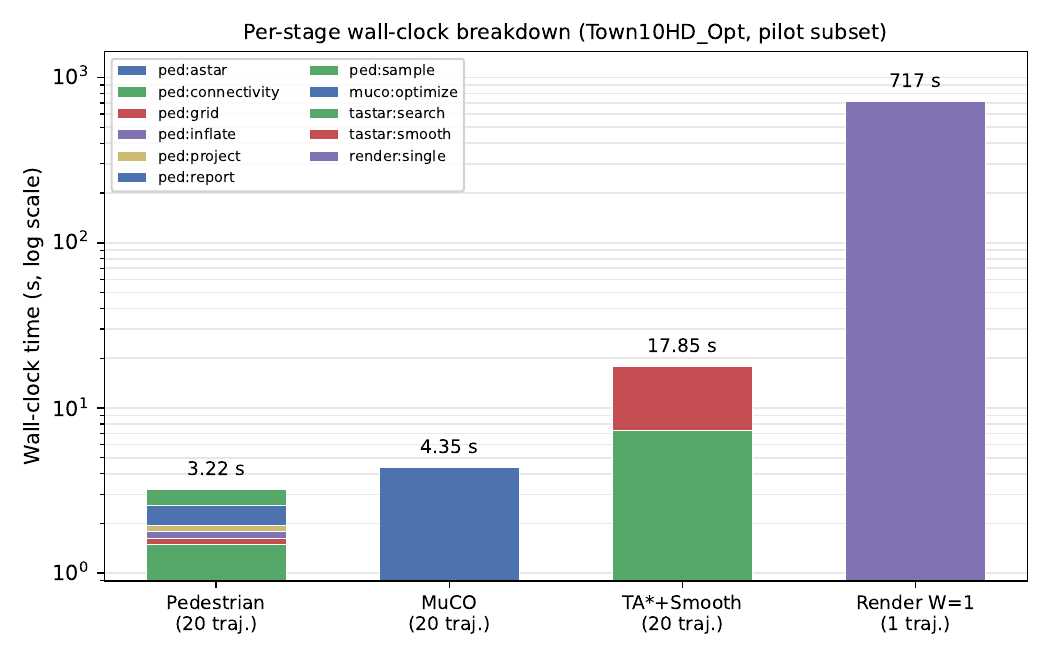}
\caption{Per-stage wall-clock budget on the 20-trajectory pilot (Town10HD\_Opt). Stage A (pedestrian, $\sim$3.2\,s for 20 trajectories), Stage B (MuCO, $\sim$4.4\,s for 20 trajectories), and Stage C (TA*+Smooth, $\sim$17.9\,s for 20 trajectories) are dwarfed by Stage D (single-process CARLA rendering, $717.4$\,s for one trajectory). The logarithmic ordinate makes the $\sim$200--500$\times$ per-trajectory gap visible and motivates the multi-process render benchmark of \cref{sec:dataset_multi}.}
\label{fig:fig17_stage_breakdown}
\end{figure}

Both drone planners consume the same scenario batch derived from Stage A. We benchmark them on the 20-trajectory pilot under the same single-knob $dt = 0.5$\,s convention as Sections \ref{sec:step4} and \ref{sec:exp_traj}, and report the per-trajectory wall-time distribution rather than a single mean. \Cref{tab:planner_dist} summarises the distribution and \cref{fig:fig17_stage_breakdown} visualises the full stage envelope on a logarithmic scale so that the four-orders-of-magnitude gap between planning and rendering remains legible. MuCO converges in $218$\,ms on average ($\sigma{=}42$\,ms, $p_{95} = 274$, max $282$\,ms) with a Rust batch of 32 worker threads, while TA*+Smooth averages $893$\,ms ($p_{95} = 1660$, max $1974$\,ms) split almost evenly between the search frontend ($367$\,ms mean) and the post-smoothing backend ($526$\,ms mean). The roughly $4\times$ wall-time penalty buys a $7.7$\,pp \emph{planner-internal 5-ray visibility} improvement (mean $0.976$ vs.\ $0.906$; this is \emph{not} the per-frame depth-buffer visibility reported in \cref{sec:dataset_stats}) and a $4$\,m shorter mean smoothed path ($\sim$3.9\% relative reduction). Note that \cref{sec:exp_traj} reports Path~0 \emph{full case-study pipeline} timings (TA*+Smooth $906$\,ms, MuCO $958$\,ms) which include orchestration, logging, and I/O overhead beyond the optimizer kernel. In the optimizer-only batch (\cref{tab:planner_dist}), Path~0 MuCO completes in $187$\,ms (below the batch median of $230$\,ms). Because the two measurements capture different scopes, \cref{tab:planner_dist} should not be compared numerically with \cref{sec:exp_traj}; the former isolates solver time, while the latter reports end-to-end single-path wall-time.

\subsection{Stage D, Single-Process Rendering}
\label{sec:dataset_single}

The single-process baseline launches one offscreen CARLA 0.9.16 server paired with the replay client at the release rendering preset (1280$\times$720, FOV $90^{\circ}$, dual augmentations per trajectory, depth + instance segmentation channels enabled), all monitored by the stage-level watchdog described in \cref{sec:dataset_watchdog}. On Path~0 from the pilot batch the session wall-time is $717.4$\,s ($\sim$12.0\,min), the pipeline emits $770$ PNG frames totalling $2.13$\,GiB, the mean GPU utilisation is $46.3\%$ and the dedicated GPU memory footprint is $7{,}674$\,MiB. No CARLA restart and no client retry was needed, giving a clean single-worker baseline of $1.07$\,fps that downstream rows are normalised against.

\subsection{Stage D, Multi-Process Rendering}
\label{sec:dataset_multi}

\begin{figure}[t]
\centering
\includegraphics[width=0.95\columnwidth]{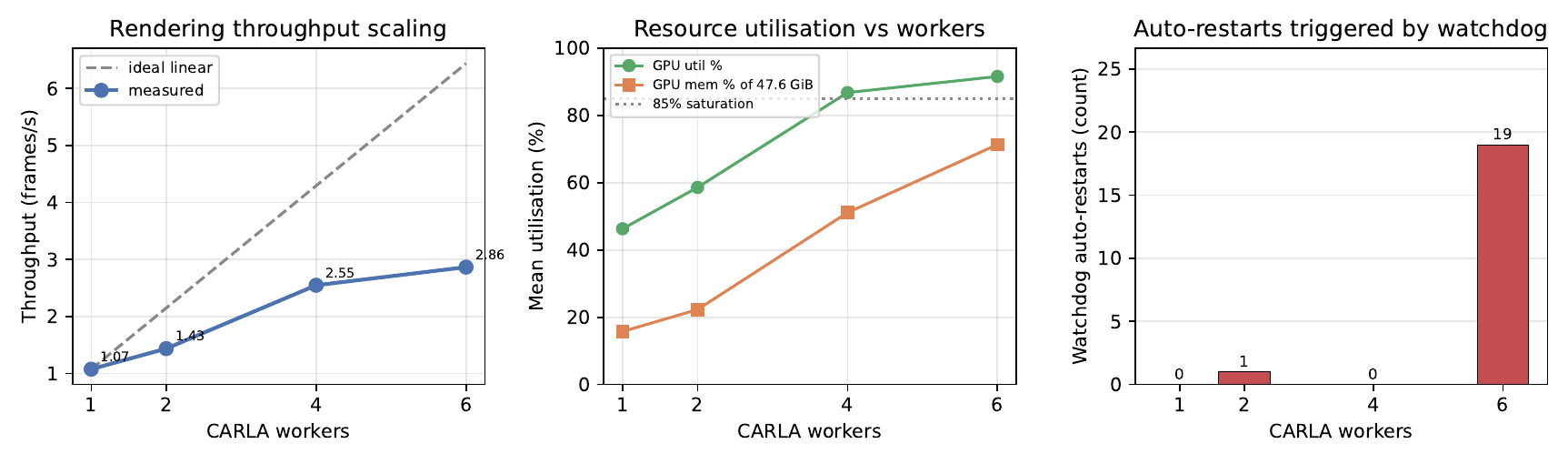}
\caption{Measured CARLA rendering at $W \in \{1, 2, 4, 6\}$ workers. Left: throughput in frames-per-second against the contention-free ``ideal linear'' upper bound; the gap visualises the cost of GPU and shader contention. Middle: mean GPU utilisation and mean GPU memory fraction of the 47.6\,GiB on-board budget; the dotted line marks the $85\%$ saturation threshold beyond which we start observing watchdog restart events. Right: total number of watchdog auto-restart events recorded for each worker count. At W=6 the mean GPU memory reaches $34.0$\,GiB (peak $40.6$\,GiB), approaching saturation and triggering 19 restart events under contention; W=4 is the largest configuration that completes with zero restarts in this pilot.}
\label{fig:fig18_render_scaling}
\end{figure}

We measure 2, 4, and 6 parallel CARLA configurations against the same baseline trajectories in our benchmark run. Each worker $i$ launches its own offscreen CARLA server on RPC port $2000 + 10 i$ (CARLA reserves three consecutive ports so a stride of 10 is conservative); the workers are then dispatched in parallel and watched by the same auto-restart logic that the single-process run used. The measured per-CARLA GPU footprint is $\sim$7.7\,GiB, which moves the effective ceiling on a single RTX 6000 Ada (48\,GiB) from the projected 7 workers down to a practical $\sim$6 workers before the watchdog starts firing. \Cref{tab:render_scaling} and \cref{fig:fig18_render_scaling} report the resulting scaling envelope. Throughput climbs from $1.07$\,fps (W=1) to $1.43$\,fps (W=2, $1.34\times$), $2.55$\,fps (W=4, $2.37\times$, $86.8\%$ mean GPU util, $24.3$\,GiB GPU memory), and $2.68$\,fps successful throughput (W=6, $2.50\times$, $91.6\%$ GPU util, $34.0$\,GiB mean / $40.6$\,GiB peak GPU memory). Beyond W=4 the GPU approaches saturation: at W=6 the watchdog records $16$ in-session CARLA server restarts and $3$ client retries, and one trajectory (worker 0) is dropped after exhausting all retries. The 6-worker configuration therefore delivers only a marginal $0.13$\,fps successful-throughput gain over 4 workers at the cost of dramatically worse reliability; on this pilot run, W=4 is the best observed trade-off for this workstation. We note that these are single-run observations (each worker count tested once); a definitive sweet-spot conclusion would require repeated trials under controlled conditions.

\subsection{Watchdog-Monitored Reliability}
\label{sec:dataset_watchdog}

\begin{figure}[t]
\centering
\includegraphics[width=0.95\columnwidth]{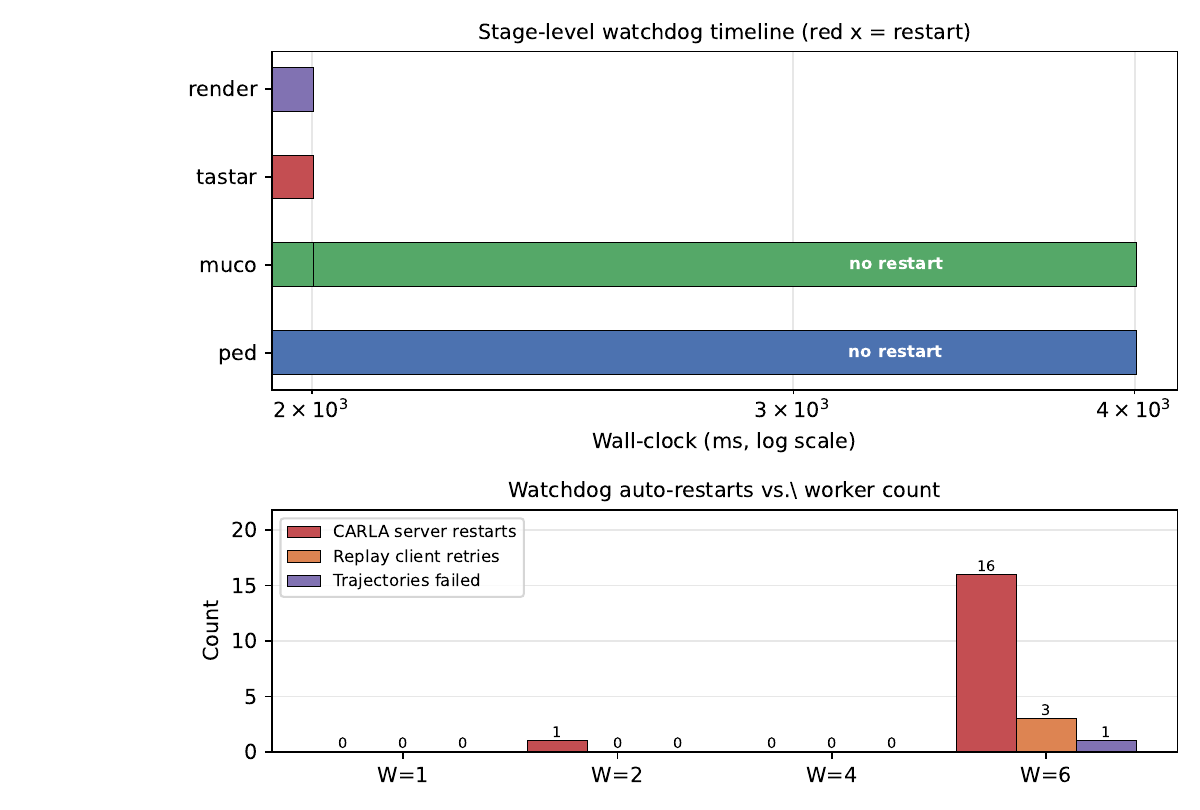}
\caption{Gantt-style watchdog timeline. Each horizontal bar represents one child-process attempt; bar length encodes wall-time on a logarithmic axis. Red crosses mark inner-CARLA auto-restart events (1 at W=2, 19 at W=6). Inline labels on the longer stage bars indicate whether a restart occurred; all four outer-wrapper stages completed without restart. The bottom panel shows per-configuration restart counts; all W=6 failures cluster on worker~0.}
\label{fig:fig19_watchdog}
\end{figure}

Every stage in \cref{tab:stage_breakdown,tab:planner_dist,tab:render_scaling} runs under a three-tier watchdog architecture. The \emph{outer stage wrapper} records each child-process launch as one attempt, restarts the child on a non-zero exit code or on a heartbeat-file stall (configurable per stage), and appends a structured event to the reproducibility log. Stage D additionally embeds two more layers inside each worker: (i)~a \emph{CARLA health-check} that pings the simulator RPC port every 5\,s and force-restarts the server after three consecutive failures, and (ii)~a \emph{client retry loop} that re-launches the replay client up to two more times before declaring the trajectory failed. \Cref{tab:watchdog_reliability} and \cref{fig:fig19_watchdog} report the resulting reliability envelope. The outer wrappers for Stages A--D produced zero auto-restarts; the inner CARLA health-check fired exactly once at W=2---the event log confirms a slow first server boot recovered automatically on the second attempt ($\sim$9\,s delay)---and a further 19 times at W=6, of which 16 were CARLA server restarts and 3 were client retries. All W=6 failures concentrated on worker~0; whether this reflects a port-assignment bias or a path-specific resource spike remains an open question that would require port-rotation or path-swap experiments to resolve. The 19 restart events kept the W=6 run alive long enough to complete 5 of 6 trajectories instead of dropping the whole batch. The failed trajectory (worker~0) produced $232$ partial frames; these partial frames are \emph{excluded} from the public dataset and do not contribute to the ``successful FPS'' metric in \cref{tab:render_scaling}.

\subsection{Dataset Statistics}
\label{sec:dataset_stats}

The first public release contains $250$ validated trajectories generated with MuCO, selected through a two-stage filtering funnel: (i)~a \emph{planner-stage prefilter} that requires MuCO planner convergence, a planner-internal 5-ray visibility $\geq 40\%$, and the smoothness envelope of \cref{sec:step6}; trajectories failing this stage are dropped before rendering. (ii)~A \emph{render-stage quality check} run on the surviving trajectories that screens for collisions, frame-completeness gaps, and semantic consistency anomalies. The rendered depth-buffer visibility characterised in \cref{tab:zoom_eval} is reported as a downstream descriptor of the released trajectories and is \emph{not} used as a rejection rule. On the Town10HD\_Opt pilot batches reported in this chapter, the observed ratio of initial candidates to validated trajectories is approximately $4{:}1$ to $5{:}1$. The same funnel is applied across all maps in the multi-map production pipeline; TA*+Smooth is evaluated only as a comparison baseline (\cref{sec:exp_traj}). Each trajectory contains approximately $618$--$770$ rendered frames (variation arises from differing pedestrian path lengths at the fixed $dt = 0.5$\,s / 2\,Hz sampling rate); for the release we subsample to a uniform $400$ frames per trajectory via stride selection, yielding about $100{,}000$ images total. Per-frame annotations include:
\begin{itemize}[nosep]
    \item RGB image (1280$\times$720 pixels)
    \item Metric depth map (32-bit float NumPy array)
    \item Semantic segmentation mask (CARLA built-in labels)
    \item Complete 6-DOF drone pose: position $(x, y, z)$ and orientation (yaw, pitch, roll)
    \item Target world position and per-frame rendered depth-buffer visibility flag / score (\cref{sec:step6})
    \item Natural language navigation instruction (Chinese and English)
\end{itemize}

The 3D environment statistics highlight the complexity of the urban scene: 65,614 raw bounding boxes (95.38\% vegetation) are simplified to 2,067 boxes, reflecting the dense obstacle environment in which pedestrian and drone trajectories must navigate. We use two distinct visibility quantities and make the distinction explicit here. The \emph{per-trajectory planner-internal 5-ray visibility}, measured on the 20-trajectory pilot subset by the 5-ray test used inside the planner, has mean $0.906$ and range $[0.641, 1.000]$ (\cref{tab:quality_stats}); it is the metric used by the planner-stage prefilter of \cref{sec:step6}, and every pilot trajectory in this paper sits above the $40\%$ prefilter threshold. The \emph{per-frame rendered depth-buffer visibility} aggregated over the four FOV configurations (\cref{tab:zoom_eval}) covers a wider range, from $52.1\%$ at the most occluded telephoto setting to $89.2\%$ at the wide-angle setting, providing a natural distribution of tracking difficulty levels at the frame level; it is a downstream descriptor and is not used as a rejection rule.

\FloatBarrier

\section{Baseline Experiments}
\label{sec:experiments}

This section consolidates three baseline experiments that share Town10HD\_Opt as the common map but differ in scope, sample size, and the supporting tables / figures from \cref{sec:dataset}: (1)~a trajectory-planning comparison that contrasts a Path~0 case study with a 20-trajectory optimizer-only distribution, (2)~a rendering-quality assessment that aggregates the dataset-level statistics already established in \cref{sec:dataset_stats}, and (3)~a zoom-capability evaluation on a separate 50-trajectory single-run pool. \Cref{tab:baseline_protocol} summarises the scope of each experiment so that downstream claims can be attributed to the right sample.

\begin{table}[t]
\centering
\footnotesize
\caption{Baseline experiment protocol (slim 4-column form). Every row in rows~1--5 uses Town10HD\_Opt; the public release row uses the multi-map pool. ``Single run'' denotes one execution under the listed configuration without independent repeats. Companion figures for each experiment (\cref{fig:fig17_stage_breakdown}, \cref{fig:fig18_render_scaling}, \cref{fig:fig19_watchdog}) are referenced from the prose rather than this table to keep the column readable.}
\label{tab:baseline_protocol}
\begin{enumerate}[leftmargin=*,itemsep=2pt,topsep=2pt]
\item \textbf{Path~0 case study:} $N{=}1$ on Town10HD\_Opt; released MuCO + TA*+Smooth; single-process, end-to-end wall-time. Headline metric: \cref{tab:algo_comparison}. Repetitions: single run.
\item \textbf{20-traj.\ planner pilot:} $N{=}20$ on Town10HD\_Opt; MuCO + TA*+Smooth optimizers; 32-thread Rust batch, optimizer-only. Headline metric: \cref{tab:planner_dist}. Repetitions: single run.
\item \textbf{Render scaling pilot:} MuCO release pool on Town10HD\_Opt at four worker counts ($W{=}1,2,4,6$); parallel CARLA, $1280{\times}720$, FOV $90^\circ$; per-$W$ frame counts in \cref{tab:render_scaling}. Headline metric: \cref{tab:render_scaling}. Repetitions: single per $W$.
\item \textbf{Watchdog reliability:} Same pool and worker counts as render scaling; three-tier watchdog with $2.0$\,s heartbeat. Headline metric: \cref{tab:watchdog_reliability}. Repetitions: single per $W$.
\item \textbf{Zoom evaluation:} $N{=}50$ on Town10HD\_Opt; MuCO release pool; fixed FOV per trajectory ($30^\circ$/$60^\circ$/$90^\circ$/$110^\circ$). Headline metric: \cref{tab:zoom_eval}. Repetitions: single run.
\item \textbf{Public release:} $N{=}250$ validated, multi-map; MuCO release funnel: planner-stage prefilter $\rightarrow$ rendering ($W{=}4$ recommended) $\rightarrow$ 400-frame stride subsample. Headline metric: \cref{sec:dataset_stats}. Repetitions: production.
\end{enumerate}
\end{table}

\subsection{Trajectory Planning Comparison}
\label{sec:exp_traj}

Trajectory planning is benchmarked at two complementary scopes. \Cref{tab:algo_comparison} reports an \emph{end-to-end Path~0 case study} on Town10HD\_Opt: the planner kernel together with orchestration, logging, and I/O overhead is timed as a single wall-clock measurement. \Cref{tab:planner_dist} reports the \emph{optimizer-only distribution} over the 20-trajectory pilot subset of \cref{sec:dataset_planners}, with the optimizer kernel isolated from the orchestration overhead and dispatched through the 32-thread Rust batch harness. The two measurements should not be averaged together: the Path~0 row in \cref{tab:planner_dist} (MuCO $187$\,ms optimizer-only, below the batch median of $230$\,ms) and the Path~0 row in \cref{tab:algo_comparison} (MuCO $958$\,ms end-to-end) capture different scopes of the same trajectory.

\paragraph{Unified sampling rate.}
A key design decision is to expose a single $dt$ parameter that governs the variable-speed pedestrian resampling of Step~3 and the drone trajectory in Step~4 \emph{simultaneously}, so that the two planners always produce drone trajectories temporally synchronized with the ground-level target without asking the user to align two independent time steps. The two paradigms still differ substantially in path length and runtime as reported in \cref{tab:algo_comparison,tab:planner_dist}, but neither produces drift relative to the pedestrian. The ``building-circling'' artifact described in \cref{sec:step4} is unrelated to $dt$: it is a visibility-gradient cycling issue and its dedicated fixes are documented in that section.

\paragraph{Planner roles and trade-off.}
We adopt MuCO as the \emph{production planner} for the public release on the basis of the optimizer-only batch in \cref{tab:planner_dist} ($218$\,ms mean vs.\ $893$\,ms mean, an approximately $4\times$ wall-time advantage), at the cost of an approximately $3.9\%$ longer mean smoothed path and a $7.0$\,pp lower planner-internal 5-ray visibility ($0.906$ vs.\ $0.976$). This adoption criterion uses optimizer-only timing, not the Path~0 end-to-end case study (where MuCO's $958$\,ms is in fact slightly slower than TA*+Smooth's $906$\,ms, \cref{tab:algo_comparison}); end-to-end timing is scope-dependent and dominated by orchestration / I/O overhead rather than solver cost. TA*+Smooth is retained as a \emph{comparison baseline} so that downstream users can swap planners under the same single-knob $dt$ convention. On the 20-trajectory pilot, the MuCO release pilot passes the planner-stage prefilter of \cref{sec:step6} with zero rejections (\cref{tab:quality_stats}); the same prefilter applied to TA*+Smooth outputs is not reported in this paper.

\paragraph{Planning vs.\ rendering budget.}
\Cref{fig:fig17_stage_breakdown} visualises the stage-wise wall-time envelope on the 20-trajectory pilot: Stage A pedestrian planning ($\sim$$3.2$\,s for 20 trajectories), Stage B MuCO planning ($\sim$$4.4$\,s), and Stage C TA*+Smooth planning ($\sim$$17.9$\,s) are dwarfed by Stage D single-process CARLA rendering ($717.4$\,s for one trajectory; \cref{sec:dataset_single}). On this single-machine pilot, planner time is therefore not the bottleneck relative to rendering; we do not extrapolate this observation to a guarantee at thousands of trajectories, because render scaling beyond $W=4$ degrades reliability under contention (\cref{tab:render_scaling}).

\subsection{Rendering Quality and Data Fidelity}
\label{sec:exp_render}

\paragraph{Synchronized multi-modal sample.}
\Cref{fig:dataset_sample} shows one synchronized release sample: RGB, the float32 depth array visualised through the \texttt{inferno\_r} colormap, semantic segmentation, and a 3D-box reprojection debug view all share the same simulator tick (\cref{sec:step5}). The pipeline writes one such triplet per rendered frame and emits per-frame JSON annotations carrying the intrinsic matrix and the 6-DOF drone pose; this provides synchronized RGB / depth / segmentation samples suitable for downstream VLM experiments rather than a quality claim specific to any one downstream task.

\paragraph{Render configuration.}
All release captures use a fixed resolution of $1280{\times}720$, manual exposure (auto-exposure disabled), motion blur disabled, and one of the four fixed FOV values listed in \cref{tab:zoom_config}; the three sensors are synchronized on a shared simulator tick as described in \cref{sec:step5}. Each trajectory is rendered with two augmentation passes (a clean track and a perturbed track) as detailed in \cref{app:dual_track}.

\paragraph{Depth precision.}
Storing depth as 32-bit floating-point NumPy arrays preserves full metric precision; quantization losses are characterised quantitatively in Appendix~A. On the representative $720{\times}1280$ scene with a valid depth range of $15.36$--$38.32$\,m used in \cref{tab:depth_format_comparison}, 8-bit storage yields a mean error of $2.25$\,cm, a maximum error of $4.50$\,cm, and visible banding artefacts on continuous surfaces (\cref{fig:depth_comparison}), while 32-bit float yields a mean error of $1.0\times10^{-4}$\,cm and preserves edge correlation of $1.000$ against the 64-bit ground truth. The valid-range and unit conventions used here are inherited from Appendix~A; no separate $0.5$--$80$\,m, $0.15$\,m / $0.31$\,m figures are claimed in this paper.

\paragraph{Perturbation diversity.}
Random perturbation parameters introduce controlled per-frame variation in drone and pedestrian poses; each frame falls back to the unperturbed pose whenever the perturbed pose would push the target out of the frustum or violate the safety clearance (\cref{app:dual_track}). On the 20-trajectory pilot subset (\cref{tab:quality_stats}), $20/20$ trajectories pass the per-trajectory $40\%$ visibility threshold and the smoothness envelope ($a_{\mathrm{rms}} \le 5\,\mathrm{m/s}^{2}$, $j_{\mathrm{rms}} \le 10\,\mathrm{m/s}^{3}$); per-frame fallback-to-unperturbed events are logged inside the released annotation but are not aggregated in this paper.

\paragraph{Temporal density.}
Each raw trajectory yields $618$--$770$ rendered frames at $dt = 0.5$\,s (2\,Hz), with the spread driven by pedestrian path length. For the public release each trajectory is \emph{subsampled} to $400$ frames via uniform stride selection so that all trajectories share the same temporal envelope; the released $250$-trajectory pool therefore contains approximately $100{,}000$ rendered images.

\paragraph{Pipeline reliability.}
The reliability envelope is governed by the three-tier watchdog described in \cref{sec:dataset_watchdog} and is configuration-dependent rather than a blanket end-to-end guarantee. On the render-scaling pilot (\cref{tab:render_scaling,tab:watchdog_reliability,fig:fig19_watchdog}), all four outer-stage wrappers produced zero auto-restarts; the inner CARLA health-check fired exactly once at $W=2$ (a slow first server boot that recovered after $\sim$$9$\,s) and $19$ times at $W=6$, where one trajectory on worker~0 was dropped after exhausting all retries. The conservative reliability statement supported by this pilot is therefore ``zero outer-stage restarts and zero failed trajectories at $W \le 4$''; we do not claim a wider end-to-end guarantee.

\subsection{Zoom Capability Evaluation}
\label{sec:exp_zoom}

\paragraph{Measurement protocol.}
We evaluate the four fixed FOV configurations of \cref{tab:zoom_config} on $50$ test trajectories drawn from the MuCO release pool on Town10HD\_Opt, with each trajectory held at a constant FOV for its entire duration (matching the per-trajectory FOV convention of \cref{sec:step5}). All other render settings are held constant ($1280{\times}720$, fixed manual exposure, motion blur disabled). The reported numbers are averaged over the visible frames of each trajectory and then over the $50$ trajectories; the evaluation is a single-run pool with no independent repetitions.

\paragraph{Metric definitions.}
For this evaluation we use three quantities aggregated to per-trajectory means and then to a 50-trajectory pool mean. \emph{Target Size (px)} is the width$\times$height bounding box of the projected 3D target box in image pixels. \emph{Visibility (\%)} is the rendered depth-buffer visibility ratio defined in Step~6 (\cref{sec:step6})---the per-frame fraction of the projected target 3D box that is unoccluded under the depth-buffer test, cross-checked against the segmentation pedestrian mask. This metric is distinct from the planner-internal 5-ray visibility used in \cref{tab:algo_comparison,tab:planner_dist,tab:dist_stats}. \emph{Track Quality} is the per-frame bounding-box IoU between (a)~the projected 3D target box and (b)~the axis-aligned bounding box of the connected pedestrian-labelled region in the segmentation mask, averaged only over frames where the rendered depth-buffer visibility is strictly positive. Trajectories with no visible frames in the entire sequence would be excluded by this rule; in the 50-trajectory pool no such trajectory occurred.

\begin{table}[t]
\centering
\caption{Zoom configuration evaluation on $50$ single-run Town10HD\_Opt test trajectories. ``Visibility'' is the per-frame depth-buffer ratio of \cref{sec:step6}; ``Track Quality'' is the bounding-box IoU defined in \cref{sec:exp_zoom}. Values are pool-level means over the 50 trajectories; no independent repetitions and no standard-deviation columns are available for this pilot.}
\label{tab:zoom_eval}
\begin{tabular}{lrrrr}
\toprule
Zoom Level & FOV ($^\circ$) & Target Size (px) & Visibility (\%) & Track Quality (IoU) \\
\midrule
Wide-angle & 110 & $12 \times 24$ & 89.2 & 0.72 \\
Standard   &  90 & $18 \times 36$ & 85.7 & 0.81 \\
Narrow     &  60 & $32 \times 64$ & 71.3 & 0.88 \\
Telephoto  &  30 & $68 \times 136$ & 52.1 & 0.93 \\
\bottomrule
\end{tabular}
\end{table}

\paragraph{Visibility--resolution trade-off.}
\Cref{tab:zoom_eval} exposes a monotone visibility--resolution trade-off on this single-run pool: wide-angle maximises per-frame visibility ($89.2\%$) at the cost of small target representation, while telephoto produces large targets and the highest track-quality IoU ($0.93$) but reduces visibility to $52.1\%$. We do not interpret this as a universal best-practice recommendation: the result is single-run, single-map, and reports pool means without confidence intervals; a definitive ranking would require repeated trials and additional maps.

\paragraph{Optical-like FOV zoom vs.\ digital zoom.}
The public release implements an \emph{optical-like FOV zoom}: the camera intrinsic is changed by selecting one of the four fixed FOV values, with the FOV held constant for the duration of each trajectory. This is implemented in CARLA as a camera-intrinsic adjustment, not as a physical focal-length change. A qualitative $5\times$ comparison against post-hoc digital zoom (centre-crop with bilinear / bicubic interpolation) on the same scene is shown in \cref{fig:zoom_fidelity} (Appendix~B); digital zoom introduces visible blurring and loss of high-frequency texture, while the FOV-based zoom preserves sharp edges. The dynamic FOV-via-actor-recreation procedure documented in \cref{app:zoom} is an optional implementation route for PTZ-style applications and is not used inside the public trajectories.

\subsection{Release-Level Summary}
\label{sec:exp_release}

\Cref{sec:dataset_stats} reports that the first public release contains $250$ validated MuCO trajectories produced by the two-stage filtering funnel (planner-internal 5-ray visibility prefilter $\geq 40\%$ together with the smoothness envelope $\rightarrow$ rendering $\rightarrow$ render-stage QC for collisions, frame-completeness gaps, and semantic anomalies), with an observed candidate-to-release ratio of approximately $4{:}1$ to $5{:}1$ on the Town10HD\_Opt pilot batches reported in this chapter; the corresponding ratios on the multi-map production batches, together with the aggregate render-stage QC drop counts, are not tabulated in the main text and are instead recorded as release-artifact metadata. Each release trajectory is subsampled to $400$ frames, yielding approximately $100{,}000$ RGB / depth / segmentation triplets; the four FOV configurations of \cref{tab:zoom_config} are drawn per trajectory and the per-FOV trajectory counts are likewise recorded in the release-artifact metadata (we do not claim uniform distribution across FOVs in this paper).

Each frame in the release carries the synchronized triplet of \cref{fig:dataset_sample}, the per-frame 6-DOF drone pose, the target world position together with the per-frame rendered depth-buffer visibility flag / score (\cref{sec:step6}), the per-frame perturbation fallback flag exposed in the annotation JSON, and the bilingual CoC caption produced by the Stage~7 student model (\cref{sec:step7}). The pilot-level baselines reported in this section characterise components of the release-level pipeline on Town10HD\_Opt only; per-map production-funnel statistics and aggregate render-stage QC counts across the multi-map release are deferred to the public release artifact.

\FloatBarrier

\section{Limitations and Conclusion}
\label{sec:conclusion}

\subsection{Limitations}
\label{sec:limitations}

We acknowledge several inherent limitations of the pipeline design.

\paragraph{Simulation-to-real gap.}
While CARLA provides realistic urban rendering, the generated data may not fully capture the visual complexity, sensor noise, and environmental variability of real-world drone footage. Domain adaptation or fine-tuning on real-world data may be necessary for deployment scenarios. The pipeline's reliance on synthetic environments means that certain real-world phenomena (e.g., motion blur, lens distortion, atmospheric effects) are not fully represented.

\paragraph{Fixed pedestrian dynamics.}
The pipeline generates pedestrian trajectories using A* path planning with speed variations based on path curvature. This approach does not capture the full complexity of real human movement patterns, including sudden stops, direction changes, or interactions with other pedestrians. More sophisticated pedestrian behavior models could enhance trajectory realism.

\paragraph{Language limitation.}
The current pipeline supports only Chinese and English captions, rather than broader multilingual coverage. Although these two languages already enable a range of training and evaluation settings, extending the pipeline to additional languages remains important for wider international applicability.

\paragraph{Viewpoint limitation.}
The current pipeline focuses exclusively on UAV aerial viewpoints for outdoor pedestrian tracking. It does not yet include complementary perspectives such as pedestrian-level views, vehicle-mounted views, traffic surveillance views, or indoor viewpoints, which could further enrich cross-view perception and multi-agent understanding across both outdoor and indoor environments.

\paragraph{Ethical considerations.}
Aerial tracking of pedestrians raises privacy concerns even in simulated environments. The \cosfly-Track dataset is designed exclusively for research purposes, and we provide usage guidelines emphasizing responsible application. The simulated nature of the data mitigates direct privacy risks, but researchers should be mindful of downstream applications.

\subsection{Conclusion}
\label{sec:conclusion_final}

We have presented \cosfly, a generalizable construction pipeline for aerial tracking built on the CARLA simulator, together with the \cosfly-Track dataset, a large-scale multi-modal aerial tracking benchmark. Our contributions include: (1) the \cosfly-Track dataset comprising 250 validated public trajectories and approximately 100,000 rendered images with complete 6-DOF pose annotations, RGB images, high-precision depth maps, semantic segmentation, and natural language navigation instructions; (2) a modular, reproducible 7-step pipeline covering the complete workflow from map export to caption generation; (3) support for configurable fixed-FOV zoom levels via per-trajectory camera-intrinsic adjustments; (4) a trajectory-planning analysis contrasting two-stage frontend/backend planning with direct multi-constraint gradient planning; and (5) baseline experiments establishing community benchmarks for aerial tracking dataset construction.

The pipeline demonstrates that simulation-based generation can produce diverse, richly annotated aerial tracking data at a fraction of the cost of real-world collection. The paired natural language instructions enable VLM-based drone navigation research, addressing a critical gap in existing aerial datasets.

Future work includes expanding to additional CARLA maps, enriching annotation modalities to better serve downstream VLM-based drone navigation, and conducting simulation-to-real transfer experiments. To further enhance the realism and diversity of synthetic data, we plan to explore two complementary directions: (1) incorporating 3D Gaussian Splatting as a photorealistic rendering backend, which enables real-time novel-view synthesis and intuitive scene-level augmentation while substantially narrowing the sim-to-real visual gap; and (2) integrating generative world models tailored for low-altitude environments, which can synthesize long-tail scenarios and predict long-horizon visual observations beyond the coverage of any fixed simulator map. The combination of geometric reconstruction and generative imagination offers a promising path toward scalable, high-fidelity aerial data synthesis. More broadly, the modular design of \cosfly makes it applicable beyond aerial tracking dataset construction: it can support data generation for low-altitude embodied intelligence, cross-view target tracking, and a wider range of robotic perception, planning, and tracking tasks across diverse simulation backends and environments. We invite the community to build on this foundation for next-generation embodied robotics research and deployment.

\FloatBarrier

\bibliographystyle{plain}
\bibliography{refs}

\clearpage
\appendix

\section{Justification for 32-bit Float Depth Map Storage}
\label{app:depth}

This section provides a comprehensive quantitative and qualitative analysis to justify the use of the 32-bit floating-point (\texttt{float32}) format for storing and processing depth maps in our experiments. To avoid circular reasoning, we establish a theoretical lossless ground truth (GT) using a 64-bit double-precision (\texttt{float64}) format, which incorporates sub-millimeter precision information. We systematically evaluate various bit-depth formats (8-bit, 10-bit, 12-bit, 16-bit, and 32-bit) against this 64-bit GT across multiple dimensions, including quantization error, information entropy, edge preservation, and storage efficiency.

\subsection{Community Practices for Depth Map Storage}

The choice of depth storage format has been shaped by the evolution of depth estimation benchmarks. As summarized in \Cref{tab:dataset_format_survey}, 16-bit integer and 32-bit floating-point formats dominate the research landscape, while no major benchmark adopts 8-bit or 64-bit formats for ground truth depth storage.

Among 16-bit benchmarks, KITTI~\citep{Geiger2012CVPR} and NYU Depth V2~\citep{Silberman2012ECCV} pioneered large-scale depth evaluation, storing depth as \texttt{uint16} PNG with application-specific scaling (e.g., \texttt{depth\_m = pixel / 256.0} in KITTI). Subsequent indoor datasets---ScanNet~\citep{Dai2017CVPR} and ScanNet++~\citep{Yeshwanth2023ICCV}---adopted millimeter-scale \texttt{uint16} encoding. More recent datasets such as WildRGB-D~\citep{Xia2024CVPR} and Aria Synthetic Environments~\citep{Aria2024ASE} continue this convention for in-the-wild and synthetic indoor scenes, respectively.

In contrast, benchmarks demanding sub-pixel or high-precision depth uniformly adopt 32-bit float storage. Early datasets such as MPI Sintel~\citep{Butler2012ECCV} and Middlebury 2014~\citep{Scharstein2014GCPR} use PFM or binary float formats for disparity. Later synthetic datasets---TartanAir~\citep{Wang2020IROS}, Hypersim~\citep{Roberts2021ICCV}, Spring~\citep{Mehl2023CVPR}, and MatrixCity~\citep{Li2023ICCV}---store metric depth as 32-bit floats via \texttt{.npy}, HDF5, or OpenEXR containers. This practice reflects the community consensus that 32-bit float preserves the continuous nature of depth without introducing discretization artifacts.

Notably, all recent foundation models for depth estimation adopt 32-bit float as their native output format to avoid banding artifacts and preserve continuous gradients. Depth Anything V2~\citep{Yang2024NeurIPS} and Marigold~\citep{Ke2024CVPR} (2024) output raw float32 depth maps, while the latest video diffusion models---DepthCrafter~\citep{Hu2025CVPR} and Video Depth Anything~\citep{Chen2025CVPR} (2025)---similarly produce 32-bit float predictions. The convergence of both dataset curation and model inference on the 32-bit float format further validates our design choice. \Cref{tab:dataset_format_survey} provides a comprehensive overview of representative datasets and models organized by their depth storage format.

\subsection{Quantization Error and Information Loss}
\label{sec:appendix_quant_error}

Depth estimation in complex environments (e.g., autonomous driving or large indoor scenes) requires capturing both vast distance ranges and fine-grained local geometric variations. The choice of bit depth fundamentally limits the depth resolution. As summarized in \Cref{tab:depth_format_comparison}, lower bit-depth formats introduce severe quantization errors.

\begin{table}[H]
\centering
\caption{Quantitative comparison of depth map storage formats on a representative $720 \times 1280$ scene with a valid depth range of 15.36--38.32\,m. All metrics are computed against the 64-bit float ground truth.}
\label{tab:depth_format_comparison}
\resizebox{\linewidth}{!}{%
\begin{tabular}{lccccccccc}
\toprule
\textbf{Format} & \textbf{Unique} & \textbf{Max Err.} & \textbf{Mean Err.} & \textbf{RMSE} & \textbf{PSNR} & \textbf{Entropy} & \textbf{Edge} & \textbf{Storage} \\
 & \textbf{Values} & \textbf{(cm)} & \textbf{(cm)} & \textbf{(cm)} & \textbf{(dB)} & \textbf{(bits)} & \textbf{Corr.} & \textbf{(KB)} \\
\midrule
8-bit           & 252       & 4.5019 & 2.2494 & 2.5979 & 58.9  & 7.45  & 0.999980 & 900   \\
10-bit          & 988       & 1.1222 & 0.5543 & 0.6418 & 71.1  & 9.40  & 0.999999 & 1,125 \\
12-bit          & 3,824     & 0.2803 & 0.1436 & 0.1646 & 82.9  & 10.75 & 1.000000 & 1,350 \\
16-bit          & 48,048    & 0.0175 & 0.0088 & 0.0101 & 107.1 & 12.71 & 1.000000 & 1,800 \\
\textbf{32-bit float} & \textbf{457,274} & \textbf{0.0002} & \textbf{0.0001} & \textbf{0.0001} & \textbf{151.5} & \textbf{12.71} & \textbf{1.000000} & \textbf{3,600} \\
64-bit float (GT) & 921,598 & 0 & 0 & 0 & $\infty$ & 12.71 & 1.000000 & 7,200 \\
\bottomrule
\end{tabular}%
}
\end{table}

For a scene with a valid depth range of approximately 23\,m, an 8-bit format (256 discrete levels) yields a depth resolution of only 9.00\,cm per level, resulting in a maximum quantization error of 4.50\,cm and a Root Mean Square Error (RMSE) of 2.60\,cm. While a 16-bit format improves the RMSE to 0.01\,cm, it still discretizes the continuous physical world into 65,536 levels. In contrast, the \texttt{float32} format captures 457,274 unique depth values in a single frame, achieving an RMSE of $6.1 \times 10^{-5}$\,cm (0.61\,$\mu$m) compared to the 64-bit GT. This sub-micrometer error is negligible for all practical 3D perception tasks, yielding a Peak Signal-to-Noise Ratio (PSNR) of 151.5\,dB.

\subsection{Banding Artifacts and Structural Degradation}
\label{sec:appendix_banding}

\begin{figure}[t]
\centering
\includegraphics[width=0.88\columnwidth]{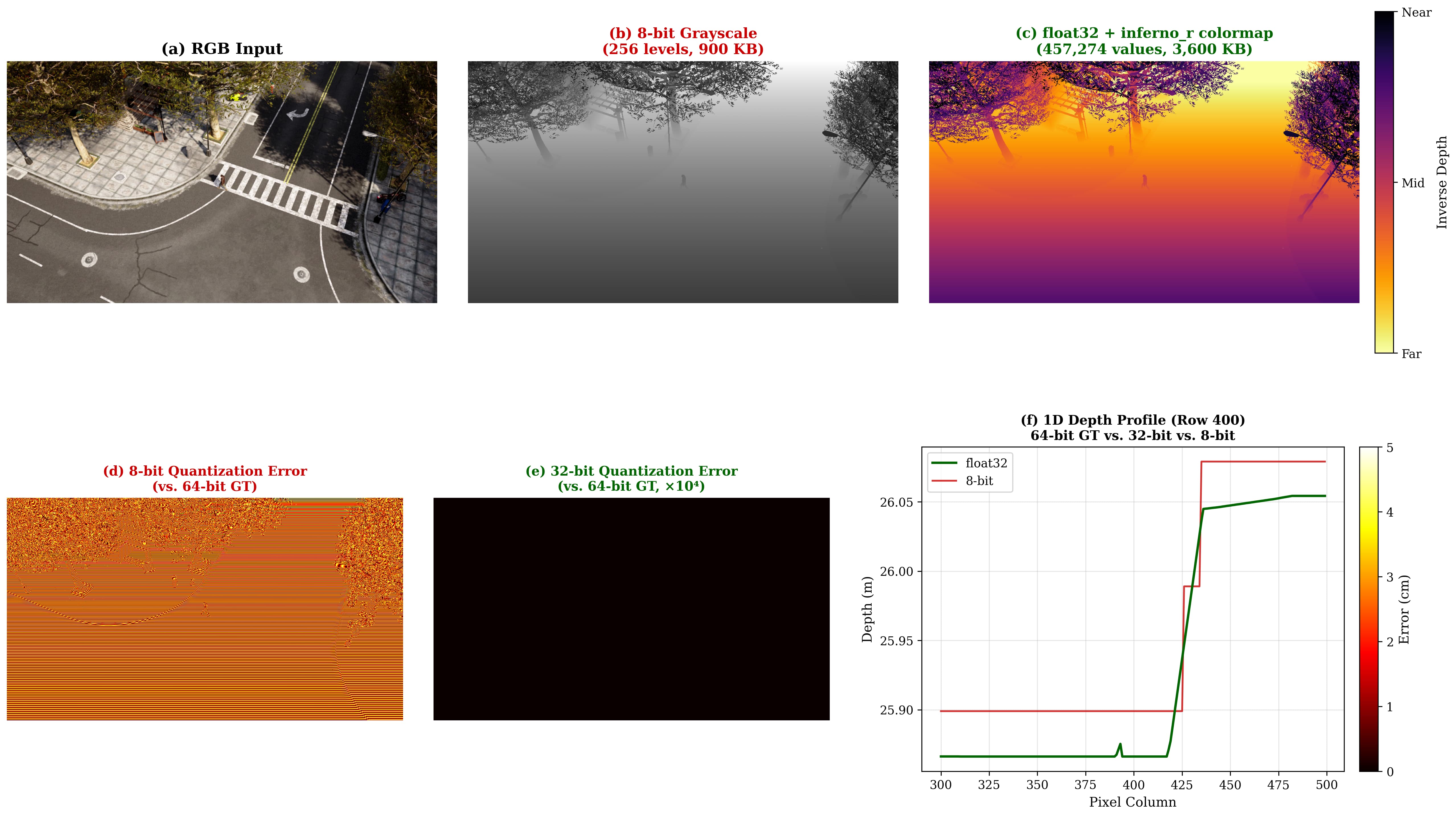}
\caption{Visual comparison of depth map storage and visualization methods. (a) RGB input. (b) 8-bit grayscale rendering. (c) float32 with inferno\_r colormap. (d) 8-bit quantization error map (vs.\ 64-bit GT). (e) 32-bit quantization error map (vs.\ 64-bit GT, $\times 10^4$). (f) 1D depth profile comparing 64-bit GT, 32-bit float, and 8-bit.}
\label{fig:depth_comparison}
\end{figure}

The most severe consequence of low bit-depth storage is the introduction of ``banding'' or ``staircase'' artifacts. As illustrated in \Cref{fig:depth_comparison}, continuous surfaces such as roads or walls are discretized into step-like structures when quantized to 8-bit or even 10-bit formats. These artifacts destroy local surface normals and edge gradients, which are critical features for point cloud reconstruction, obstacle avoidance, and 3D bounding box regression.

The edge preservation score, measured by the Pearson correlation coefficient of Sobel gradients between the quantized map and the 64-bit GT, drops to 0.999980 in 8-bit. The \texttt{float32} format, however, maintains a perfect edge correlation of 1.000000, perfectly preserving the structural integrity of the scene.

\subsection{Storage--Accuracy Pareto Optimality}
\label{sec:appendix_pareto}

While higher precision inherently requires more storage, the \texttt{float32} format represents the optimal ``sweet spot'' on the storage--accuracy Pareto front. As shown in \Cref{tab:depth_format_comparison}, the 64-bit GT requires 7,200\,KB per uncompressed $720 \times 1280$ frame. The \texttt{float32} format halves this storage requirement to 3,600\,KB while retaining 100.0\% of the information entropy (12.71\,bits) present in the 64-bit GT.

Further reducing the bit depth to 16-bit saves an additional 1,800\,KB but introduces measurable quantization errors (RMSE 0.01\,cm) and reduces the unique value count by an order of magnitude. Given the critical importance of geometric fidelity in our experiments, the storage footprint of \texttt{float32} is a necessary and highly acceptable cost to maintain near-zero quantization error and preserve full information entropy.

\begin{table*}[!t]
\centering
\caption{Representative depth estimation datasets and state-of-the-art (SOTA) models categorized by their ground truth depth map storage and output formats. The table demonstrates that 16-bit and 32-bit float are the dominant formats in the research community. Notably, all recent foundation models (2024--2025) output 32-bit float depth maps to preserve continuous gradients and avoid banding artifacts.}
\label{tab:dataset_format_survey}
\resizebox{\textwidth}{!}{%
\begin{tabular}{llllll}
\toprule
\textbf{Bit Depth} & \textbf{Dataset / Model} & \textbf{Year} & \textbf{File Format} & \textbf{Encoding Details} & \textbf{Domain} \\
\midrule
\multirow{6}{*}{16-bit uint}
 & KITTI~\citep{Geiger2012CVPR}            & 2012 & 16-bit PNG   & \texttt{depth\_m = pixel / 256.0}     & Autonomous driving \\
 & NYU Depth V2~\citep{Silberman2012ECCV}   & 2012 & 16-bit PNG / MAT & Kinect raw: mm; processed: HDF5  & Indoor \\
 & ScanNet~\citep{Dai2017CVPR}              & 2017 & 16-bit PNG   & \texttt{depth\_m = pixel / 1000.0}    & Indoor (RGB-D) \\
 & ScanNet++~\citep{Yeshwanth2023ICCV}      & 2023 & 16-bit PNG   & \texttt{depth\_m = pixel / 1000.0}    & Indoor (high-res) \\
 & WildRGB-D~\citep{Xia2024CVPR}            & 2024 & 16-bit PNG   & \texttt{depth\_m = pixel / 1000.0}    & In-the-wild objects \\
 & Aria Synthetic Env.~\citep{Aria2024ASE}  & 2024 & 16-bit PNG   & \texttt{depth\_m = pixel / 1000.0}    & Synthetic indoor \\
\midrule
\multirow{12}{*}{32-bit float}
 & \multicolumn{5}{l}{\textbf{Datasets}} \\
 & MPI Sintel~\citep{Butler2012ECCV}        & 2012 & Custom \texttt{.dpt} & Binary float32, depth in meters  & Synthetic (movie) \\
 & Middlebury 2014~\citep{Scharstein2014GCPR} & 2014 & PFM          & Float32 disparity in pixels           & Indoor (stereo) \\
 & TartanAir~\citep{Wang2020IROS}           & 2020 & \texttt{.npy} (NumPy) & Float32, depth in meters        & Synthetic (diverse) \\
 & Hypersim~\citep{Roberts2021ICCV}         & 2021 & HDF5         & Float32/16 channels, depth in meters  & Synthetic (indoor) \\
 & Spring~\citep{Mehl2023CVPR}              & 2023 & HDF5         & Float32 disparity in pixels           & High-res stereo/flow \\
 & MatrixCity~\citep{Li2023ICCV}            & 2023 & OpenEXR      & Float32 depth in cm                   & City-scale synthetic \\
 \cmidrule{2-6}
 & \multicolumn{5}{l}{\textbf{SOTA Models (Output Format)}} \\
 & Depth Anything V2~\citep{Yang2024NeurIPS} & 2024 & \texttt{.npy} (NumPy) & Float32 raw depth map           & Foundation model \\
 & Marigold~\citep{Ke2024CVPR}              & 2024 & \texttt{.npy} (NumPy) & Float32 affine-invariant depth  & Diffusion-based \\
 & DepthCrafter~\citep{Hu2025CVPR}          & 2025 & \texttt{.npy} (NumPy) & Float32 normalized disparity    & Video diffusion \\
 & Video Depth Anything~\citep{Chen2025CVPR} & 2025 & \texttt{.npy} (NumPy) & Float32 depth map               & Video foundation \\
\bottomrule
\end{tabular}%
}
\end{table*}

\FloatBarrier

\section{Zoom Capability Implementation}
\label{app:zoom}

The CARLA simulator abstracts camera intrinsics through a horizontal field-of-view (FOV) parameter rather than providing direct control over the physical focal length (expressed in millimeters). Consequently, optical zoom is realized indirectly by varying the FOV setting. This section formalizes the mapping between FOV and focal length and provides practical reference configurations.

\subsection{FOV to Focal Length Conversion}

The equivalent focal length in pixels can be computed from FOV using:
\begin{equation}
f_{\text{pixels}} = \frac{W}{2 \cdot \tan(\text{FOV} \cdot \pi / 360)}
\end{equation}
where $W$ is image width (1280 pixels in our configuration) and FOV is in degrees. \Cref{tab:zoom_conversion} provides reference values.

\begin{table}[tb]
\centering
\caption{FOV to focal length conversion for 1280$\times$720 images.}
\label{tab:zoom_conversion}
\begin{tabular}{lrrr}
\toprule
Lens Type & FOV ($^\circ$) & $f$ (pixels) & Equiv.\ 35mm (mm) \\
\midrule
Ultra-wide & 120 & 369 & 14 \\
Wide & 110 & 485 & 18 \\
Standard & 90 & 640 & 24 \\
Normal & 70 & 918 & 35 \\
Narrow & 60 & 1109 & 43 \\
Telephoto & 45 & 1546 & 58 \\
Long telephoto & 30 & 2391 & 90 \\
\bottomrule
\end{tabular}
\end{table}

\subsection{Optional Dynamic Zoom via Actor Recreation}

CARLA does not support runtime FOV modification. To simulate dynamic zoom (e.g., PTZ cameras), the pipeline can optionally destroy and recreate the camera actor with updated FOV settings between segments. This optional route introduces brief discontinuities at each FOV change; it is not used inside the public \cosfly-Track trajectories (which hold the per-trajectory FOV constant per \cref{sec:step5}) but is available to downstream users that explicitly need PTZ-style behaviour.

\subsection{Camera Intrinsic Matrix}

For 3D-2D projection or camera calibration tasks, the intrinsic matrix $K$ can be constructed as:
\begin{equation}
K = \begin{bmatrix} f_x & 0 & c_x \\ 0 & f_y & c_y \\ 0 & 0 & 1 \end{bmatrix}
\end{equation}
where $f_x = f_y = f_{\text{pixels}}$ (square pixels), $c_x = W/2 = 640$, and $c_y = H/2 = 360$.

\subsection{Visual Demonstration of Zoom Capabilities}

\begin{figure}[H]
\centering
\includegraphics[width=0.88\columnwidth]{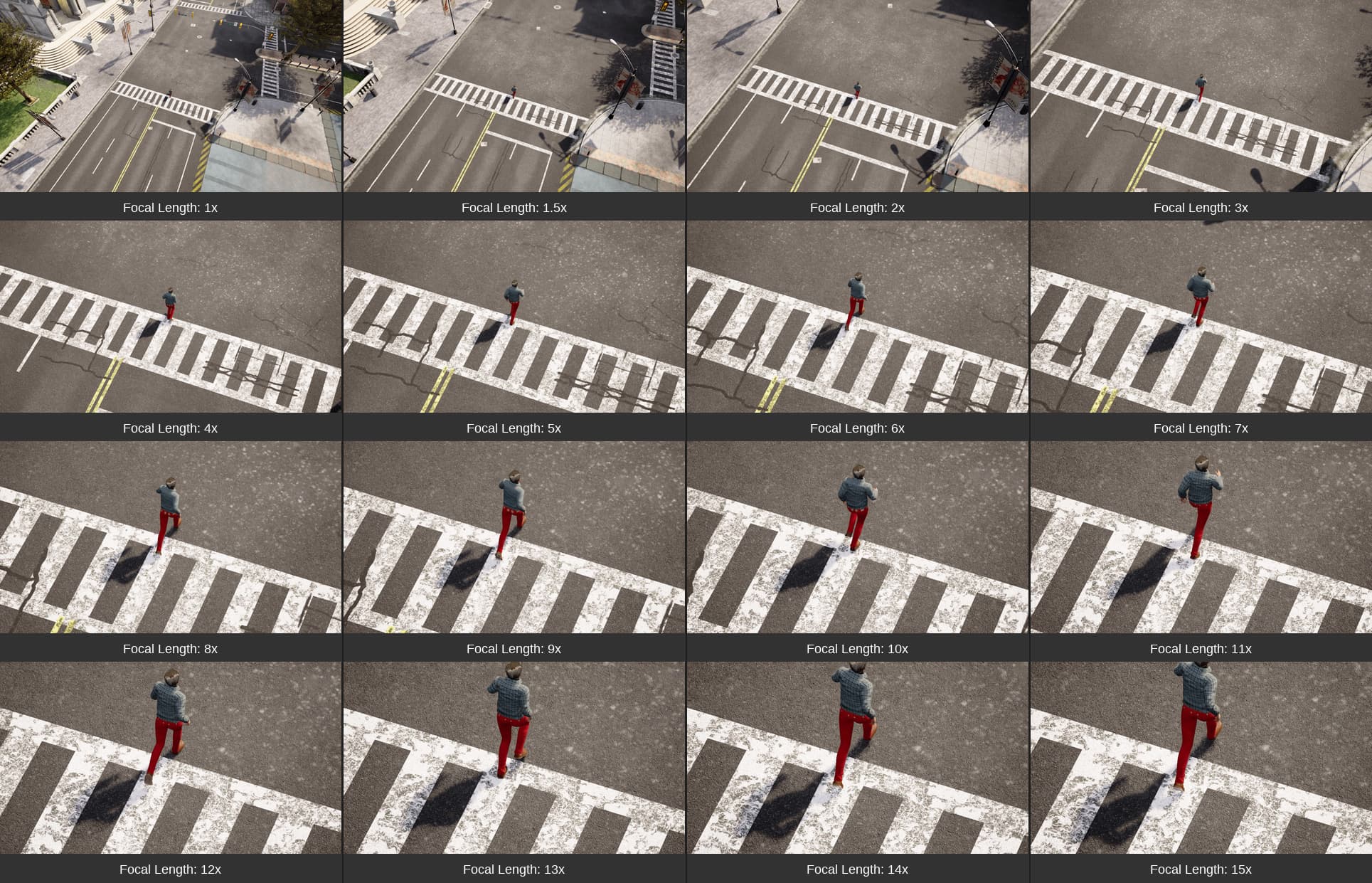}
\caption{Visual demonstration of zoom capabilities across four FOV configurations. Each row shows different tracking scenarios, while columns demonstrate the effect of varying focal lengths. Left to right: wide-angle (110$^\circ$), standard (90$^\circ$), narrow (60$^\circ$), and telephoto (30$^\circ$). Longer focal lengths (lower FOV) provide larger target representations suitable for fine-grained tracking, while shorter focal lengths provide broader situational awareness.}
\label{fig:zoom_demo}
\end{figure}

\Cref{fig:zoom_demo} demonstrates the visual effect of different zoom levels on the same scene. The figure shows a 4$\times$4 grid comparing wide-angle (110$^\circ$ FOV), standard (90$^\circ$ FOV), narrow (60$^\circ$ FOV), and telephoto (30$^\circ$ FOV) configurations. As the FOV decreases, the target pedestrian becomes progressively larger in the frame, enabling finer-grained visual recognition at the cost of reduced peripheral context. This capability is essential for aerial tracking applications where the target distance varies significantly during flight, requiring adaptive zoom to maintain consistent target visibility and resolution.

\begin{figure}[!b]
\centering
\includegraphics[width=0.82\columnwidth]{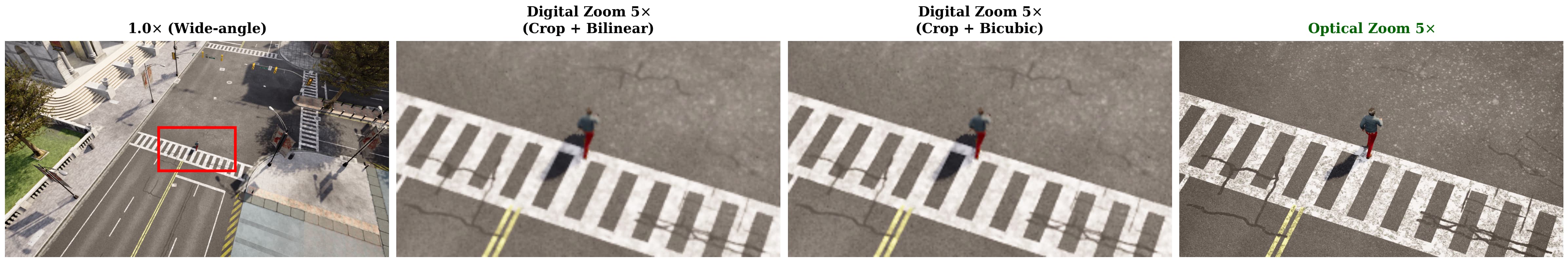}
\caption{Qualitative comparison between digital zoom and optical zoom at 5$\times$ magnification. From left to right: the original wide-angle image (1.0$\times$) with the crop region highlighted in red, digital zoom via center-crop with bilinear interpolation, digital zoom via center-crop with bicubic interpolation, and the corresponding optical zoom capture. Despite employing higher-order interpolation kernels, digital zoom introduces visible blurring and loss of fine-grained texture, whereas optical zoom preserves sharp edges and high-frequency details faithfully.}
\label{fig:zoom_fidelity}
\end{figure}

\paragraph{Optical vs.\ digital zoom fidelity.}
As shown in \Cref{fig:zoom_fidelity}, a qualitative comparison at 5$\times$ magnification reveals that digital zoom---whether using bilinear or bicubic interpolation---introduces visible blurring and loss of fine-grained texture, whereas optical zoom preserves sharp edges and high-frequency details faithfully. This confirms that optical zoom captures genuinely richer visual information that digital upscaling cannot recover, justifying the pipeline's native FOV-based zoom mechanism.

\section{Dual-Track Data Augmentation Design}
\label{app:dual_track}

To enhance the diversity and robustness of the \cosfly dataset, we develop a sophisticated dual-track data augmentation system that generates paired original and perturbed trajectories. This augmentation scheme is designed to support both denoising and prediction tasks in downstream VLM training.

\subsection{Joint Perturbation Framework}

The rendering system implements a joint probability-based perturbation mechanism that independently controls position and rotation augmentations on a per-frame basis. For each augmented trajectory, two independent Bernoulli events determine the application of perturbations:

\paragraph{Position perturbation ($P_{\text{pos}}$).}
With probability $P_{\text{pos}}$ (default 0.6), position perturbations are applied to both the pedestrian and drone. The pedestrian is displaced within a configurable radius $R_h$ using either cubic or spherical sampling strategies, with the vertical coordinate constrained by the ground height function. The drone position is similarly perturbed within radius $R_d$. Failed perturbations (e.g., due to pixel visibility constraints) trigger automatic fallback to unperturbed positions.

\paragraph{Rotation perturbation ($P_{\text{rot}}$).}
With probability $P_{\text{rot}}$ (default 0.6), the camera viewing direction is perturbed by sampling a 3D offset around the pedestrian target position. The offset is sampled within a configurable radius using cubic or spherical strategies, and the resulting viewing angle difference is clipped to a maximum deviation (default 5$^\circ$) per axis to maintain visual coherence.

\begin{figure}[t]
\centering
\includegraphics[width=0.86\columnwidth]{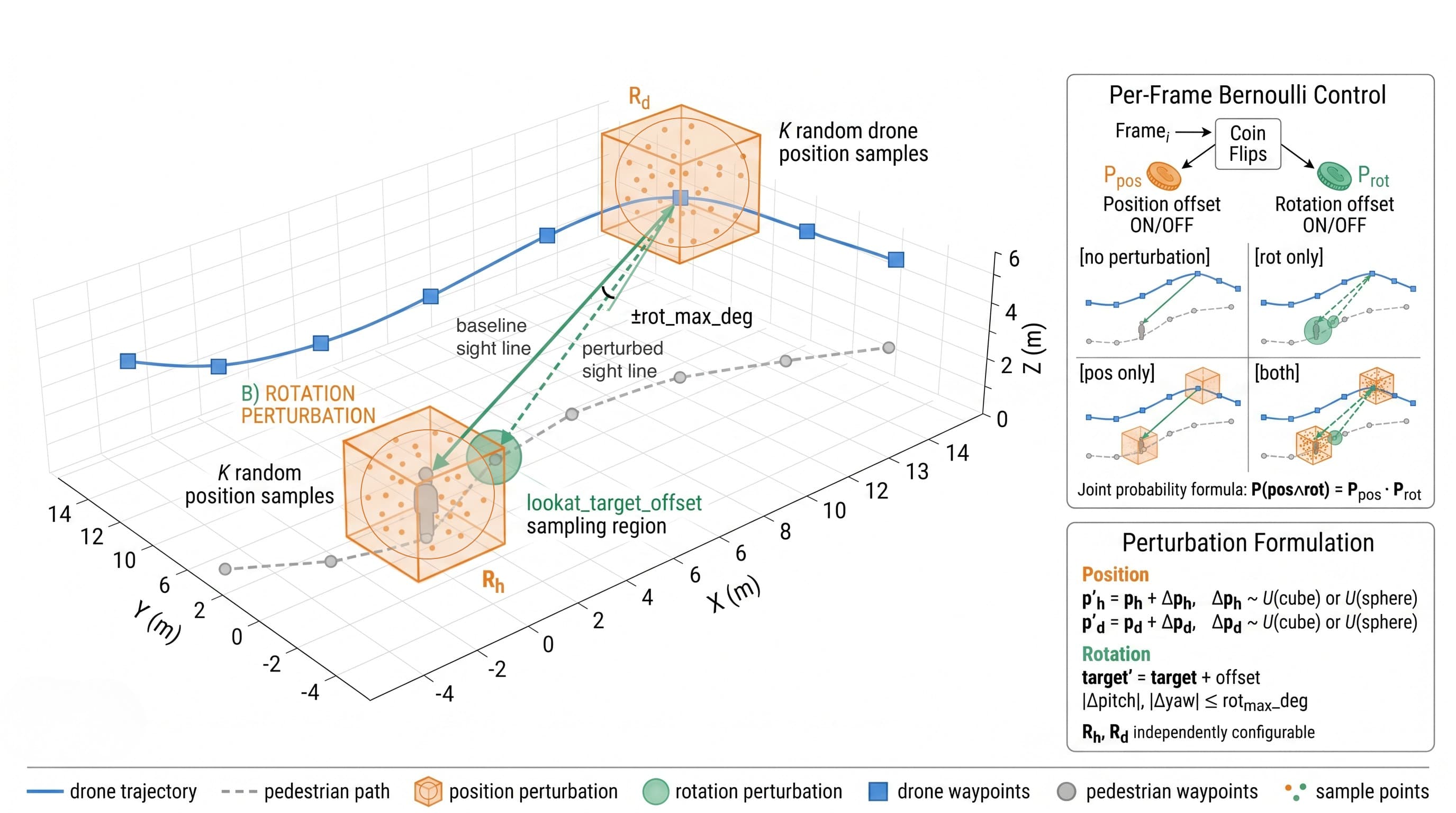}
\caption{Joint perturbation system overview. The figure shows the perturbation parameter space including position offsets ($R_h$ for pedestrian, $R_d$ for drone) and rotation offsets (viewing direction deviation). Independent Bernoulli events $P_{\text{pos}}$ and $P_{\text{rot}}$ control the per-frame application of these perturbations.}
\label{fig:joint_perturbation}
\end{figure}

\Cref{fig:joint_perturbation} illustrates the perturbation parameter space and the effect of different perturbation configurations on the rendered frames.

\paragraph{Augmentation state distribution.}
Since position and rotation perturbations are applied independently, each frame falls into one of four augmentation states. \Cref{tab:augmentation_states} shows the theoretical probability distribution based on the default parameters $P_{\text{pos}} = P_{\text{rot}} = 0.6$. This design ensures that 36\% of frames receive full perturbation (both position and rotation), while 16\% remain unperturbed, providing a balanced mix of augmented and clean samples for robust model training.

\begin{table}[tb]
\centering
\caption{Augmentation state distribution ($P_{\text{pos}} = P_{\text{rot}} = 0.6$).}
\label{tab:augmentation_states}
\begin{tabular}{lcccl}
\toprule
\textbf{State} & \textbf{Position} & \textbf{Rotation} & \textbf{Probability} & \textbf{Description} \\
\midrule
Unperturbed & \xmark & \xmark & 16\% & Clean ground-truth frame \\
Position only & \cmark & \xmark & 24\% & Spatial displacement applied \\
Rotation only & \xmark & \cmark & 24\% & Viewing angle deviation applied \\
Full perturbation & \cmark & \cmark & 36\% & Both augmentations applied \\
\midrule
\multicolumn{3}{l}{\textbf{Total perturbed}} & \textbf{84\%} & \\
\bottomrule
\end{tabular}
\end{table}

\subsection{Dual-Trajectory Data Synthesis}

The data synthesis scheme constructs training samples by combining original ground-truth trajectories with their perturbed counterparts using a sliding window mechanism. This design enables simultaneous training for trajectory denoising and prediction tasks.

\paragraph{Core data structure.}
Each trajectory consists of two parallel tracks: (1) the \textbf{original trajectory} representing the ground-truth flight path with 170--200 consecutive waypoints, and (2) the \textbf{perturbed trajectory} generated by applying random noise to each waypoint to simulate sensor observation errors.

\paragraph{Sliding window sampling.}
A fixed-length window of 10 frames slides along the trajectory with a configurable stride. Within each window position $i$, the frames are partitioned into:
\begin{itemize}[nosep]
    \item \textbf{Observation window} (frames 0--4): The first 5 frames contain perturbed trajectory observations that serve as model input.
    \item \textbf{Prediction horizon} (frames 5--9): The last 5 frames contain only ground-truth positions for future prediction targets.
\end{itemize}

\paragraph{Multi-task label construction.}
For each window sample, the model receives:
\begin{itemize}[nosep]
    \item \textbf{Input}: Perturbed positions $[\tilde{P}_i, \tilde{P}_{i+1}, \tilde{P}_{i+2}, \tilde{P}_{i+3}, \tilde{P}_{i+4}]$
    \item \textbf{Denoising target}: Original positions $[P_i, P_{i+1}, P_{i+2}, P_{i+3}, P_{i+4}]$ for the observation window
    \item \textbf{Prediction target}: Original positions $[P_{i+5}, P_{i+6}, P_{i+7}, P_{i+8}, P_{i+9}]$ for the prediction horizon
\end{itemize}

\begin{figure}[t]
\centering
\includegraphics[width=0.85\columnwidth]{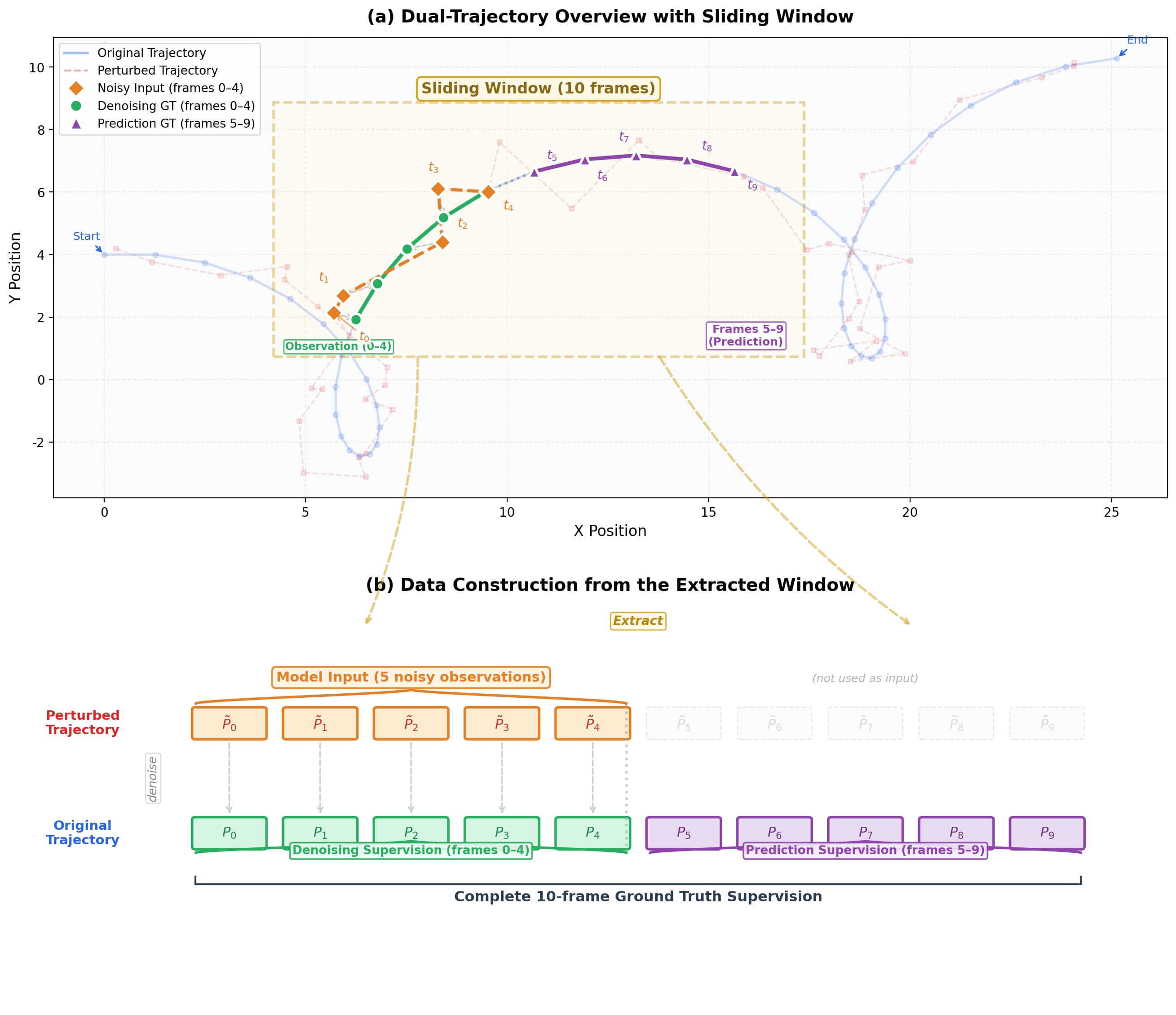}
\caption{Dual-trajectory data synthesis pipeline. (a) A sliding window of size 10 moves along the paired original and perturbed trajectories, partitioning each segment into an observation zone (frames 0--4) and a prediction horizon (frames 5--9). (b) Data construction from the extracted window: the first 5 perturbed frames serve as noisy model input, while the complete 10-frame original trajectory provides ground truth supervision for both denoising (frames 0--4) and prediction (frames 5--9) tasks.}
\label{fig:dual_trajectory_pipeline}
\end{figure}

\Cref{fig:dual_trajectory_pipeline} illustrates the data pipeline and the construction of multi-task training samples from the dual-track trajectory representation.

\subsection{Configuration Parameters}

\Cref{tab:augmentation_params} summarizes the key configuration parameters for the augmentation system.

\begin{table}[H]
\centering
\small
\caption{Data augmentation configuration parameters.}
\label{tab:augmentation_params}
\begin{tabular}{lll}
\toprule
Parameter & Default & Description \\
\midrule
\multicolumn{3}{l}{\textit{Joint Perturbation}} \\
$P_{\text{pos}}$ & 0.6 & Position perturbation probability \\
$P_{\text{rot}}$ & 0.6 & Rotation perturbation probability \\
$R_h$ (pedestrian) & 2.0\,m & Pedestrian position radius \\
$R_d$ (drone) & 3.0\,m & Drone position radius \\
$\theta_{\max}$ & 5$^\circ$ & Maximum viewing angle deviation \\
\midrule
\multicolumn{3}{l}{\textit{Sliding Window Sampling}} \\
Window size & 10 frames & Total frames per sample \\
Observation window & 5 frames & Input sequence length \\
Prediction horizon & 5 frames & Target sequence length \\
Stride & 3 frames & Window sliding step \\
\bottomrule
\end{tabular}
\end{table}

\begin{figure}[!t]
\centering
\includegraphics[width=0.9\columnwidth]{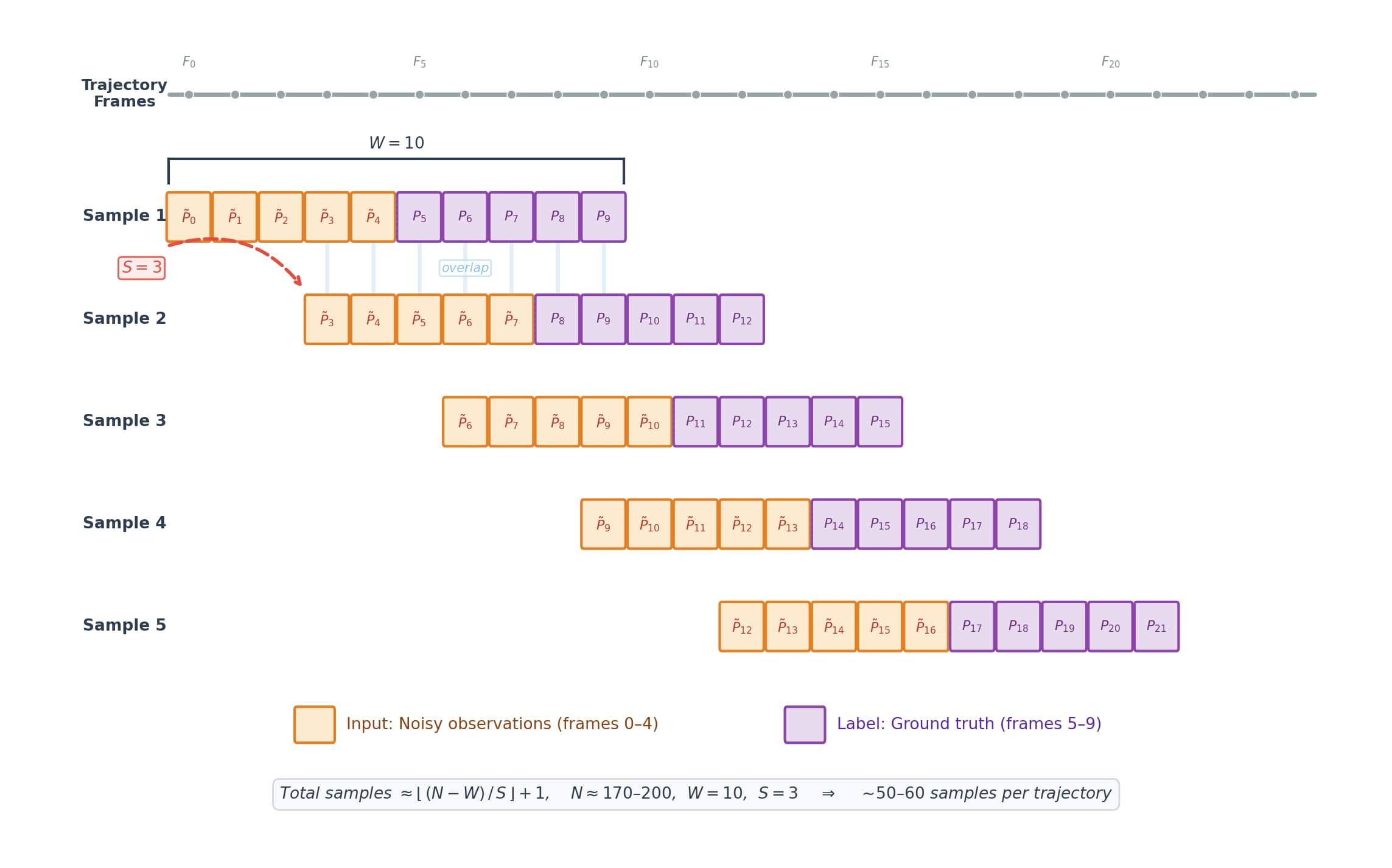}
\caption{Training sample generation through sliding window sampling. Each trajectory of 170--200 waypoints yields 50--60 overlapping training samples with window size 10 and stride 3, effectively amplifying the dataset size while preserving temporal continuity between adjacent samples.}
\label{fig:sample_generation}
\end{figure}

\paragraph{Sample generation statistics.}
For a trajectory of length $N$ with window size 10 and stride $S$, approximately $(N - 10) / S + 1$ training samples are generated per trajectory. With typical trajectories of 170--200 waypoints and stride 3, each trajectory yields approximately 50--60 training samples, significantly amplifying the effective dataset size for model training. \Cref{fig:sample_generation} illustrates this amplification process.

\section{Rendering Efficiency and System Utilization}
\label{app:rendering_efficiency}

The generation of the CosFly-Track dataset requires substantial computational resources to render high-fidelity multi-modal sensor data across diverse simulation environments. To ensure scalable and robust data production, we implemented a distributed rendering pipeline with comprehensive system monitoring and automated fault recovery mechanisms. This section details the rendering efficiency, hardware utilization, and system stability observed during the large-scale data generation process.

\subsection{Distributed Rendering Architecture}
\label{sec:appendix_rendering_arch}

The rendering pipeline is deployed across a heterogeneous cluster of four high-performance computing nodes, collectively providing 6 GPUs and 47 concurrent rendering workers. As summarized in \Cref{tab:machine_configs}, the cluster combines single-GPU and dual-GPU machines equipped with NVIDIA RTX 6000 Ada Generation (48\,GiB) and NVIDIA RTX PRO 6000 Blackwell (96\,GiB) GPUs. Worker counts are dynamically tuned to maximize hardware utilization while staying within the available VRAM budget of each node.

\begin{table}[tb]
\centering
\caption{Hardware configuration and rendering performance across the distributed cluster during the v7 data generation phase.}
\label{tab:machine_configs}
\resizebox{\linewidth}{!}{%
\begin{tabular}{lcccccc}
\toprule
\textbf{Node ID} & \textbf{GPU Config} & \textbf{Workers} & \textbf{Duration (h)} & \textbf{Completed Traj.} & \textbf{Mean GPU Util.} & \textbf{Role} \\
\midrule
Machine A (193) & $2\times$ RTX PRO 6000 Blackwell (96GB) & 16 & 94.7 & 814 & 87.0\% & High-throughput rendering \\
Machine B (37)  & $1\times$ RTX 6000 Ada Generation (48GB) & 12 & 91.5 & 497 & 94.0\% & Dense map rendering \\
Machine C (195) & $2\times$ RTX PRO 6000 Blackwell (96GB) & 18 & 85.0 & 1,178 & 89.5\% & Batch processing \\
Machine D (38)  & $1\times$ RTX 6000 Ada Generation (48GB) & 1 & 86.5 & 175 & N/A & Debugging and fallback \\
\midrule
\textbf{Total} & \textbf{6 GPUs} & \textbf{47} & \textbf{--} & \textbf{2,664} & \textbf{--} & \textbf{--} \\
\bottomrule
\end{tabular}%
}
\end{table}

The pipeline utilizes a watchdog mechanism to automatically monitor and restart the CARLA simulator and rendering processes. This ensures continuous operation despite occasional simulator crashes caused by memory leaks or physics engine instability in complex maps.

\subsection{Cumulative Production and Map Completion}
\label{sec:appendix_production}

\begin{figure}[t]
\centering
\includegraphics[width=\columnwidth]{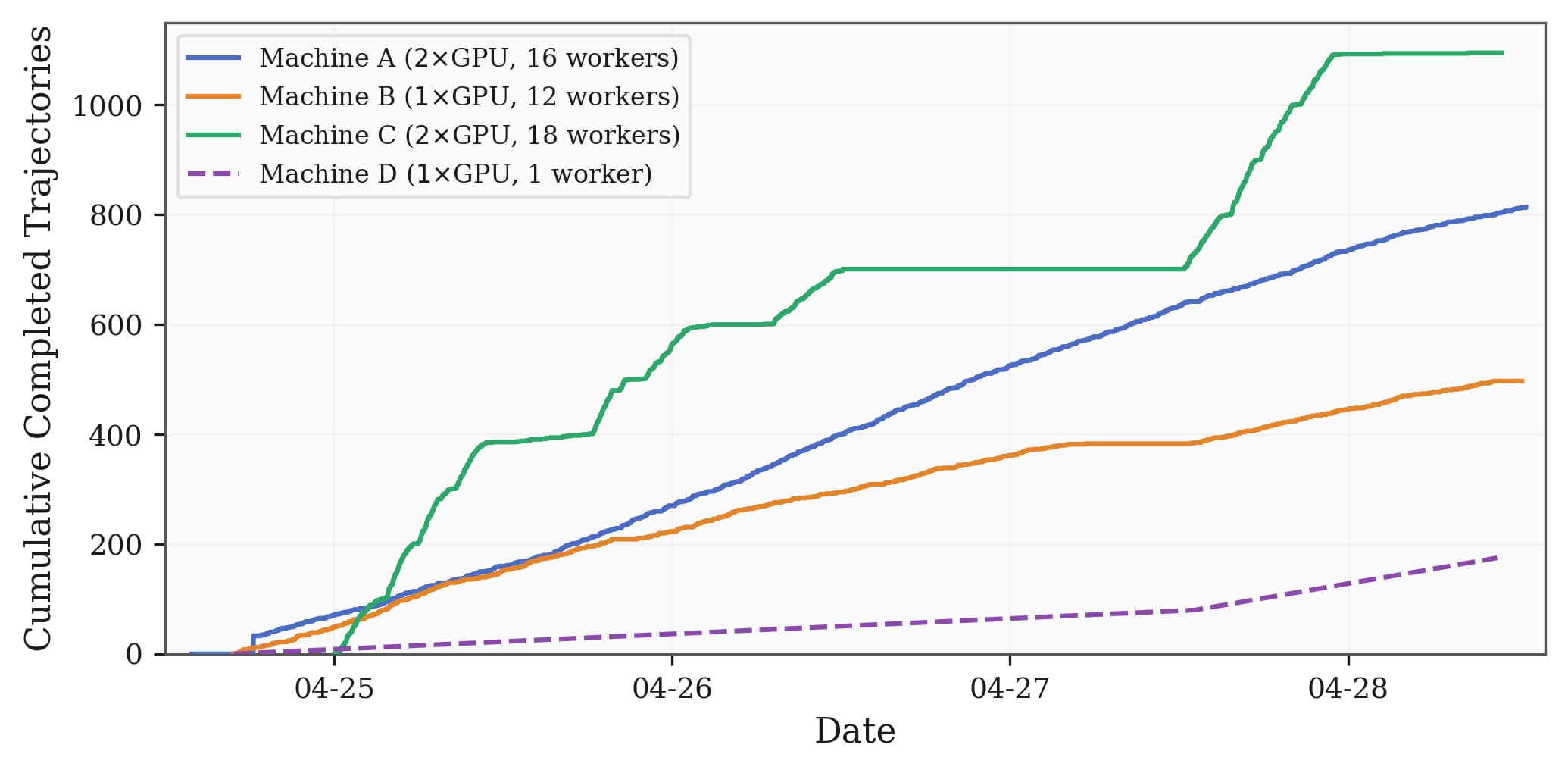}
\caption{Cumulative trajectory production across the distributed cluster over a four-day period. Dual-GPU nodes (Machines A and C) exhibit steeper production curves, while the single-GPU node (Machine B) maintains a steady but lower throughput. Machine D was primarily used for targeted map completion.}
\label{fig:cumulative_trajectories}
\end{figure}

The data generation process successfully produced 2,664 complete trajectories across 15 distinct maps. \Cref{fig:cumulative_trajectories} illustrates the cumulative trajectory production over a four-day continuous rendering period. The dual-GPU nodes (Machines A and C) demonstrated significantly higher throughput, with Machine C achieving the highest production rate by completing 1,178 trajectories.

\begin{figure}[tb]
\centering
\includegraphics[width=\columnwidth]{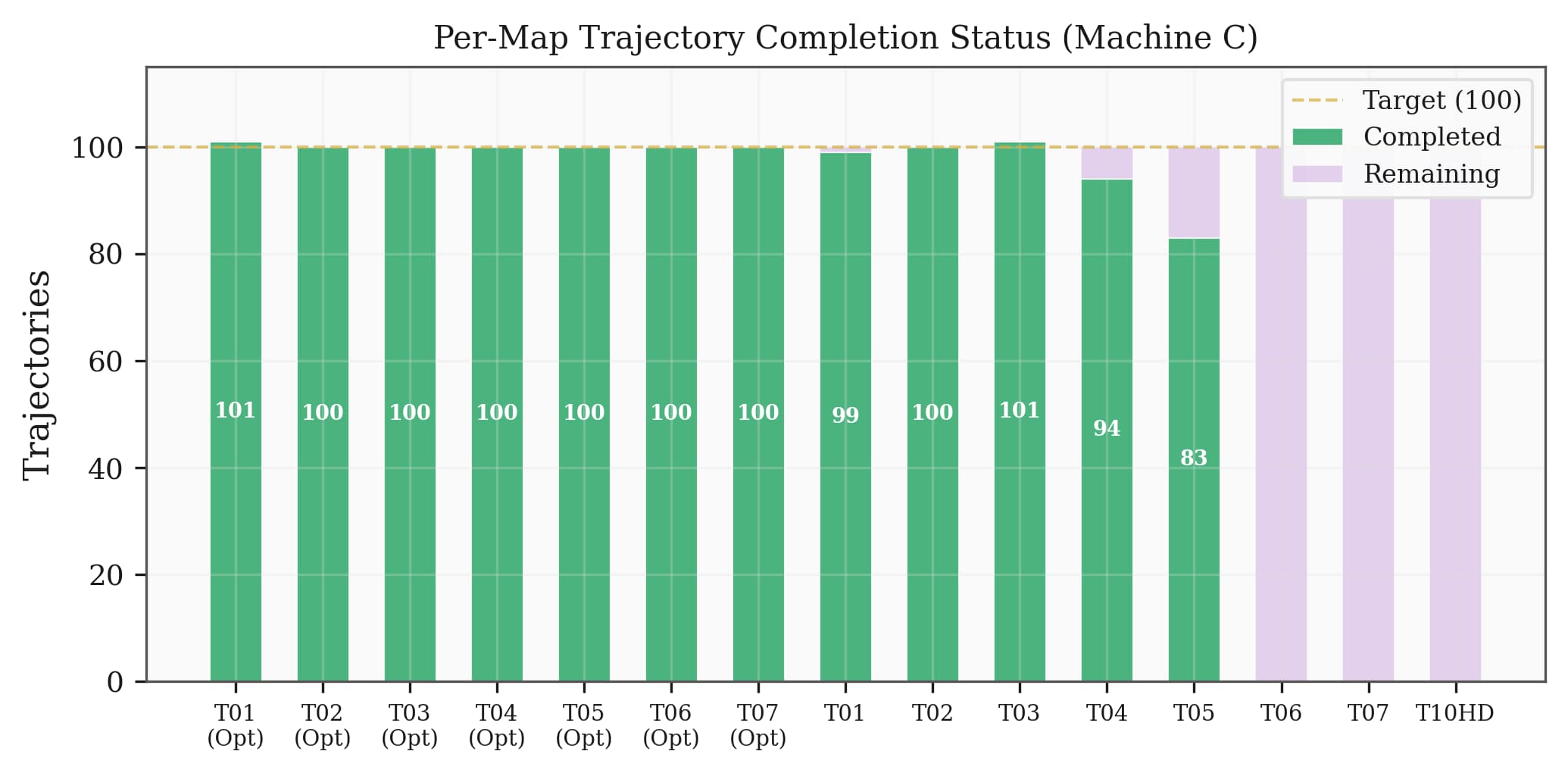}
\caption{Per-map trajectory completion status on Machine C. The system successfully generated the target 100 trajectories for all optimized maps (T01--T07 Opt) and most standard maps, demonstrating the robustness of the automated rendering pipeline across diverse environments.}
\label{fig:per_map_completion}
\end{figure}

The rendering workload was distributed across both optimized (\texttt{\_Opt}) and standard maps. \Cref{fig:per_map_completion} details the per-map completion status on Machine C, which was tasked with rendering 100 trajectories per map. The system successfully reached the target for all optimized maps and most standard maps, with minor shortfalls in computationally heavy environments like Town04 and Town05 due to simulator timeouts.

\subsection{Hardware Utilization and System Stability}
\label{sec:appendix_hardware}

\begin{figure}[!htbp]
\centering
\includegraphics[width=0.80\columnwidth]{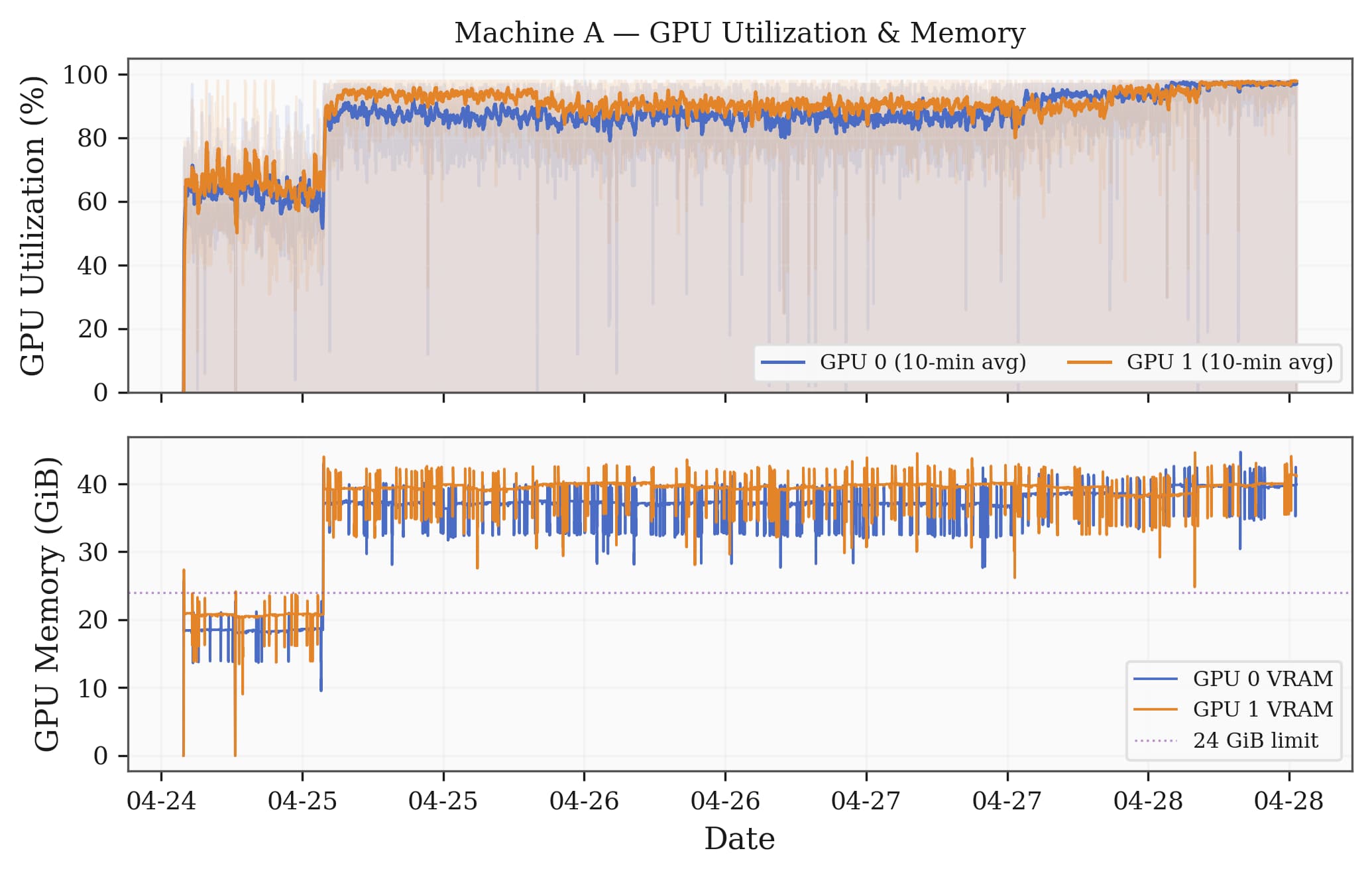}
\caption{GPU utilization and memory usage on Machine A (dual-GPU) over the rendering period. The pipeline maintains high, stable utilization (averaging 87\%) while keeping VRAM usage safely below the 96\,GiB per-GPU hardware limit. The brief drops in utilization correspond to automated simulator restarts triggered by the watchdog mechanism.}
\label{fig:gpu_utilization}
\end{figure}

Maximizing GPU utilization while maintaining system stability is a critical challenge in large-scale simulation. \Cref{fig:gpu_utilization} presents the GPU utilization and VRAM usage for Machine A, which is equipped with dual NVIDIA RTX PRO 6000 Blackwell GPUs. The system consistently maintained high GPU utilization (averaging 87.0\%) across both GPUs. VRAM usage remained well below the 96\,GiB per-GPU hardware limit, validating our choice of 16 concurrent workers for this node.

\begin{figure}[H]
\centering
\includegraphics[width=0.68\columnwidth]{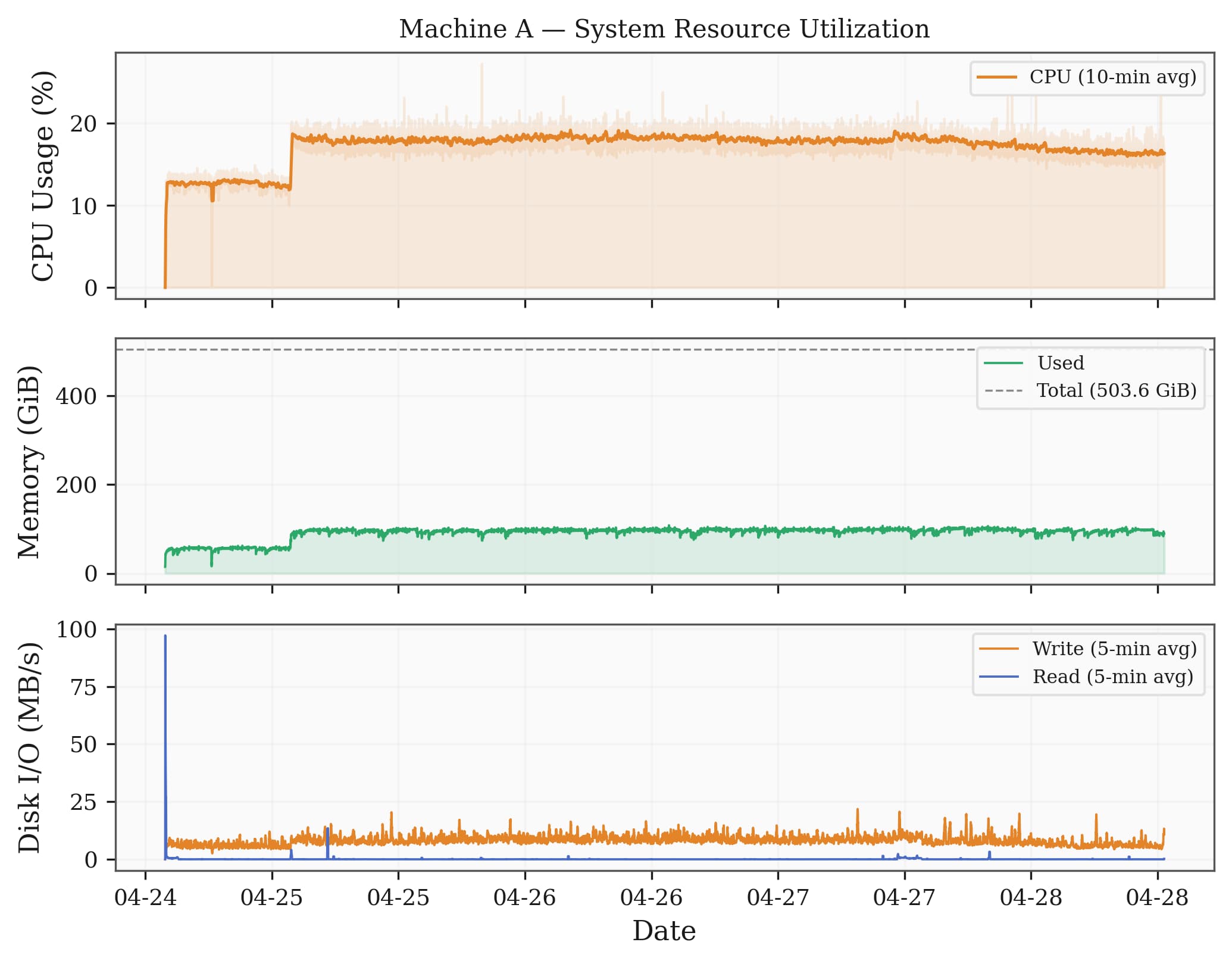}
\caption{System resource utilization on Machine A. CPU usage remains stable at $\sim$18\%, confirming the GPU-bound nature of the rendering task. System memory plateaus at $\sim$100\,GiB, and disk I/O shows consistent write operations for saving multi-modal frames.}
\label{fig:system_resources}
\end{figure}

\Cref{fig:system_resources} further details the CPU, system memory, and disk I/O metrics. CPU usage remained stable at around 18\%, indicating that the rendering process is heavily GPU-bound rather than CPU-bound. System memory usage plateaued at approximately 100\,GiB, well within the 503.6\,GiB capacity. Disk I/O shows consistent write operations corresponding to the continuous saving of multi-modal frames, with minimal read operations after the initial map loading phase.

\FloatBarrier
\subsection{Automated Fault Recovery}
\label{sec:appendix_fault_recovery}

\begin{figure}[!htbp]
\centering
\includegraphics[width=0.82\columnwidth]{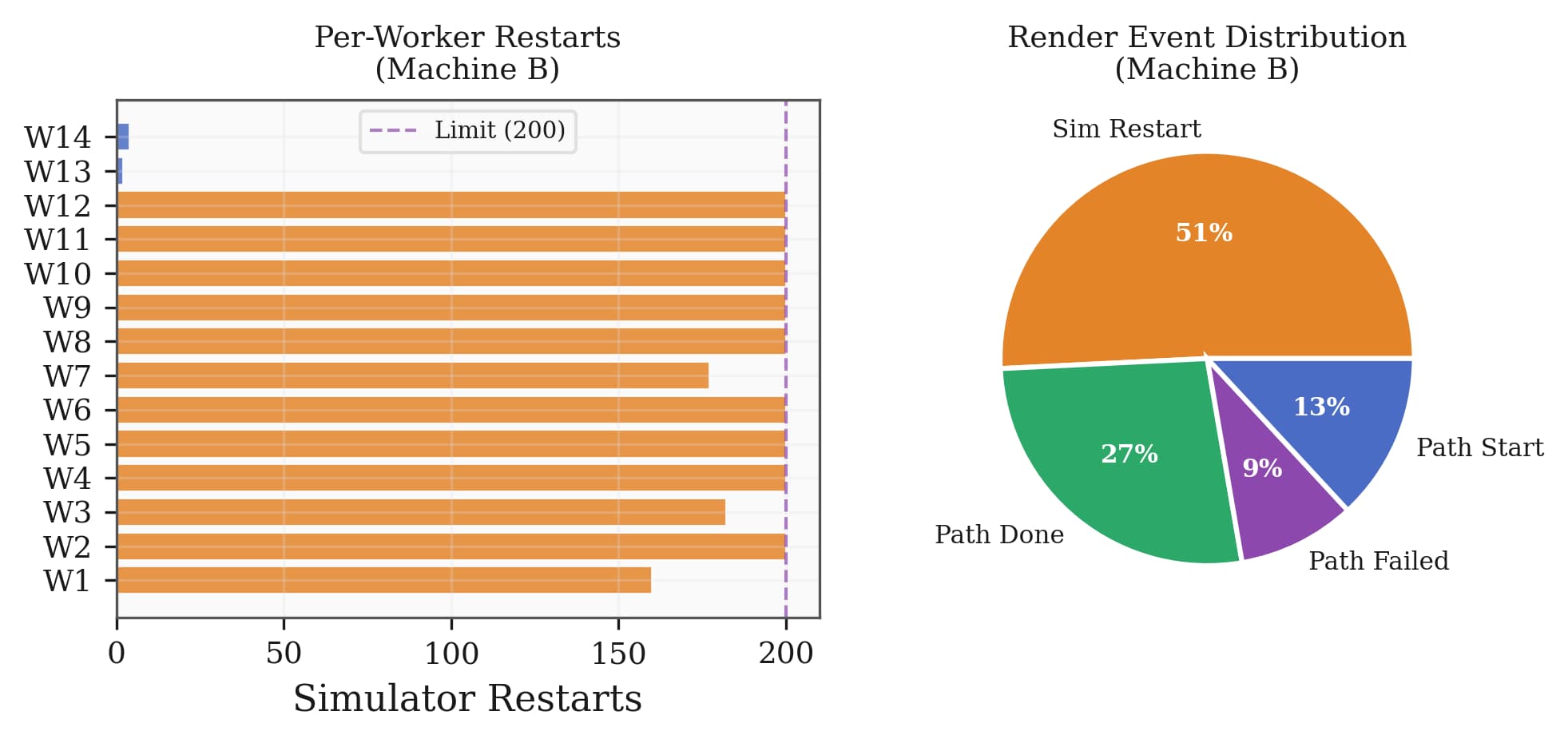}
\caption{Simulator restart analysis on Machine B. Left: Per-worker restart counts, showing that most workers reached or approached the configured maximum limit of 200 restarts. Right: Render event distribution, highlighting that automated simulator restarts account for 51\% of all logged events, underscoring the critical role of the watchdog mechanism in maintaining continuous operation.}
\label{fig:simulator_restarts}
\end{figure}

Long-running simulations in CARLA are prone to occasional crashes. Our watchdog mechanism automatically detects stalled processes and restarts the simulator. \Cref{fig:simulator_restarts} analyzes the restart events on Machine B. Over the 91.5-hour period, the system executed 2,325 automated simulator restarts across 12 workers. The event distribution shows that 51\% of logged events were simulator restarts, highlighting the necessity of the watchdog system for continuous data generation. Despite these interruptions, the pipeline successfully completed 27\% of path rendering attempts, ensuring steady progress without manual intervention.

\FloatBarrier

\FloatBarrier
\section{UAV Chain-of-Cause (CoC) Data Pipeline}
\label{sec:appendix_e}

\begin{figure}[tb]
    \centering
    \includegraphics[width=\linewidth]{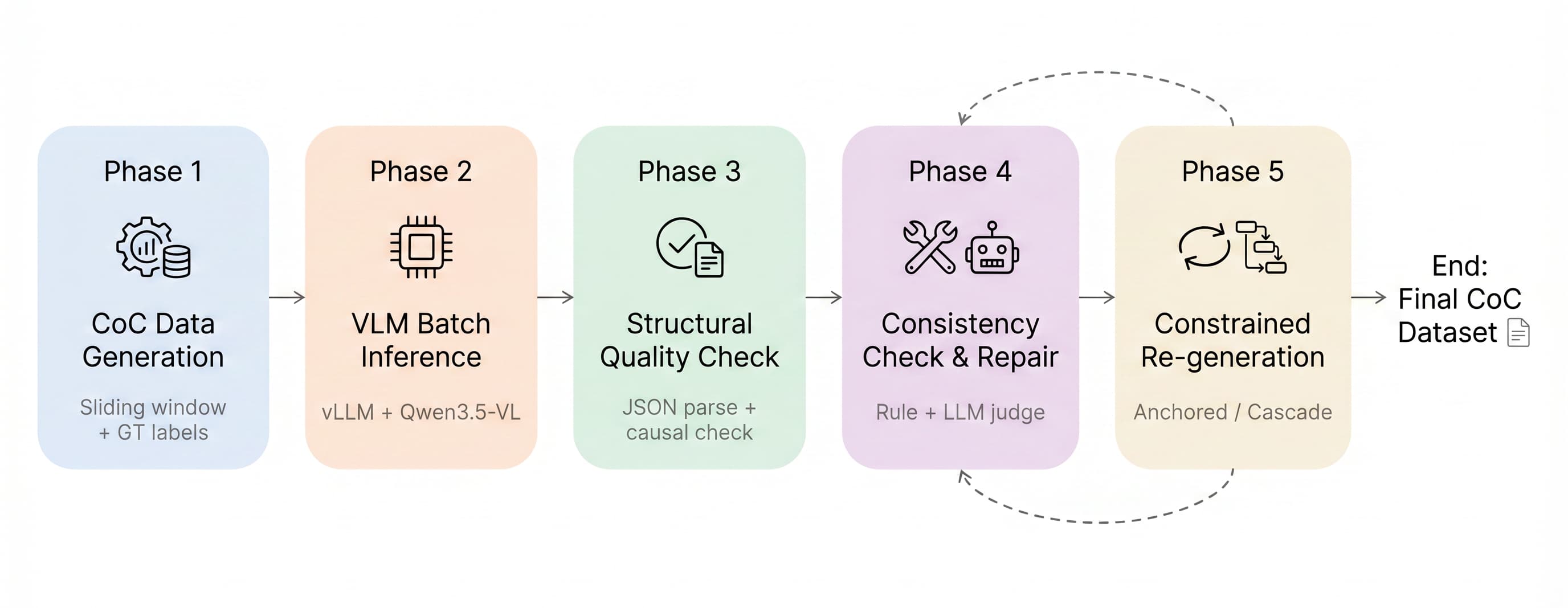}
    \caption{The five-phase UAV CoC data production pipeline. The workflow ensures causal safety through sliding window sampling and guarantees geometric consistency via a multi-stage verification and repair mechanism.}
    \label{fig:pipeline_flowchart}
\end{figure}

To train the UAV-VLN multimodal large language model, we developed a comprehensive Chain-of-Cause (CoC) data production pipeline. This pipeline transforms raw flight trajectories into high-quality, causally safe instruction-tuning data. The pipeline consists of five distinct phases: data generation, batch inference, structural quality check, trajectory consistency check, and constrained re-generation. \Cref{fig:pipeline_flowchart} illustrates the end-to-end workflow.

\subsection{Causal Locality and Sliding Window Sampling}

The CoC paradigm, following the reasoning-and-action formulation introduced by Alpamayo-R1~\citep{wang2025alpamayo}, is designed to enforce \textit{causal locality}. Unlike traditional Chain-of-Thought (CoT) approaches where the model is provided with both historical and future frames, our CoC pipeline strictly limits the model's observation to a historical sliding window. This ensures that the model learns to make decisions based solely on observable past information, preventing future information leakage during training.

\begin{figure}[tb]
    \centering
    \includegraphics[width=\linewidth]{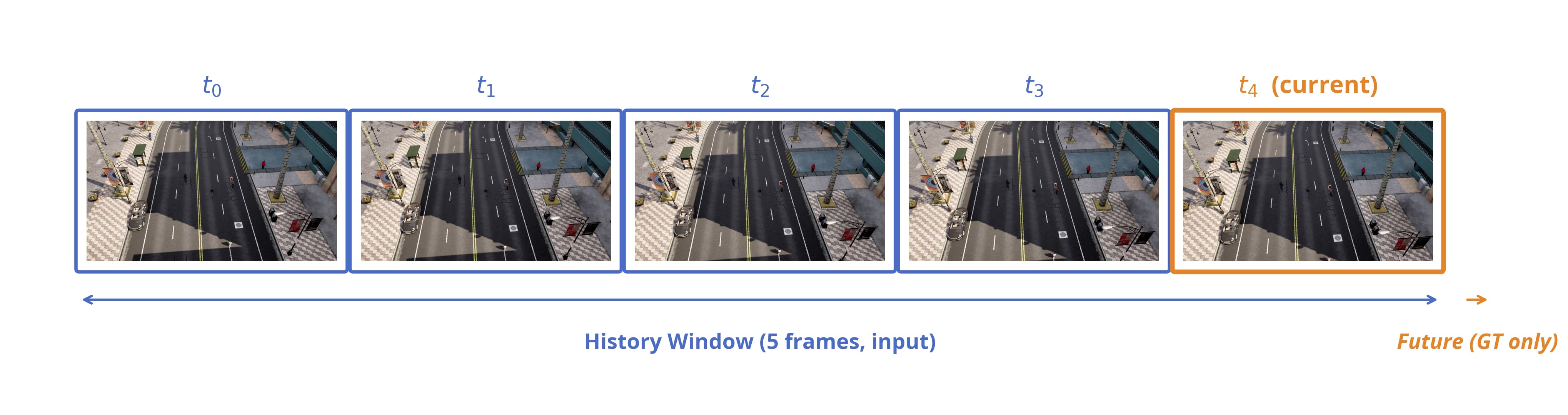}
    \caption{Sliding window observation sequence. The model receives 5 historical frames ($t_0$ to $t_4$) and corresponding flight state data as input. The future trajectory is hidden and used only to derive the ground-truth label.}
    \label{fig:sliding_window}
\end{figure}

As shown in \Cref{fig:sliding_window}, each sample is constructed using a 5-frame history window (sampled at 0.5s intervals). The model must analyze the target's motion trend, environmental obstacles, and the drone's flight state within this window to predict the required flight adjustment. The future trajectory (4 frames) is entirely hidden from the model and is used exclusively to extract the ground-truth (GT) flight decision label.

\subsection{Structured CoC Generation and Batch Inference}

\begin{figure}[tb]
    \centering
    \includegraphics[width=\linewidth]{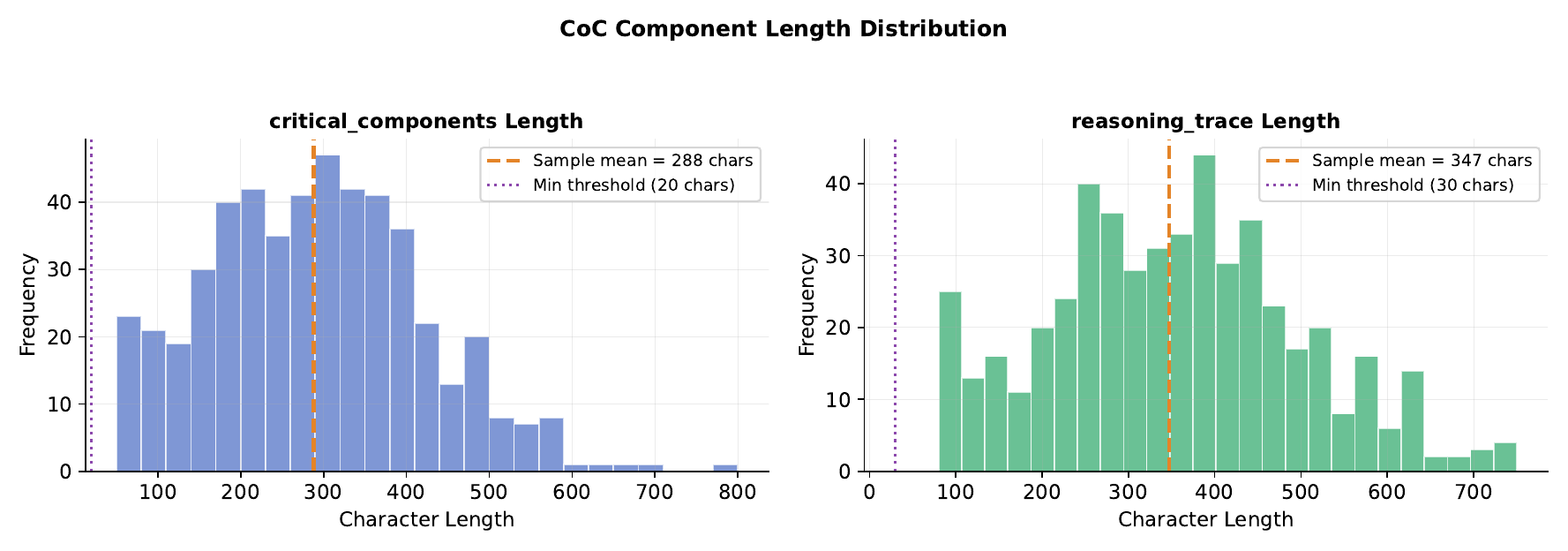}
    \caption{Character length distribution of the generated CoC components. The \texttt{critical\_components} and \texttt{reasoning\_trace} consistently exceed the minimum quality thresholds (20 and 30 characters, respectively), providing rich causal supervision.}
    \label{fig:coc_lengths}
\end{figure}

The model is instructed to output a structured JSON response containing three components: \texttt{critical\_components} (key observable factors), \texttt{reasoning\_trace} (explicit causal logic), and \texttt{flight\_decision} (selected from a predefined 17-option closed set). \Cref{fig:coc_lengths} shows the length distribution of the generated reasoning components, demonstrating that the model produces detailed and substantial causal analysis.

To handle the massive scale of the dataset, we employ vLLM with continuous batching and PagedAttention for high-throughput offline inference. Using the Qwen3.5-397B-A17B-FP8 vision-language teacher with tensor parallelism, the pipeline achieves efficient large-scale data generation while supporting automatic resumption from checkpoints.

\subsection{Trajectory Consistency Verification}

A fundamental challenge of the causal locality principle is that the model's predicted flight decision may diverge from the drone's actual future trajectory. To address this, we implement a rigorous consistency verification mechanism that compares the semantic intent of the model's decision against the geometric properties of the actual trajectory (yaw angle, altitude change, and average speed).

\begin{figure}[tb]
    \centering
    \includegraphics[width=\linewidth]{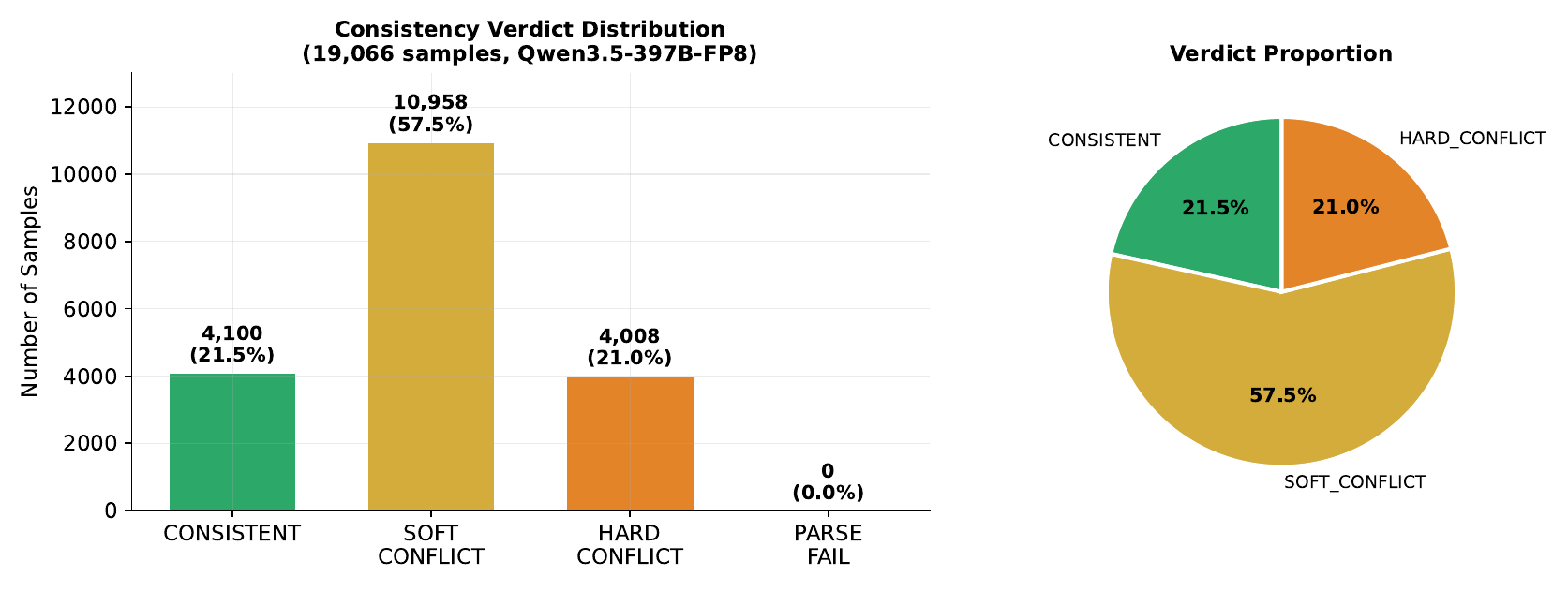}
    \caption{Consistency verdict distribution across 19,066 generated samples. While 21.5\% of samples match the GT exactly, the majority (57.5\%) exhibit soft conflicts where the model's decision is a reasonable alternative to the actual trajectory.}
    \label{fig:consistency_verdict}
\end{figure}

As detailed in \Cref{fig:consistency_verdict}, the verification process classifies each sample into one of three categories:
\begin{itemize}
    \item \textbf{CONSISTENT (21.5\%):} The model's decision exactly matches the GT label.
    \item \textbf{SOFT CONFLICT (57.5\%):} The decisions differ, but there is no direct geometric contradiction. For example, the GT might be ``track straight'', but the model suggests ``yaw left to follow'' to correct a minor visual offset.
    \item \textbf{HARD CONFLICT (21.0\%):} The model's decision directly contradicts the actual trajectory (e.g., suggesting a left turn when the drone actually turned right).
\end{itemize}

\begin{figure}[tb]
    \centering
    \includegraphics[width=0.85\linewidth]{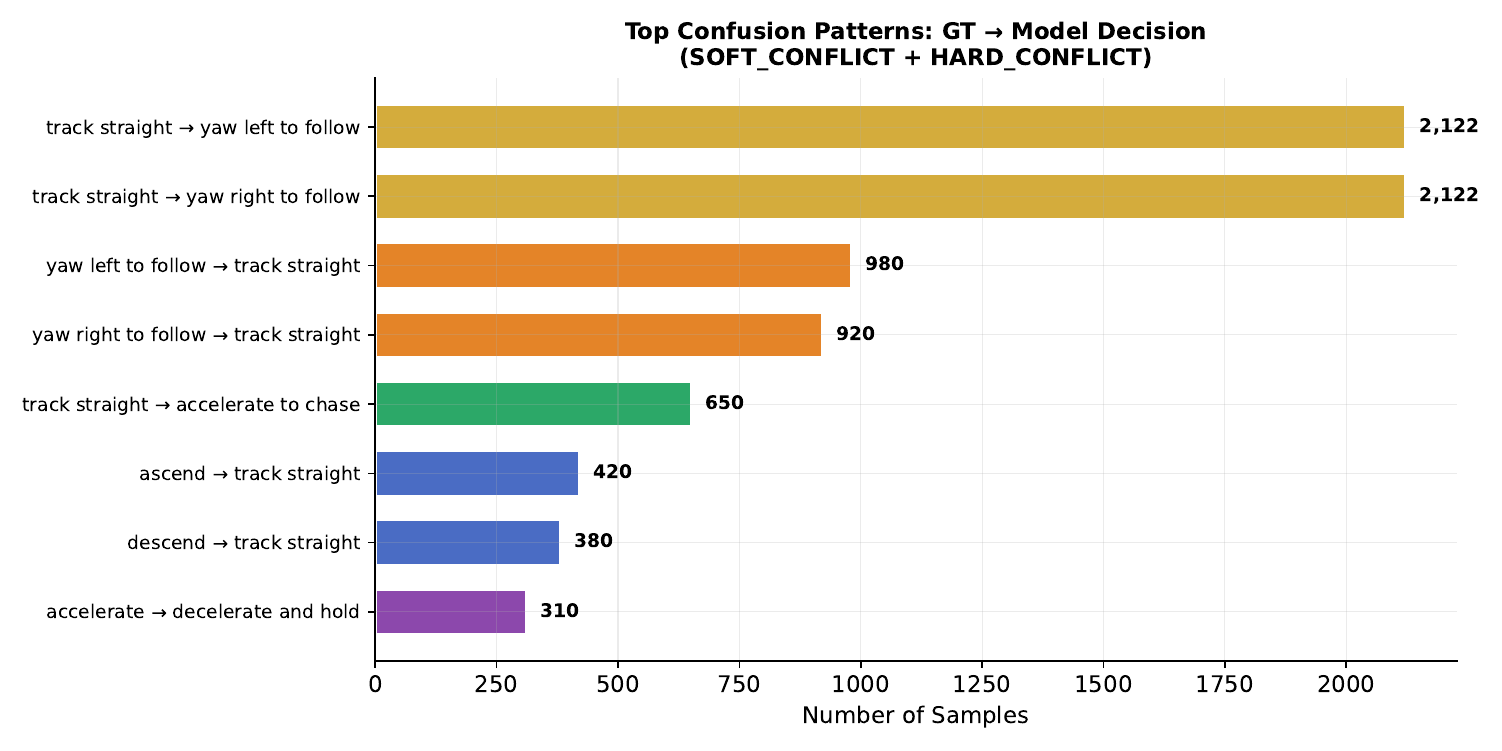}
    \caption{Top confusion patterns between ground-truth labels and model predictions. The most common divergence occurs when the GT indicates straight flight, but the model proactively suggests yaw corrections based on visual target offsets.}
    \label{fig:confusion_patterns}
\end{figure}

\Cref{fig:confusion_patterns} highlights the most frequent confusion patterns. The dominant source of divergence arises from the GT heuristic's strict 3$^\circ$ yaw threshold, whereas the VLM proactively suggests yaw adjustments based on the target's pixel offset in the image frame. For conflict samples, an optional LLM judge evaluates the reasoning trace to determine if the decision is an acceptable alternative, fixable by label replacement, or requires complete re-generation.

\subsection{Constrained Re-generation Strategies}

\begin{figure}[tb]
    \centering
    \includegraphics[width=\linewidth]{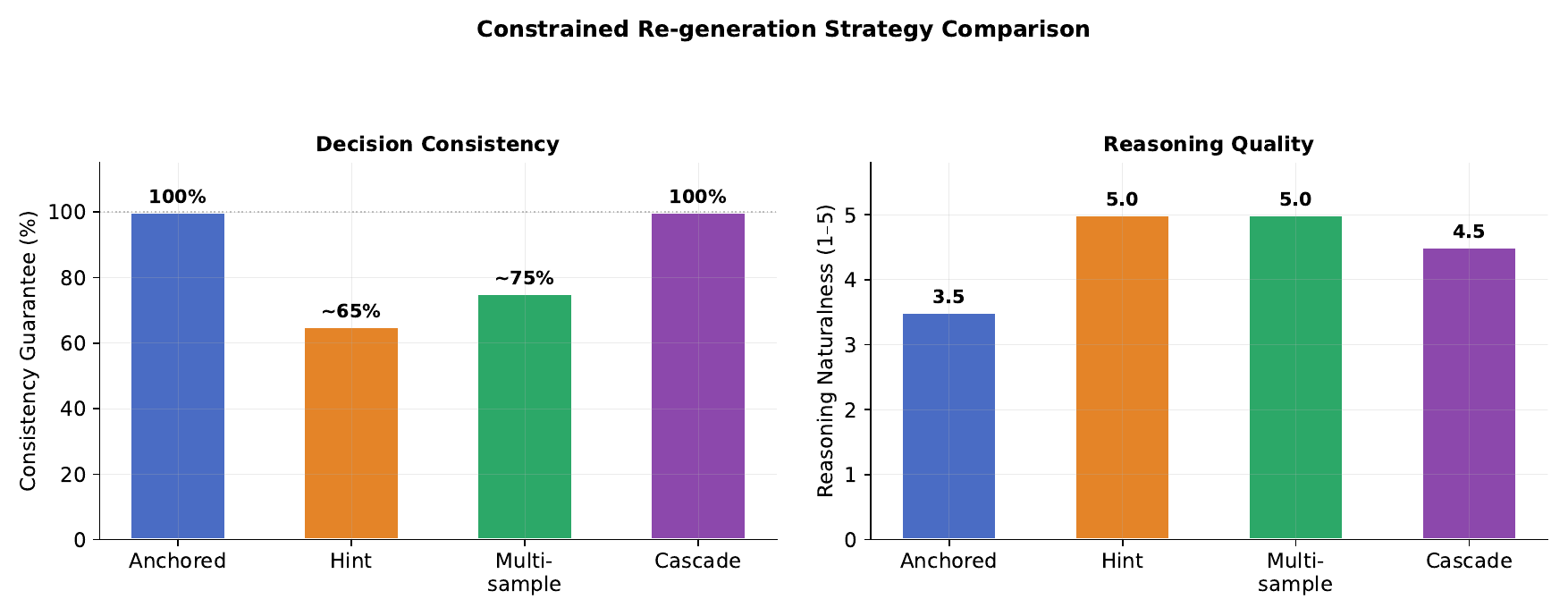}
    \caption{Comparison of constrained re-generation strategies. The \textit{Anchored} strategy guarantees 100\% consistency by injecting the GT decision into the prompt, while the \textit{Cascade} strategy balances naturalness and consistency by falling back through multiple methods.}
    \label{fig:regen_strategy}
\end{figure}

Samples with fundamentally flawed reasoning (REJECT\_REGEN) cannot be fixed by simple label replacement. Furthermore, simply re-running inference with the original prompt often yields the same incorrect result since the model still lacks future visibility. To resolve this, we developed four constrained re-generation strategies, compared in \Cref{fig:regen_strategy}.

The default \textit{Anchored} strategy injects the correct GT flight decision directly into the system prompt as a hard constraint. The model is then tasked with performing ``backward reasoning''—finding observable evidence in the historical frames to justify the provided decision. This approach guarantees 100\% decision consistency while maintaining the structural integrity of the CoC data. For optimal quality, the \textit{Cascade} strategy sequentially attempts hint-based generation and multi-sampling before falling back to the anchored approach, ensuring both high naturalness and guaranteed consistency for the final training dataset.

\subsection{Bilingual Sample Case}
\label{sec:appendix_e_sample}

To make the Chain-of-Cause (CoC) format concrete, we present one bilingual sample from trajectory \texttt{trajectory\_1776047127\_ORI\_frame\_00004}.

\begin{figure}[H]
    \centering
    \includegraphics[width=0.58\linewidth]{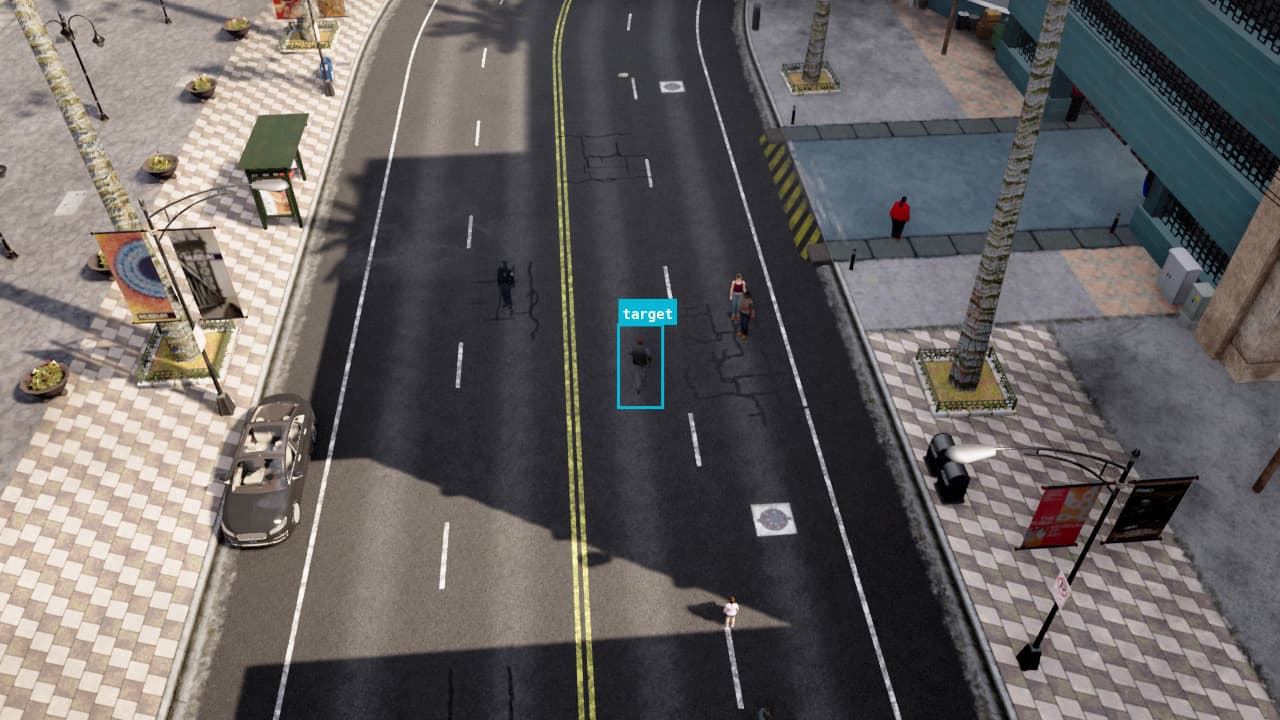}
    \caption{Observation frame for sample \texttt{trajectory\_1776047127\_ORI\_frame\_00004}.}
    \label{fig:coc_sample_frame}
\end{figure}

\noindent Tables~\ref{tab:coc_sample_en} and~\ref{tab:coc_sample_zh} present aligned English and Chinese CoC samples for the same observation window. \Cref{fig:coc_sample_frame} shows the corresponding current frame (the fifth and newest observation image in the 5-frame sliding window) for this sample.

\vspace{-0.8\baselineskip}

\begin{table}[H]
\centering
\scriptsize
\caption{English CoC sample for sample \texttt{trajectory\_1776047127\_ORI\_frame\_00004}.}
\label{tab:coc_sample_en}
\setlength{\tabcolsep}{3pt}
\renewcommand{\arraystretch}{0.95}
\begin{tabular}{@{}l@{\hspace{0.5em}}l@{}}
\toprule
\textbf{Field} & \textbf{Content} \\
\midrule
\texttt{sample\_id} &
\begin{minipage}[t]{0.72\linewidth}\raggedright
\texttt{trajectory\_1776047127\_ORI\_frame\_00004}
\end{minipage} \\[1pt]
\texttt{system\_prompt} &
\begin{minipage}[t]{0.72\linewidth}\raggedright
You are a UAV target-tracking flight decision system. Analyze the drone's historical observation images and flight state, then output a structured Chain-of-Cause (CoC) reasoning in strict JSON.
\end{minipage} \\[1pt]
\texttt{user\_prompt} &
\begin{minipage}[t]{0.72\linewidth}\raggedright
5 observation images (oldest $\rightarrow$ newest; last = current frame) with flight data. Velocity: [0.0, 2.92, 1.46, 0.87, 1.02] m/s. Altitude: [22.0, 21.2, 20.7, 20.6, 20.6] m. Yaw: [0.0, 3.2, 4.4, 4.8, 4.8] deg. Target ground speed: 1.00 m/s. Visibility: [visible, visible, visible, visible, visible].
\end{minipage} \\[1pt]
\texttt{critical\_components} &
\begin{minipage}[t]{0.72\linewidth}\raggedright
Target pedestrian walking straight in the center lane. Bounding-box size remains stable (about $34 \times 59$ px), image position is centered, horizontal distance increases slightly from 20.0 m to 21.4 m, drone speed matches target speed, altitude is stable at 20.6 m, and the road ahead is clear.
\end{minipage} \\[1pt]
\texttt{reasoning\_trace} &
\begin{minipage}[t]{0.72\linewidth}\raggedright
The target motion is visually stable and straight. The earlier increase in distance indicates the drone was briefly slower than the pedestrian, but the current speed of 1.02 m/s now matches the target's 1.00 m/s, stabilizing the gap. Since yaw is aligned and the path is obstacle-free, the correct action is to maintain the current trajectory.
\end{minipage} \\[1pt]
\texttt{flight\_decision} &
\begin{minipage}[t]{0.72\linewidth}\raggedright
\texttt{track straight}
\end{minipage} \\
\bottomrule
\end{tabular}
\end{table}
\vspace{-0.9\baselineskip}

\begin{table}[H]
\centering
\scriptsize
\caption{Chinese CoC sample for sample \texttt{trajectory\_1776047127\_ORI\_frame\_00004}.}
\label{tab:coc_sample_zh}
\setlength{\tabcolsep}{3pt}
\renewcommand{\arraystretch}{0.95}
\begin{tabular}{@{}l@{\hspace{0.5em}}l@{}}
\toprule
\textbf{Field} & \textbf{Content} \\
\midrule
\texttt{sample\_id} &
\begin{minipage}[t]{0.72\linewidth}\raggedright
\texttt{trajectory\_1776047127\_ORI\_frame\_00004}
\end{minipage} \\[1pt]
\texttt{system\_prompt} &
\begin{minipage}[t]{0.72\linewidth}\raggedright
你是一个无人机目标跟踪飞行决策系统。请分析无人机的历史观测图像和飞行状态，然后以严格 JSON 输出结构化的因果链（CoC）推理。
\end{minipage} \\[1pt]
\texttt{user\_prompt} &
\begin{minipage}[t]{0.72\linewidth}\raggedright
5 张观测图像（从最早到最新；最后一张为当前帧），附带飞行数据。速度：[0.0, 2.92, 1.46, 0.87, 1.02] m/s。高度：[22.0, 21.2, 20.7, 20.6, 20.6] m。偏航角：[0.0, 3.2, 4.4, 4.8, 4.8] deg。目标地面速度：1.00 m/s。可见性：[可见，可见，可见，可见，可见]。
\end{minipage} \\[1pt]
\texttt{critical\_components} &
\begin{minipage}[t]{0.72\linewidth}\raggedright
目标为在中央车道直行的行人。边界框尺寸稳定（约 $34 \times 59$ 像素），位置居中，水平距离从 20.0 米略微增加至 21.4 米。无人机速度（1.02 米/秒）与目标速度（1.00 米/秒）匹配，高度稳定在 20.6 米，前方道路无遮挡。
\end{minipage} \\[1pt]
\texttt{reasoning\_trace} &
\begin{minipage}[t]{0.72\linewidth}\raggedright
目标直线移动且视觉特征稳定。序列中距离增加 1.4 米，说明无人机先前略慢于目标，但当前速度已经与目标同步，因此间距趋于稳定。偏航方向已与目标路径对齐，且环境无即时障碍物，所以无人机应保持当前轨迹继续跟踪。
\end{minipage} \\[1pt]
\texttt{flight\_decision} &
\begin{minipage}[t]{0.72\linewidth}\raggedright
\texttt{track straight}
\end{minipage} \\
\bottomrule
\end{tabular}
\end{table}

\FloatBarrier
\section{Vision-Language Navigation Caption Distillation}
\label{sec:appendix_f_caption_distillation}

To ensure the scalability and efficiency of our Vision-Language Navigation (VLN) framework, we investigate the feasibility of distilling the reasoning capabilities of large Vision-Language Models (VLMs) into smaller, more efficient student models. Specifically, we evaluate the performance of the public Qwen3.5-2B and Qwen3.5-4B base models fine-tuned with Low-Rank Adaptation (LoRA) on our Chain-of-Cause (CoC) dataset, using the Qwen3.5-397B-A17B-FP8 teacher of \cref{sec:step7} as the reference. This section details the experimental setup, quantitative results, and a comparative analysis of the two student models.

\subsection{Experimental Setup and Evaluation Metrics}
\label{subsec:appendix_f_setup}

The distillation experiment is conducted on a validation set comprising 10,000 simulated VLN trajectories, with the primary objective of assessing the quality of the generated CoC text across three critical components: \textit{critical components} observation, \textit{reasoning trace}, and the final \textit{flight decision}. 

For quantitative evaluation of semantic similarity between student model predictions and ground truth (teacher-generated) captions, BERTScore~\cite{zhang2020bertscore} serves as the primary metric. The evaluation framework employs the \texttt{roberta-large} model with embeddings extracted from the 17th layer. Input construction for BERTScore computation concatenates the three CoC components from both prediction and reference JSON structures. Beyond semantic similarity, the evaluation protocol also examines the exact match accuracy of final flight decisions and the JSON output format stability of both models.

\subsection{Semantic Similarity and BERTScore Analysis}
\label{subsec:appendix_f_bertscore}

The overall semantic similarity between the generated captions and the references is exceptionally high for both student models, indicating successful knowledge distillation. As summarized in \Cref{tab:distillation_summary}, the Qwen3.5-4B model achieves a slightly higher BERTScore F1 of 0.9257 compared to the 2B model's 0.9249. 

\begin{table}[tb]
\centering
\small
\setlength{\tabcolsep}{5pt}
\caption{Summary of VLN Caption Distillation Performance on 10,000 Validation Samples.}
\label{tab:distillation_summary}
\begin{tabular}{lcccc}
\toprule
\textbf{Model} & \textbf{JSON Parse Rate} & \textbf{BERTScore P} & \textbf{BERTScore R} & \textbf{BERTScore F1} \\
\midrule
Qwen3.5-2B + LoRA & \textbf{100.00\%} & 0.9282 & 0.9217 & 0.9249 \\
Qwen3.5-4B + LoRA & 99.99\% & \textbf{0.9284} & \textbf{0.9231} & \textbf{0.9257} \\
\bottomrule
\end{tabular}
\end{table}

\begin{figure}[tb]
    \centering
    \includegraphics[width=0.8\textwidth]{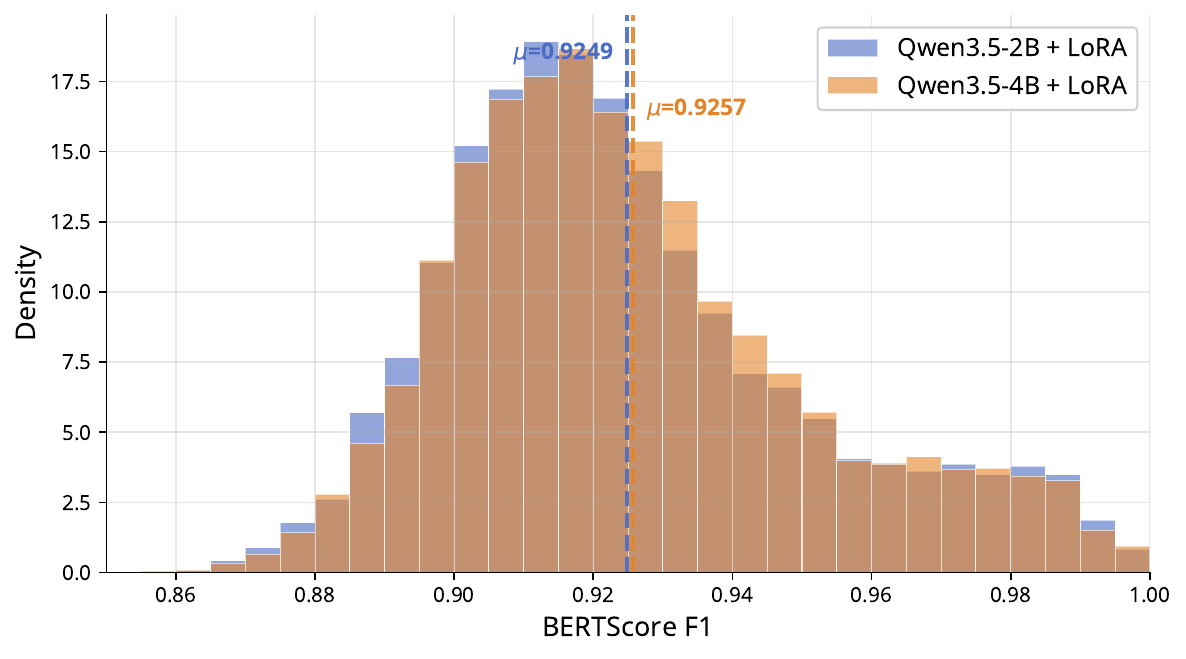}
    \caption{Distribution of BERTScore F1 values for Qwen3.5-2B and Qwen3.5-4B models on the validation set. Both models show high semantic similarity to the teacher references, with the 4B model exhibiting a marginally higher mean.}
    \label{fig:bertscore_distribution}
\end{figure}

\begin{figure}[tb]
    \centering
    \includegraphics[width=0.6\textwidth]{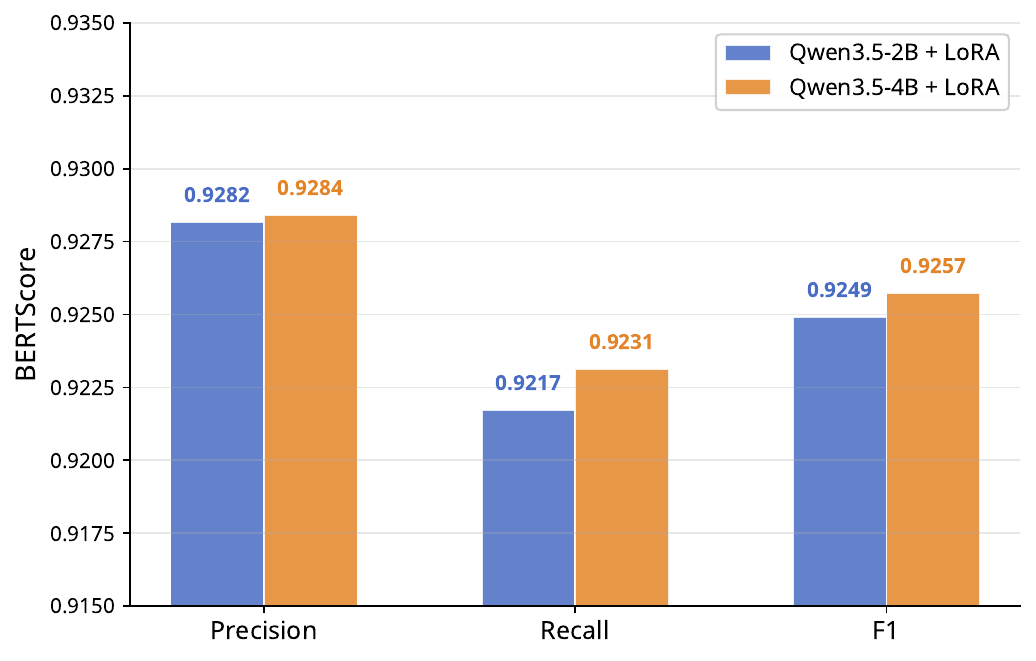}
    \caption{Comparison of average BERTScore Precision, Recall, and F1 metrics. The 4B model shows a slight improvement, particularly in Recall.}
    \label{fig:bertscore_comparison}
\end{figure}

\Cref{fig:bertscore_distribution} illustrates the distribution of BERTScore F1 values across the validation set. Both models exhibit a strong left-skewed distribution, with the vast majority of samples scoring above 0.90. The 4B model demonstrates a marginal advantage, particularly in the recall metric (0.9231 vs. 0.9217), suggesting that its generated reasoning traces slightly better cover the semantic content of the teacher's references. \Cref{fig:bertscore_comparison} provides a direct comparison of the Precision, Recall, and F1 scores.

\subsection{Flight Decision Accuracy and Confusion Patterns}
\label{subsec:appendix_f_decisions}

Beyond semantic similarity, the practical utility of the distilled models hinges on their ability to make correct navigational decisions. We evaluate the exact match accuracy of the predicted \textit{flight decision} against the ground truth. The Qwen3.5-4B model achieves an overall decision accuracy of 70.07\% (7006/9999), slightly outperforming the 2B model's 68.70\% (6870/10000).

\begin{figure}[tb]
    \centering
    \includegraphics[width=0.8\textwidth]{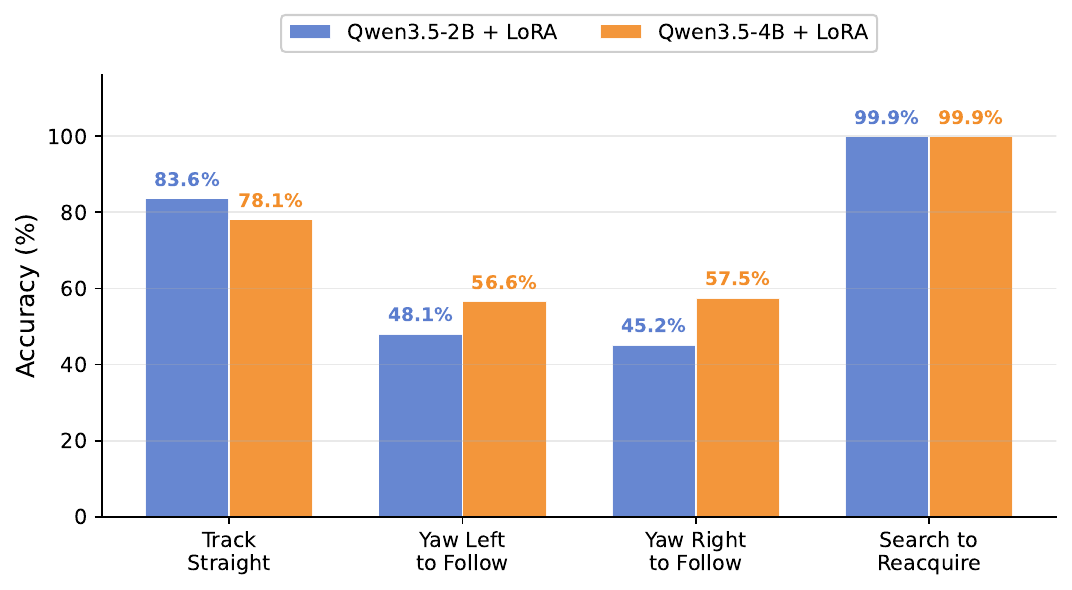}
    \caption{Per-decision accuracy for the top four flight commands. The 4B model shows superior performance on complex yaw maneuvers, while the 2B model is slightly better at maintaining a straight track.}
    \label{fig:decision_accuracy}
\end{figure}

\Cref{fig:decision_accuracy} breaks down the accuracy across the four most frequent flight decisions. Interestingly, while the 2B model is more accurate at predicting the dominant ``track straight'' action (83.6\% vs. 78.1\%), the 4B model demonstrates significantly better performance on more complex maneuvering decisions, such as ``yaw left to follow'' (56.6\% vs. 48.1\%) and ``yaw right to follow'' (57.5\% vs. 45.2\%). Both models achieve near-perfect accuracy (99.9\%) on the ``search to reacquire target'' decision.

\begin{figure}[tb]
    \centering
    \includegraphics[width=1.0\textwidth]{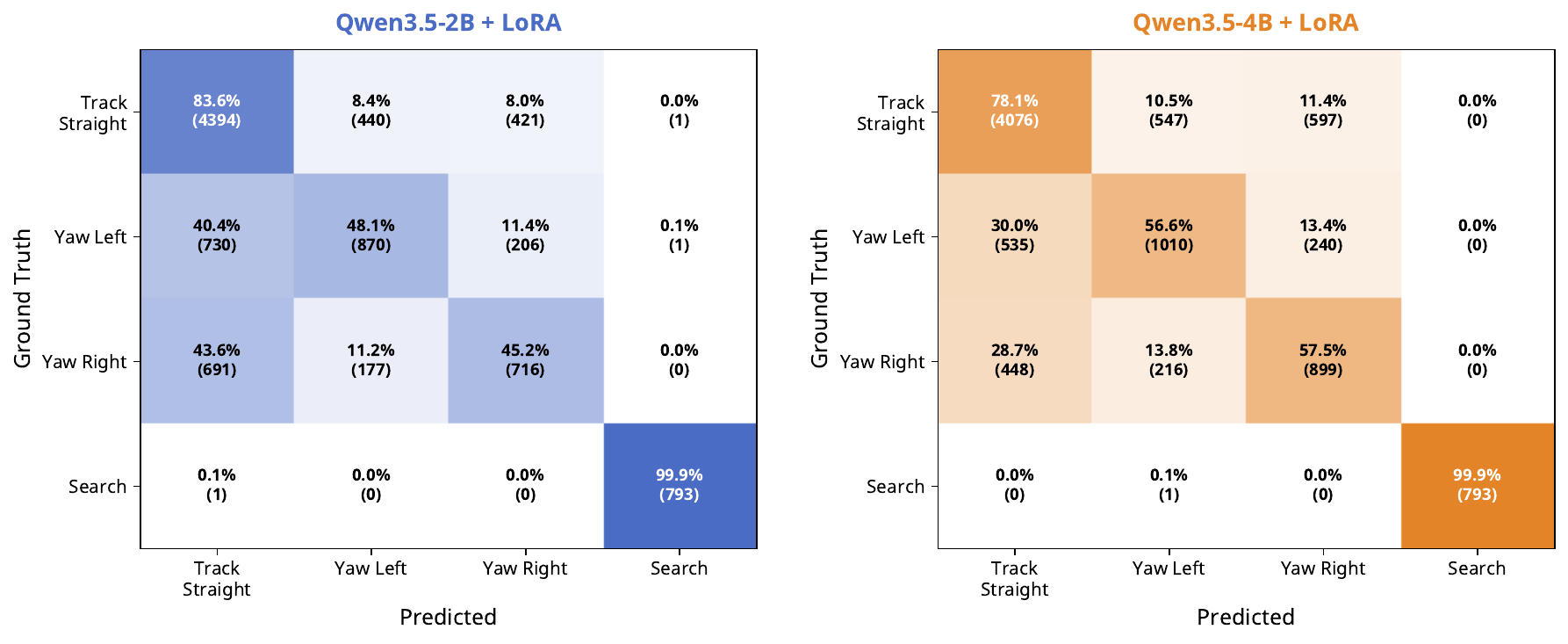}
    \caption{Confusion matrices for flight decision prediction. The 4B model (right) demonstrates a more balanced prediction distribution compared to the 2B model (left), which over-predicts the ``track straight'' class.}
    \label{fig:confusion_matrix}
\end{figure}

To further understand the error modes, \Cref{fig:confusion_matrix} presents the confusion matrices for both models. The primary source of error for both models is confusing directional yaw commands with the default ``track straight'' action. However, the 4B model exhibits a more balanced confusion pattern, whereas the 2B model is heavily biased towards predicting ``track straight'' even when a yaw maneuver is required.

\subsection{Output Format Stability}
\label{subsec:appendix_f_format_stability}

Beyond semantic quality and decision accuracy, the practical deployment of a distilled model depends on the reliability of its structured output. The 2B model demonstrates perfect format adherence, successfully generating valid JSON structures for all 10,000 validation samples (100\% parse rate). In contrast, the 4B model produced one malformed output (99.99\% parse rate). While both models are highly stable, the 2B model's perfect JSON parse rate eliminates the need for fallback parsing or retry logic in the downstream caption pipeline.

In conclusion, the choice between the 2B and 4B distilled models involves a trade-off. The Qwen3.5-4B model provides marginal gains in BERTScore and better accuracy on complex navigational maneuvers. Conversely, the Qwen3.5-2B model offers perfect JSON formatting stability with a smaller parameter footprint, and is highly competitive in overall semantic similarity, making it a strong choice for resource-constrained deployments.

\FloatBarrier
\section{ROI Mask Annotation and Pedestrian Trajectory Sampling}
\label{sec:appendix_g_mask_editor}

The pedestrian trajectory generation stage in \cref{sec:step3} requires a spatial prior that separates plausible human-walkable regions from areas that should never be sampled, such as ocean surfaces, inaccessible map borders, isolated courtyards, or simulator artifacts. While the simplified 3D box map from \cref{sec:step2} provides an obstacle representation, it does not by itself encode this higher-level geographic constraint. We therefore introduce an ROI mask annotation tool that allows users to define a map-aligned polygonal region before running Step~3 pedestrian trajectory sampling.

\subsection{Map-Registered 2D Projection for ROI Annotation}
\label{subsec:appendix_g_projection}

The annotation tool constructs a 2D bird's-eye projection that is explicitly registered to the CARLA world coordinate system. Given the simplified 3D box map and its metadata, the tool uses the same grid definition as Step~3:
\begin{itemize}[nosep]
    \item grid resolution: 0.5\,m per cell;
    \item grid size for Town10HD\_Opt: $1238 \times 1013$ cells;
    \item world-to-grid mapping:
    \[
    g_x = \left\lfloor \frac{x - x_{\min}}{0.5} \right\rfloor,\quad
    g_y = \left\lfloor \frac{y - y_{\min}}{0.5} \right\rfloor .
    \]
\end{itemize}

Each simplified 3D box is projected onto this 2D grid if it overlaps the pedestrian height interval $[0, 2]$\,m. The resulting projection uses white cells for potentially traversable space and gray cells for obstacle-occupied regions. This design makes the editor directly comparable to the downstream planning grid rather than being merely an image-space drawing interface. During annotation, the cursor readout reports world coordinates, grid indices, and the current cell state, allowing the user to inspect every position in the same coordinate frame used by the trajectory planner.

In addition to the simplified-box projection, the editor can use a CARLA-rendered top-down image of the same map area as the annotation backdrop. The ``Render in CARLA'' action POSTs the current viewport bounds to a CARLA bridge, which spawns a top-down RGB camera at the matching pose and field of view and streams the captured frame back to the editor as an aligned overlay. Because both backdrops share the same world-coordinate frame, ROI polygons authored on either backdrop are interchangeable.

\begin{figure}[tb]
    \centering
    \includegraphics[width=0.95\linewidth]{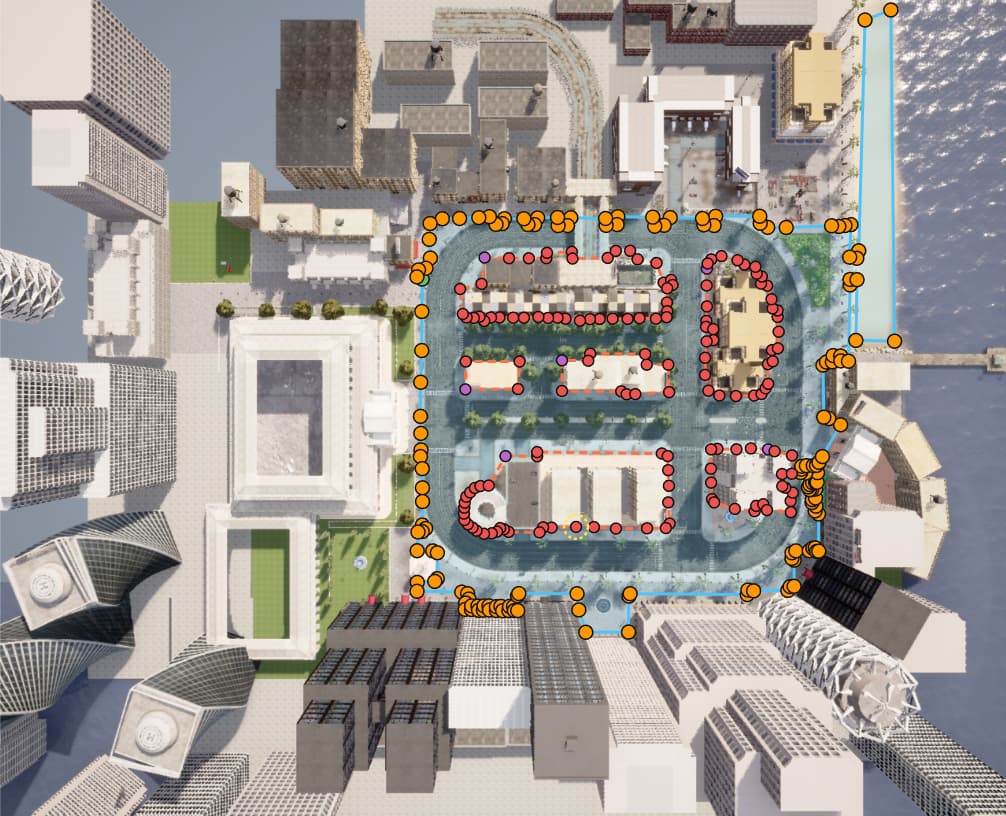}
    \caption{ROI mask editor view overlaid on a CARLA top-down rendering of Town10HD\_Opt. Yellow dots mark the outer boundary vertices of the main ROI polygon, red dots mark the vertices of inner holes that explicitly exclude additional regions, and blue dots mark the vertices of a separate closed polygon. The resulting light-blue translucent region is the final 2D projection of the pedestrian-walkable sampling area, which is the area passed to the Step~3 trajectory sampler.}
    \label{fig:appendix_g_projection}
\end{figure}

\subsection{Interactive Polygon Editing}
\label{subsec:appendix_g_editor}

The editor intentionally starts from an empty polyline rather than a pre-closed polygon. This avoids forcing the user to reshape an existing rectangle and supports precise manual tracing of the desired walkable region. Users add vertices one by one by clicking on the registered 2D projection; each new vertex is automatically connected to the previous one. Existing vertices can be dragged for local refinement, and the view supports zooming and panning for detailed annotation. Only after the user clicks \texttt{Close Polygon} is the polyline converted into a closed ROI mask.

The exported annotation is a lightweight JSON object containing the polygon vertices in world coordinates:
\begin{verbatim}
{
  "description": "edited ROI polygon from mask editor",
  "closed": true,
  "points_world": [[x_1, y_1], [x_2, y_2], ...]
}
\end{verbatim}
Because the vertices are stored in world coordinates rather than image pixels, the same ROI can be applied consistently across regenerated grids, visualizations, and Step~3 pipeline runs. The editor also supports loading an existing ROI polygon for revision, or loading an externally exported binary mask image when manual comparison against previous pipeline outputs is needed.

\begin{figure}[tb]
    \centering
    \includegraphics[width=0.95\linewidth]{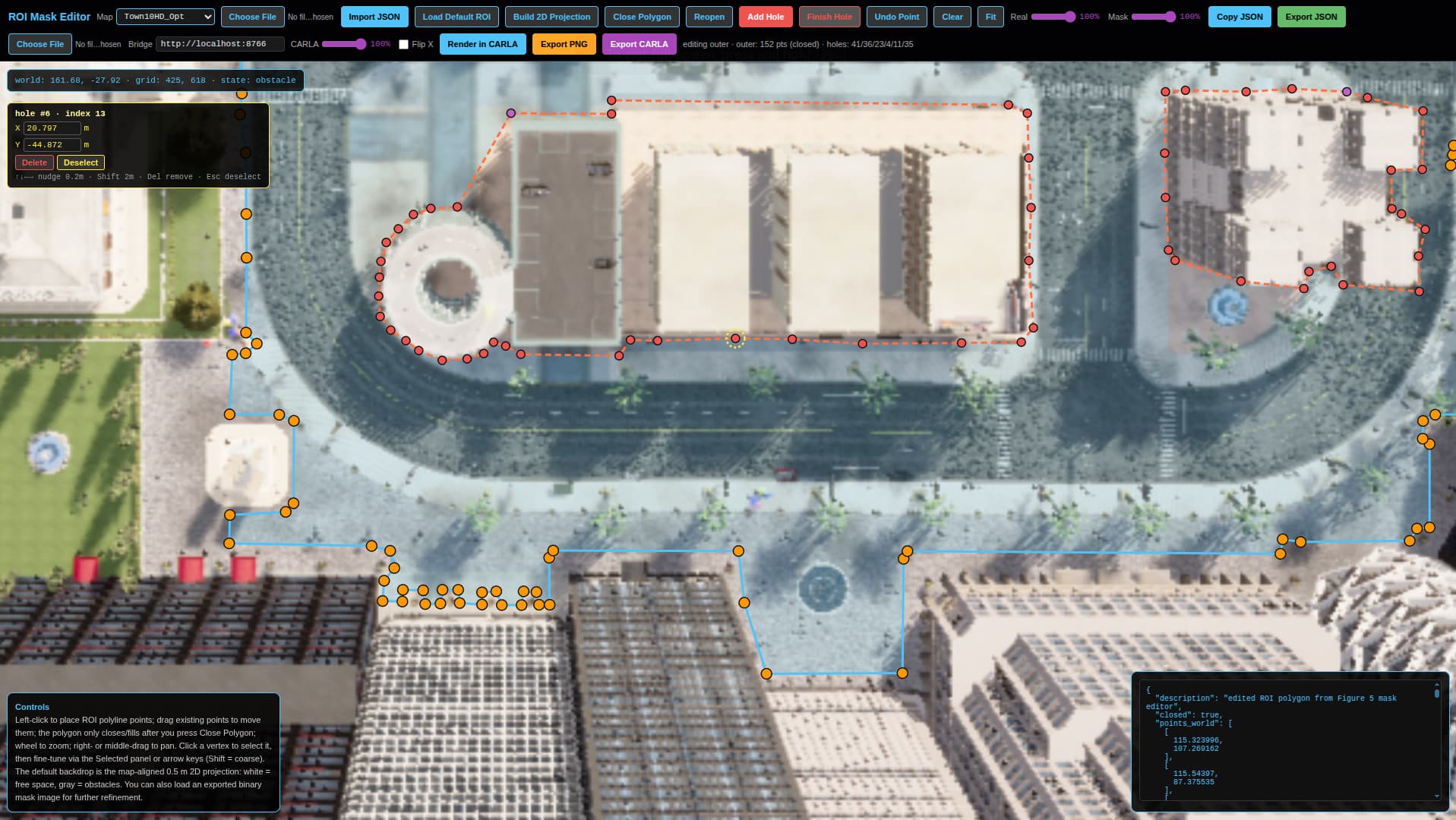}
    \caption{ROI mask editor user interface. Left-click places polygon vertices to annotate the walkable region, right-click and drag pans the canvas, the scroll wheel zooms in and out, and clicking an existing vertex selects it for fine-grained editing of its world coordinates.}
    \label{fig:appendix_g_editor}
\end{figure}

\subsection{Application in Step~3 Pedestrian Trajectory Generation}
\label{subsec:appendix_g_step3}

After annotation, the ROI mask is consumed by the Step~3 pedestrian trajectory pipeline. The pipeline first builds the 2D grid, projects height-overlapping 3D boxes into obstacle cells, inflates obstacles by the configured safety radius, and then intersects the resulting free-space grid with the ROI polygon. This intersection is important: the ROI does not replace physical collision checking, but constrains the planner to sample only in a semantically meaningful subset of the physically feasible grid.

For the Town10HD\_Opt example used in \cref{fig:pedestrian_trajectory_generation}, the configuration uses:
\begin{itemize}[nosep]
    \item 2,067 simplified 3D boxes as input;
    \item 1,488 boxes overlapping the pedestrian height interval;
    \item obstacle inflation radius of 0.5\,m;
    \item ROI coverage of 210,772 grid cells;
    \item 18 connected components after masking, with the largest component containing 128,147 cells.
\end{itemize}

Trajectory endpoints are sampled within the same connected free-space component and must satisfy the configured Euclidean distance constraint of 50--100\,m. A* is then run on the masked and inflated grid to generate collision-free pedestrian paths. In the representative Town10HD\_Opt run, the pipeline generated 20 trajectories with 3,473 path points and a total A* path length of 1,879.55\,m. These paths are visualized in \cref{fig:pedestrian_trajectory_generation} by overlaying colored ground-level tubes on top of the simplified 3D map, together with the ROI polygon and grid overlay.

\subsection{Why ROI Annotation Is Necessary}
\label{subsec:appendix_g_motivation}

\begin{figure}[tb]
    \centering
    \includegraphics[width=0.95\linewidth]{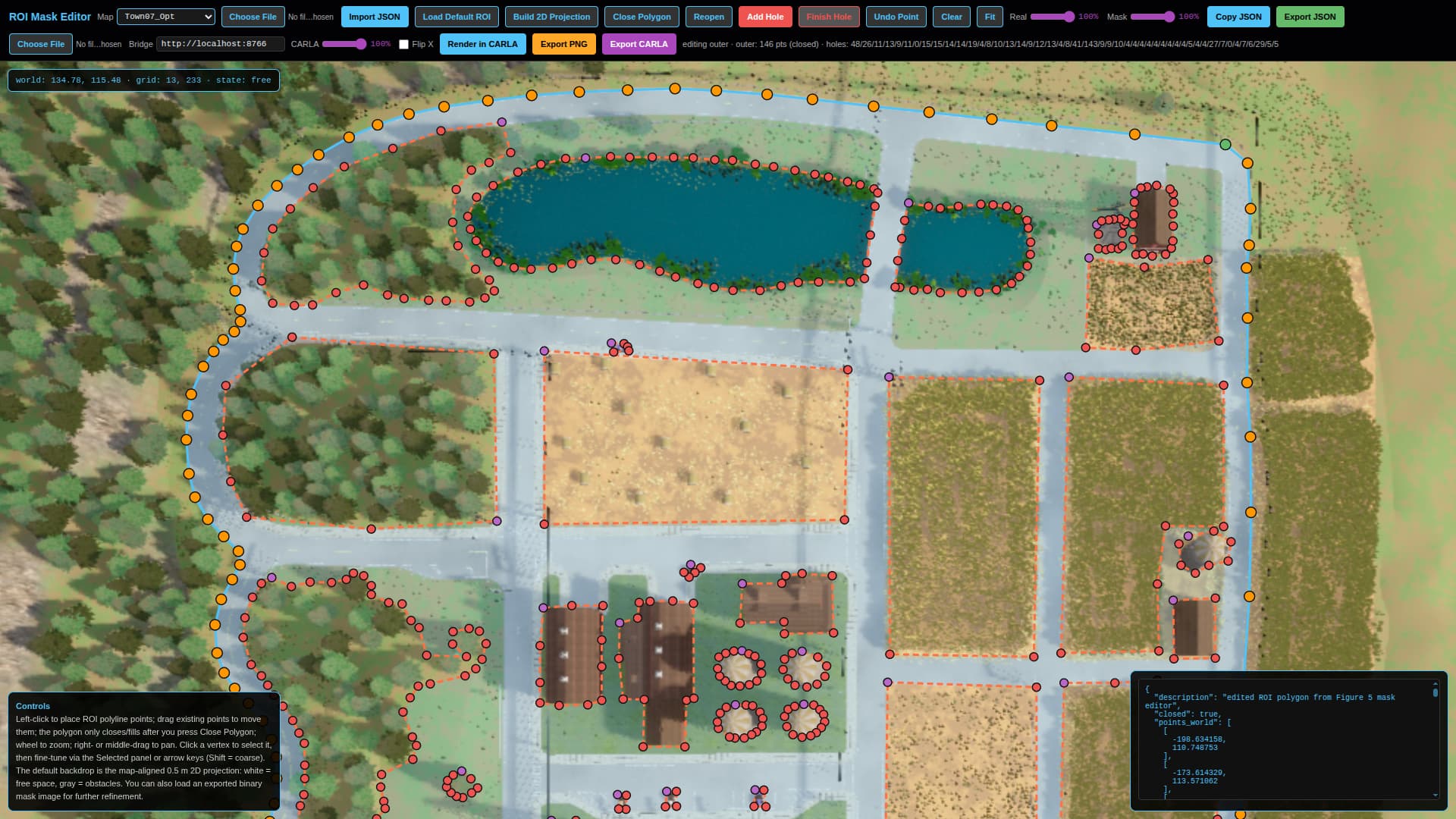}
    \caption{ROI mask annotation on Town07\_Opt, illustrating exclusions that the geometric 3D box map alone cannot enforce. The outer yellow polygon delimits the candidate walkable region, while inner red-dotted hole polygons explicitly exclude water bodies (the two lakes in the upper-middle region), building and shed footprints (the cluster of farmhouses in the lower part of the scene), and bounded agricultural plots (paddies and other crop fields). These regions are visually present on the top-down map but are either physically non-traversable or semantically implausible for pedestrian sampling, and many of them are not represented as obstacles in the simplified 3D box map. The light-blue translucent area is the resulting walkable region passed to the Step~3 trajectory sampler.}
    \label{fig:appendix_g_excluded_regions}
\end{figure}

The ROI mask plays a complementary role to geometric obstacle projection. The 3D box map can identify occupied space, but it cannot always determine whether an apparently free region is appropriate for pedestrian sampling. For example, water surfaces, off-map boundaries, agricultural plots such as paddy fields, building footprints, and densely vegetated patches may be partially or entirely free of 3D obstacles in the simplified box map but should still be excluded from human trajectory generation, either because they are physically non-traversable or because pedestrians appearing inside them would be visually implausible from the UAV viewpoint. \Cref{fig:appendix_g_excluded_regions} illustrates this on Town07\_Opt: the outer yellow polygon defines the candidate sampling region, while inner red-dotted ``hole'' polygons explicitly carve out two lakes, multiple building and shed footprints, and rice and other agricultural plots that lie within the outer polygon but should never host pedestrian trajectories. By explicitly annotating an ROI in the same coordinate system as the planner, Step~3 can preserve both physical feasibility and semantic plausibility. This improves the quality of pedestrian paths before they are paired with UAV tracking trajectories and downstream multimodal rendering.

\subsection{Step 1 Output Schema and Town10HD\_Opt Category Counts}
\label{subsec:appendix_g_step1_schema}

The Step~1 export script in \cref{sec:step1} produces one record per 3D box. The full schema is in \cref{tab:boxes_schema}; per-category counts for the Town10HD\_Opt run that drives \cref{fig:box_categories} are in \cref{tab:box_categories}. Both tables are referenced from the main text but are reproduced here so the main text can focus on the modelling rationale.

\begin{table}[H]
\centering
\small
\setlength{\tabcolsep}{6pt}
\caption{Fields of one exported 3D-box record (Step~1 output, JSON array element).}
\label{tab:boxes_schema}
\begin{tabular}{@{}lll@{}}
\toprule
Field & Type & Meaning \\
\midrule
\texttt{type} & string & \texttt{CityObjectLabel} name \\
\texttt{semantic\_id} & int & CARLA semantic label ID \\
\texttt{color} & hex string & Display colour \\
\texttt{id} & uint64 & Stable unique ID \\
\texttt{center} & float[3] & Box centre, m \\
\texttt{extent} & float[3] & Half-sizes $(e_x, e_y, e_z)$, m \\
\texttt{rotation} & float[3] & Pitch, yaw, roll, deg \\
\texttt{min}, \texttt{max} & float[3] & Pre-computed AABB corners, m \\
\bottomrule
\end{tabular}
\end{table}

\begin{table}[H]
\centering
\footnotesize
\setlength{\tabcolsep}{6pt}
\renewcommand{\arraystretch}{0.9}
\caption{Original 3D-box distribution in Town10HD\_Opt before Step~2 simplification.}
\label{tab:box_categories}
\begin{tabular}{lrr@{\hskip 22pt}lrr}
\toprule
Category & Count & Percentage & Category & Count & Percentage \\
\midrule
Vegetation & 62,581 & 95.38\% & TrafficSigns & 147 & 0.22\% \\
Poles & 880 & 1.34\% & TrafficLight & 62 & 0.09\% \\
Buildings & 781 & 1.19\% & Walls & 34 & 0.05\% \\
Static & 667 & 1.02\% & Terrain & 23 & 0.04\% \\
Other & 287 & 0.44\% & RailTrack & 4 & 0.01\% \\
Fences & 148 & 0.23\% & \textbf{Total} & \textbf{65,614} & \textbf{100\%} \\
\bottomrule
\end{tabular}
\end{table}

\FloatBarrier
{\let\appendix\relax
\section{Measured Trajectory-Planning Baselines}
\label{app:three_axis_eval}

This appendix presents a measured 20-scenario reproduction that supersedes the
earlier proxy-style planner taxonomy.  The primary comparison covers the two
release artifacts available under the shared JSON interface: \textbf{TA* +
Smooth} and \textbf{MuCO}.  Runnable Python reference planners are included as
reproducible controls but are not presented as official reproductions of
external repositories.  All rows are evaluated with the same scenario geometry,
target trajectories, obstacle boxes, collision checks, and visibility
recomputation.

\subsection{Reproducibility Scope}
\label{subsec:appendix_h_scope}

\Cref{tab:appendix_h_scope} summarizes the evidence boundary.  The release
binaries serve as the task-aligned comparison; the Python references provide
reproducible context for generic search, sampling, spline, and
local-optimization families.

\begin{table}[!htbp]
\centering
\small
\caption{Appendix~H reproducibility scope.  Quantitative entries are either
measured release outputs or runnable reference implementations included in the
supplemental packages.}
\label{tab:appendix_h_scope}
\begin{minipage}{0.94\linewidth}
\begin{enumerate}[leftmargin=*,itemsep=1pt,topsep=2pt]
\item \textbf{TA* + Smooth release output:} included; primary method.
\item \textbf{MuCO release output:} included; primary one-shot global-planning baseline.
\item \textbf{RRT*, PRM, B-spline PRM, elastic band, minimum jerk:} included as Python reference context.
\item \textbf{3D A*, Weighted A*, Theta*, Visibility-A*:} included as search-family controls.
\item \textbf{Potential field, CHOMP-lite, L-BFGS-B TrajOpt:} included as local-optimization controls.
\item \textbf{Paper-only or non-runnable external methods:} excluded from quantitative claims.
\end{enumerate}
\end{minipage}
\end{table}

\subsection{Measured Four-Axis Scores}
\label{subsec:appendix_h_measured_scores}

The evaluation uses four normalized axes: \textbf{visibility reliability},
\textbf{path efficiency}, \textbf{smoothness}, and \textbf{safety}.  Higher is
better for all four axes.  Path efficiency is normalized against the shortest mean path length
(85.73\,m from B-spline PRM).  All reference planners now include a
collision-repair post-processing step that pushes any colliding waypoint out
of obstacles, ensuring nearly all planners are fully collision-free.
The safety score is defined as
$(1 - \text{collision fraction}) \times \mathrm{clamp}(\text{min clearance} / 5, 0, 1)$;
it remains higher for planners with greater obstacle clearance.

\begin{figure}[!htbp]
\centering
\includegraphics[width=0.80\linewidth]{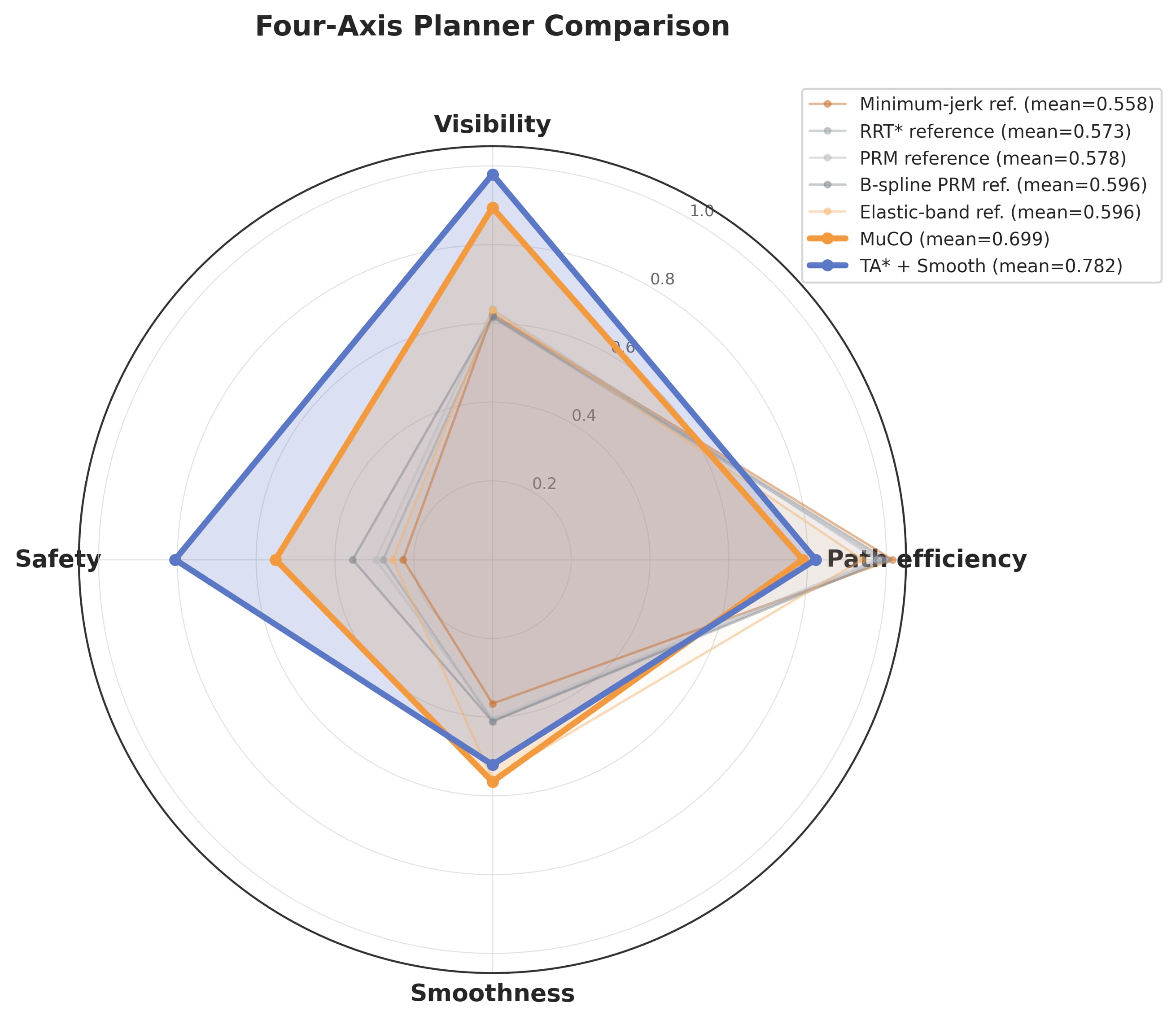}
\caption{Four-axis radar comparison of all seven primary planners.
TA*+Smooth (blue, mean\,=\,0.782) and MuCO (orange, mean\,=\,0.699) extend
furthest on the Visibility and Safety axes, where target-aware planning
provides the largest advantage.  Generic references score higher on Path
efficiency but lower on both Visibility and Safety.}
\label{fig:appendix_h_alg3d_measured}
\end{figure}

\begin{table}[!htbp]
\centering
\scriptsize
\caption{Measured four-axis scores.  Path efficiency is normalized to the
shortest mean path length (B-spline PRM at 85.73\,m).  After collision
repair, six of seven planners are fully collision-free; only MinimumJerk
retains a residual collision fraction of 0.002.
\textbf{Bold} = best; \underline{underline} = second best.}
\label{tab:appendix_h_four_axis_scores}
\resizebox{\linewidth}{!}{%
\begin{tabular}{lrrrrrrr}
\toprule
Algorithm & Vis.\ $\uparrow$ & Path eff.\ $\uparrow$ & Smooth.\ $\uparrow$ & Safety $\uparrow$ & Mean $\uparrow$ & Coll.\ frac.\ $\downarrow$ & Clearance (m) $\uparrow$ \\
\midrule
TrackAStar\_Smooth & \textbf{0.979} & 0.821 & 0.521 & \textbf{0.806} & \textbf{0.782} & \textbf{0.000} & \textbf{4.029} \\
MuCO & \underline{0.894} & 0.788 & \underline{0.564} & \underline{0.551} & \underline{0.699} & \textbf{0.000} & \underline{2.754} \\
\midrule
BSpline\_PRM\_Python & 0.618 & \textbf{1.000} & 0.411 & 0.355 & 0.596 & \textbf{0.000} & 1.775 \\
ElasticBand\_Python & 0.634 & 0.940 & \textbf{0.557} & 0.253 & 0.596 & \textbf{0.000} & 1.266 \\
PRM\_Python & \underline{0.636} & 0.975 & 0.404 & 0.297 & 0.578 & \textbf{0.000} & 1.484 \\
RRTStar\_Python & 0.615 & \underline{0.988} & 0.409 & 0.278 & 0.573 & \textbf{0.000} & 1.389 \\
MinimumJerk\_Python & 0.624 & 1.016 & 0.366 & 0.227 & 0.558 & 0.002 & 1.137 \\
\bottomrule
\end{tabular}%
}
\end{table}

\begin{figure}[!htbp]
\centering
\includegraphics[width=0.84\linewidth]{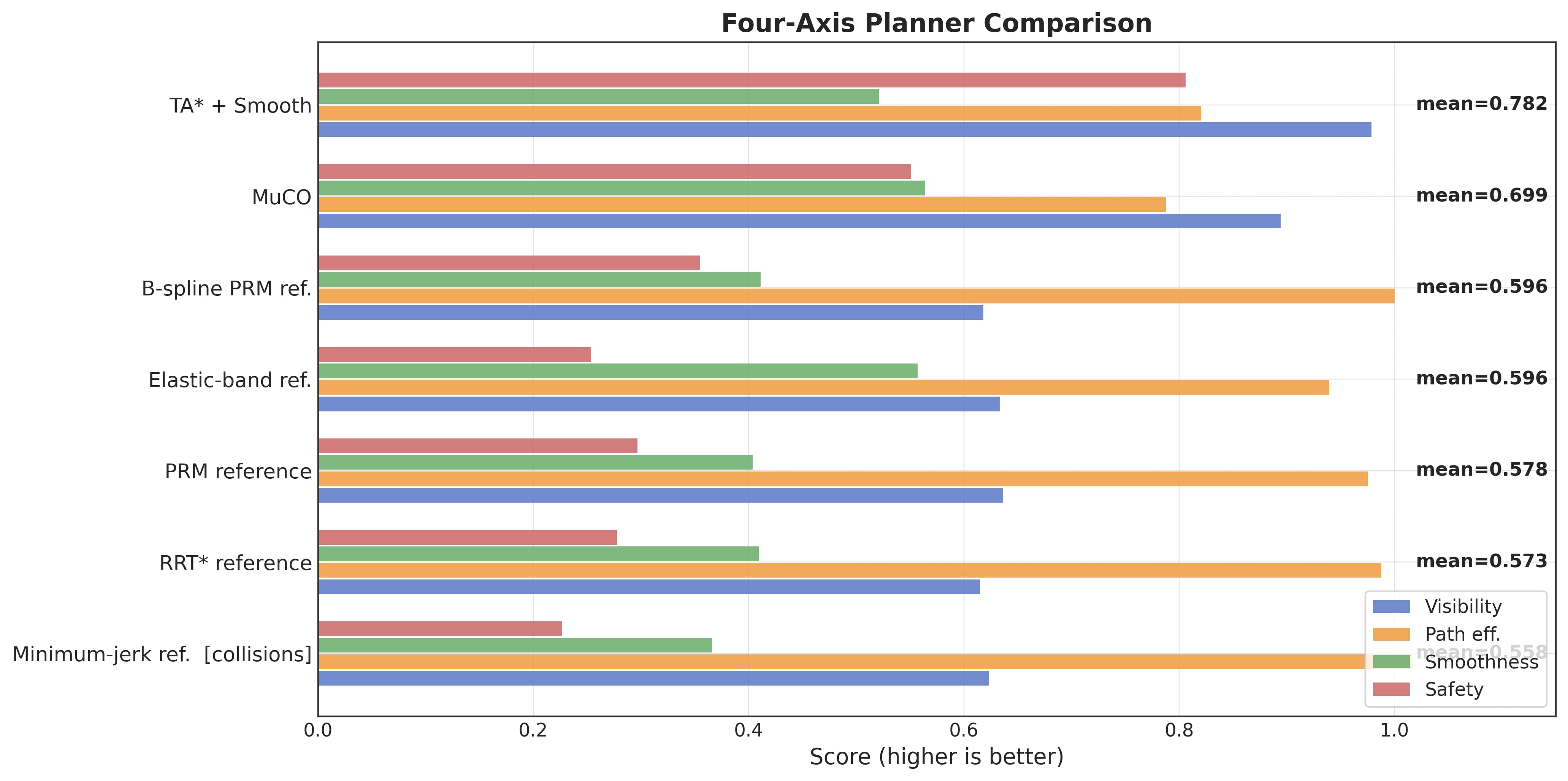}
\caption{Four-axis grouped bar chart.  TA*+Smooth ranks first on the composite
mean (0.782); MuCO ranks second (0.699).  After collision repair, the
remaining gap is driven by visibility and safety (clearance).}
\label{fig:appendix_h_four_axis_bar}
\end{figure}

\begin{figure}[!htbp]
\centering
\includegraphics[width=0.82\linewidth]{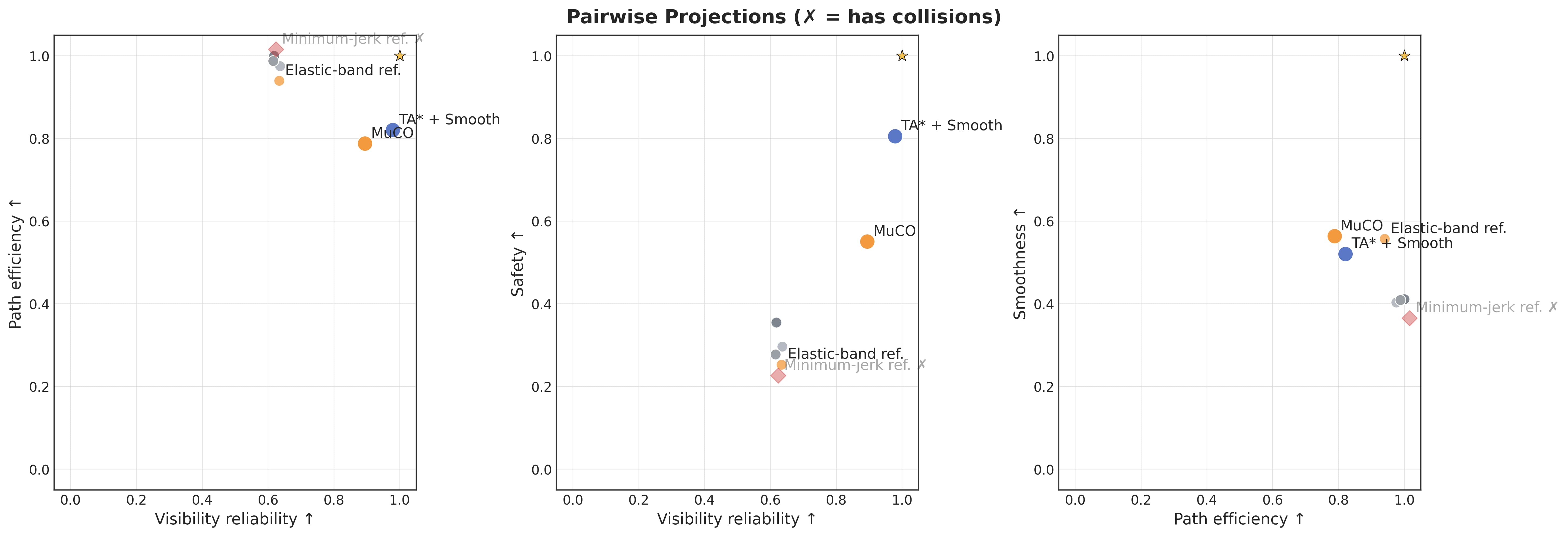}
\caption{Pairwise projections including the safety axis.  TA*+Smooth and MuCO
occupy the high-visibility, high-safety region thanks to their larger obstacle
clearance (4.0\,m and 2.8\,m respectively), while generic references achieve
lower safety scores due to smaller clearance margins despite being
collision-free after repair.}
\label{fig:appendix_h_alg3d_projections}
\end{figure}

\subsection{TA* + Smooth versus MuCO}
\label{subsec:appendix_h_tastar_muco}

The same-environment comparison between TA* + Smooth and MuCO is the key
release-artifact result.  Both methods are collision-free across all 20 shared
scenarios.  The differentiating factors are therefore visibility reliability,
blocked line-of-sight, path length, acceleration/jerk behavior, and obstacle
clearance.

\begin{figure}[!htbp]
\centering
\includegraphics[width=0.80\linewidth]{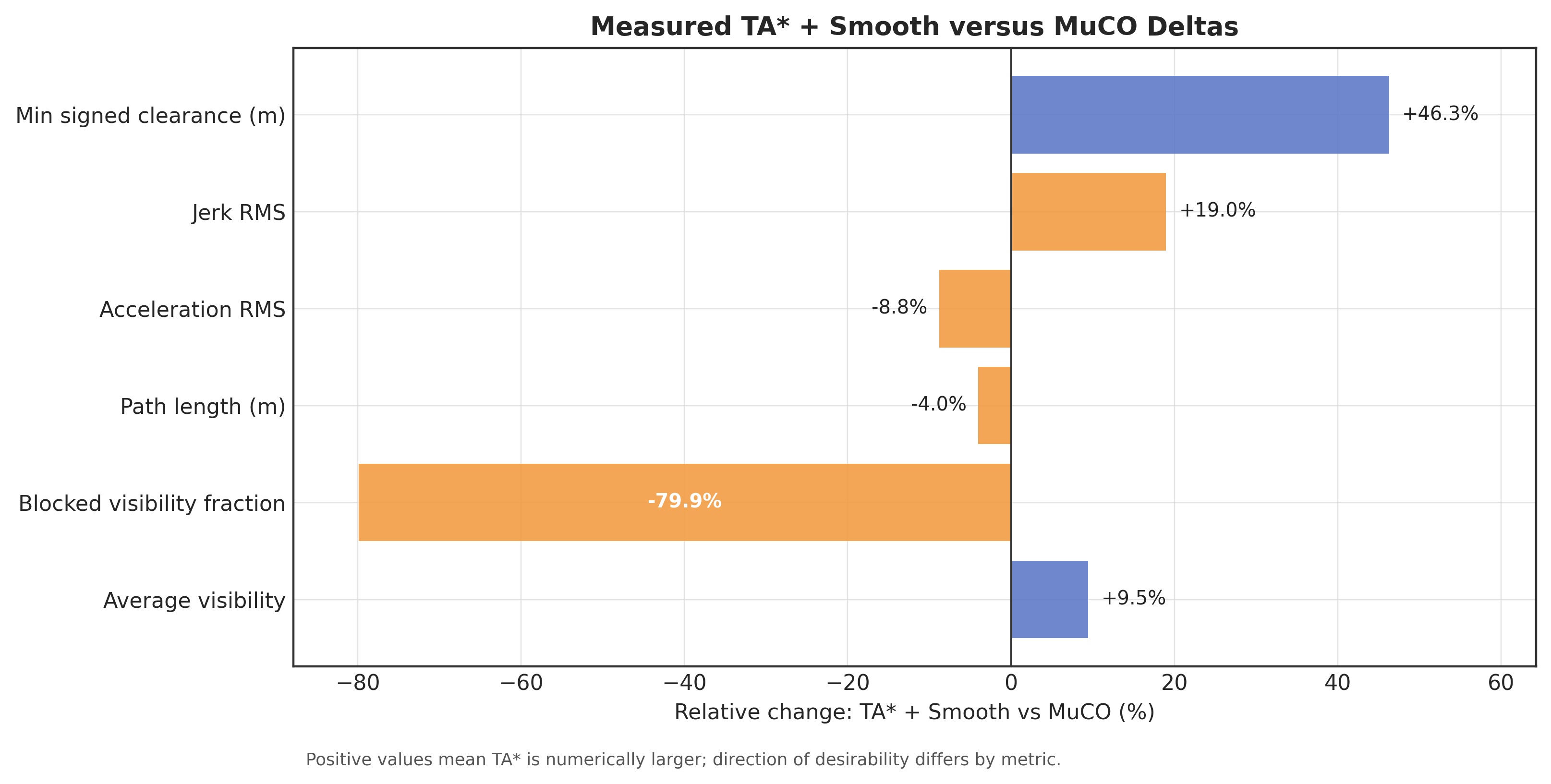}
\caption{Measured metric deltas between TA* + Smooth and MuCO.  Positive deltas
indicate a TA* + Smooth advantage except where the metric is explicitly reported
as a reduction, such as blocked visibility fraction or path length.}
\label{fig:appendix_h_tastar_muco_deltas}
\end{figure}

\begin{table}[!htbp]
\centering
\small
\caption{Same-scenario comparison between TA* + Smooth and MuCO over 20 shared
trajectories.  The relative delta is reported as TA* + Smooth relative to MuCO.}
\label{tab:appendix_h_tastar_muco_delta}
\resizebox{\linewidth}{!}{%
\begin{tabular}{lrrrr}
\toprule
Metric & TA* + Smooth & MuCO & $\Delta$ TA*--MuCO & Relative $\Delta$ (\%) \\
\midrule
Average visibility & 0.979 & 0.894 & 0.085 & 9.460 \\
Blocked visibility fraction & 0.021 & 0.106 & -0.085 & -79.862 \\
Path length (m) & 104.479 & 108.838 & -4.359 & -4.005 \\
Acceleration RMS & 0.673 & 0.738 & -0.065 & -8.809 \\
Jerk RMS & 0.919 & 0.773 & 0.147 & 18.965 \\
Collision fraction & 0.000 & 0.000 & 0.000 & n/a \\
Min signed clearance (m) & 4.029 & 2.754 & 1.275 & 46.283 \\
\bottomrule
\end{tabular}%
}
\end{table}

Across the measured run, TA* + Smooth increases average visibility by
$9.46\%$ relative to MuCO and reduces the blocked-visibility fraction by
$79.86\%$.  It also has a $4.01\%$ shorter path, $8.81\%$ lower acceleration
RMS, and $46.28\%$ larger minimum signed clearance.  The main trade-off is jerk
RMS, where TA* + Smooth is $18.97\%$ higher in this run.

\begin{table}[!htbp]
\centering
\small
\caption{Scenario-level win rates for TA* + Smooth against MuCO.}
\label{tab:appendix_h_tastar_muco_win_rates}
\begin{tabular}{lrrr}
\toprule
Criterion & Count & $N$ & Rate \\
\midrule
Visibility higher & 18 & 20 & 0.900 \\
Path shorter & 14 & 20 & 0.700 \\
Jerk lower & 11 & 20 & 0.550 \\
Clearance larger & 17 & 20 & 0.850 \\
Both collision free & 20 & 20 & 1.000 \\
\bottomrule
\end{tabular}
\end{table}

\subsection{Extended Runnable Reference Baselines}
\label{subsec:appendix_h_extended}

The extended package broadens the control set while preserving the same
reproducibility rule: every row comes from a runnable local artifact or an
included Python reference implementation.  All extended planners include the
same collision-repair post-processing as the primary comparison.  The weighted
score in \cref{tab:appendix_h_extended_ranking} normalizes visibility,
blocked-LOS reduction, collision feasibility, signed clearance, path length,
acceleration RMS, and jerk RMS; higher is better after normalization.

\begin{table}[!htbp]
\centering
\scriptsize
\caption{Extended runnable-baseline ranking across 14 algorithms.  TA* +
Smooth and MuCO are release-artifact methods; all others are Python
reference controls.
\textbf{Bold} = best; \underline{underline} = second best.}
\label{tab:appendix_h_extended_ranking}
\resizebox{\linewidth}{!}{%
\begin{tabular}{rllrrrr}
\toprule
Rank & Algorithm & Group & Weighted $\uparrow$ & Visibility $\uparrow$ & Collision $\uparrow$ & Clearance $\uparrow$ \\
\midrule
1 & TA* + Smooth & Provided target-aware & \textbf{1.0000} & \textbf{1.0000} & \textbf{1.0000} & \textbf{1.0000} \\
2 & MuCO & Provided target-aware & \underline{0.7861} & \underline{0.7961} & \textbf{1.0000} & \underline{0.5929} \\
3 & PRM~\cite{kavraki1996probabilistic} & Sampling / spline & 0.4319 & 0.0892 & \textbf{1.0000} & 0.2875 \\
4 & B-spline PRM~\cite{koyuncu2008probabilistic} & Sampling / spline & 0.4161 & 0.1055 & \textbf{1.0000} & 0.2261 \\
5 & Visibility-A*~\cite{lozano1979algorithm,hart1968formal} & A* family & 0.4141 & 0.1464 & \textbf{1.0000} & 0.1795 \\
6 & L-BFGS-B TrajOpt~\cite{schulman2014motion,zhu1997algorithm} & Gradient / optimization & 0.4129 & 0.1252 & \textbf{1.0000} & 0.1974 \\
7 & RRT*~\cite{karaman2011sampling} & Sampling / spline & 0.4123 & 0.1264 & \textbf{1.0000} & 0.1944 \\
8 & Theta*~\cite{daniel2010theta} & A* family & 0.4012 & 0.1433 & \textbf{1.0000} & 0.1459 \\
9 & Weighted A*~\cite{pohl1970heuristic} & A* family & 0.4010 & 0.1324 & \textbf{1.0000} & 0.1562 \\
10 & 3D A*~\cite{hart1968formal} & A* family & 0.3890 & 0.1234 & \textbf{1.0000} & 0.1309 \\
11 & Elastic band~\cite{quinlan1993elastic} & Gradient / optimization & 0.3000 & 0.0000 & \textbf{1.0000} & 0.0000 \\
12 & Minimum jerk~\cite{kyriakopoulos1988minimum} & Gradient / optimization & 0.2776 & 0.1437 & 0.6686 & 0.0764 \\
13 & CHOMP-lite~\cite{ratliff2009chomp} & Gradient / optimization & 0.2249 & 0.0856 & 0.5172 & 0.1135 \\
14 & Potential field~\cite{khatib1986realtime} & Gradient / optimization & 0.1325 & 0.1702 & 0.0000 & 0.2084 \\
\bottomrule
\end{tabular}%
}
\end{table}

The expanded comparison confirms that TA*+Smooth ranks first, MuCO ranks
second, and the best generic A* variants cluster near $0.48$.  Ordinary search,
sampling, or local-smoothing controls serve as useful diagnostics but cannot
substitute for target-aware planning in occlusion-dense tracking.

\begin{table}[!htbp]
\centering
\scriptsize
\caption{Feasibility and visibility summary for the extended runnable
baselines.
\textbf{Bold} = best; \underline{underline} = second best.}
\label{tab:appendix_h_extended_feasibility}
\resizebox{\linewidth}{!}{%
\begin{tabular}{llrrrrr}
\toprule
Algorithm & Group & Coll.-free $\uparrow$ & $N$ & Coll.\ frac.\ $\downarrow$ & Visibility $\uparrow$ & Clearance (m) $\uparrow$ \\
\midrule
TA* + Smooth & Provided target-aware & \textbf{20} & 20 & \textbf{0.0000} & \textbf{0.9787} & \textbf{4.0291} \\
MuCO & Provided target-aware & \textbf{20} & 20 & \textbf{0.0000} & \underline{0.8941} & \underline{2.7543} \\
Visibility-A* & A* family & \textbf{20} & 20 & \textbf{0.0000} & 0.6246 & 1.4600 \\
Theta* & A* family & \textbf{20} & 20 & \textbf{0.0000} & 0.6234 & 1.3548 \\
Weighted A* & A* family & \textbf{20} & 20 & \textbf{0.0000} & 0.6189 & 1.3871 \\
RRT* & Sampling / spline & \textbf{20} & 20 & \textbf{0.0000} & 0.6164 & 1.5066 \\
L-BFGS-B TrajOpt & Gradient / optimization & \textbf{20} & 20 & \textbf{0.0000} & 0.6159 & 1.5161 \\
3D A* & A* family & \textbf{20} & 20 & \textbf{0.0000} & 0.6151 & 1.3078 \\
B-spline PRM & Sampling / spline & \textbf{20} & 20 & \textbf{0.0000} & 0.6077 & 1.6059 \\
PRM & Sampling / spline & \textbf{20} & 20 & \textbf{0.0000} & 0.6009 & 1.7984 \\
Elastic band & Gradient / optimization & \textbf{20} & 20 & \textbf{0.0000} & 0.5639 & 0.8981 \\
Potential field & Gradient / optimization & 19 & 20 & 0.0074 & 0.6345 & 1.5505 \\
Minimum jerk & Gradient / optimization & 19 & 20 & \underline{0.0024} & 0.6235 & 1.1371 \\
CHOMP-lite & Gradient / optimization & 19 & 20 & 0.0036 & 0.5995 & 1.2535 \\
\bottomrule
\end{tabular}%
}
\end{table}

After collision repair, 11 of 14 algorithms are collision-free on all 20
scenarios; only Potential field, Minimum jerk, and CHOMP-lite retain residual
collisions in one scenario each.  All clearance values are now positive, but
only the two target-aware methods achieve both high visibility ($>0.89$) and
large clearance ($>2.7$\,m), confirming that target-aware planning is essential
for occlusion-dense tracking.

\begin{table}[!htbp]
\centering
\small
\caption{Group-level summary from the extended runnable baselines.
\textbf{Bold} = best; \underline{underline} = second best.}
\label{tab:appendix_h_group_summary}
\resizebox{\linewidth}{!}{%
\begin{tabular}{lrrrrrr}
\toprule
Group & Algorithms & Vis.\ $\uparrow$ & Blocked LOS $\downarrow$ & Coll.\ frac.\ $\downarrow$ & Clearance (m) $\uparrow$ & Path len.\ (m) $\downarrow$ \\
\midrule
Provided target-aware & 2 & \textbf{0.9364} & \textbf{0.0636} & \textbf{0.0000} & \textbf{3.3917} & 106.7 \\
A* family & 4 & \underline{0.6205} & \underline{0.3795} & \textbf{0.0000} & \underline{1.3774} & 86.6 \\
Sampling / spline & 3 & 0.6083 & 0.3917 & \textbf{0.0000} & 1.6370 & \textbf{86.6} \\
Gradient / optimization & 5 & 0.6075 & 0.3925 & 0.0027 & 1.2711 & 88.3 \\
\bottomrule
\end{tabular}%
}
\end{table}

\subsection{Measured Trajectory Reproduction}
\label{subsec:appendix_h_trajectory_reproduction}

The trajectory reproduction figures use the saved measured trajectory JSON files
and the same normalized obstacle geometry.  Gray rectangles or translucent boxes
denote obstacles, the dashed black curve denotes the target trajectory, and
colored curves denote planner outputs.

\begin{figure}[!htbp]
\centering
\includegraphics[width=0.92\linewidth]{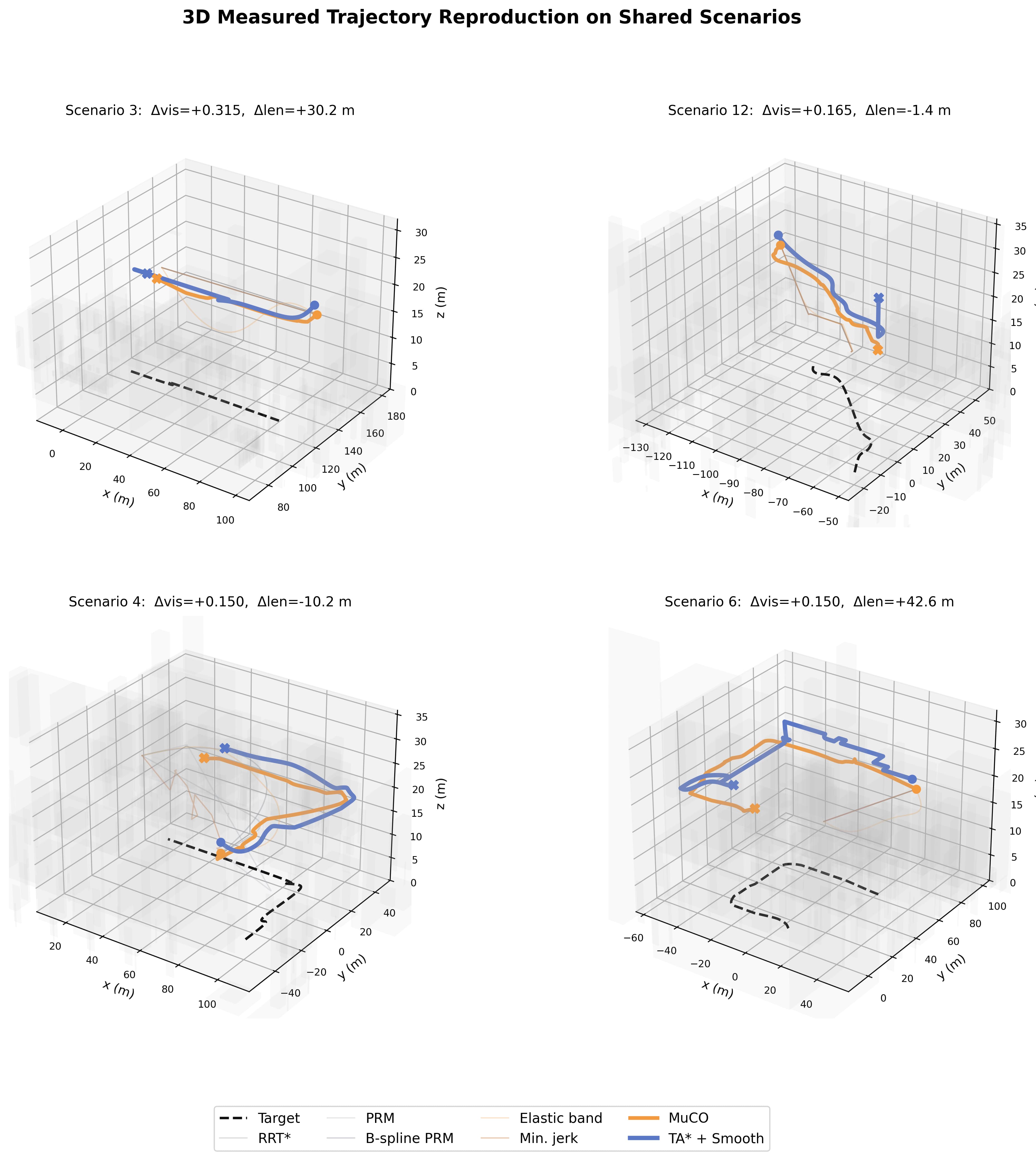}
\caption{3D measured trajectory reproduction on the four scenarios with the
largest TA*+Smooth visibility gain.  Blue = TA*+Smooth, orange = MuCO, dashed
black = target (pedestrian), gray = Python reference planners.  The 3D view
reveals the altitude separation between the drone planners and the
ground-level target.}
\label{fig:appendix_h_measured_trajectory_grid}
\end{figure}

\subsection{Reproduction}
\label{subsec:appendix_h_rerun}

The measured tables in this appendix are produced by a reproduction package
released alongside the dataset.  The package contains the primary four-axis
measurements, the extended runnable-baseline measurements, and the scripts that
regenerate every table and figure from raw per-scenario JSON.  These artefacts
suffice to reproduce every table above without relying on non-runnable external
implementations.  Step-by-step reproduction commands are provided in the release
repository.

\subsection{Planner Algorithmic Details}
\label{app:planner_details}

This subsection gives the full algorithmic specification of the two drone
trajectory planners summarized in \cref{sec:step4}.  Notation matches the main
text; default values are those of the released reference implementations.

\subsubsection{TA*+Smooth (two-stage)}
\label{subsec:appendix_planner_tastar}

\paragraph{Frontend (Track A*).}
The frontend instantiates Track A*~\citep{chen2026trackastar} on a 4D
spatio-temporal voxel grid.  The grid is built per scene with default
resolution $(\Delta_{xy}, \Delta_z) = (4.0\,\mathrm{m}, 4.0\,\mathrm{m})$, a
corridor margin of $45$\,m around the target trajectory, and an altitude
envelope of $[20, 100]$\,m.  At every layer the search is constrained by a
beam of width $2048$ (the smallest beam value observed in our local sweep
that keeps the per-scene visibility regression within $\leq 5\,\mathrm{pp}$;
the corresponding sweep tables are included in the release package).  The
visibility test inside
A* uses five rays per evaluation:
\[
(0,0,0),\ (0,0,+0.8),\ (0,0,-0.6),\ (+0.3,0,0),\ (-0.3,0,0),
\]
matching the offsets of the reference Track A* baseline so the output
visibility metric is comparable.  The A* cost combines a tracking weight
$2.0$, visibility weight $18.0$, path weight $1.0$, safety weight $8.0$, and
smoothness weight $0.15$; per-cell signed distance to obstacles is cached
across time because obstacles are static.

\paragraph{Backend (post-smoothing).}
The smoothed trajectory keeps the per-frame target association produced by the
frontend; it only modifies the spatial positions.  The backend has two
sub-stages.  First, a \emph{shortcut} pass attempts to replace runs of up to
$12$ consecutive waypoints with a straight-line interpolant; a shortcut is
accepted iff every interpolated point still passes the
\texttt{candidate\_is\_valid} check (collision-free with the configured safety
margin, and not below the local visibility anchor described below).  Second,
an \emph{elastic} relaxation runs for up to $30$ iterations: for each interior
waypoint that is not flagged as a visibility anchor, the algorithm moves the
point towards the midpoint of its two neighbours by a step
$\alpha \in [0.02, 0.35]$, with the step shrinking by a factor of $2$ if the
candidate is invalid.  The iteration terminates early when no waypoint was
updated.

\paragraph{Acceptance criteria.}
After both sub-stages, the smoothed trajectory is accepted only if (i) the
mean visibility loss with respect to the raw TA* output is at most
$0.05$, (ii) the per-frame visibility loss never exceeds $0.10$ at any frame
that is not above the anchor threshold $0.999$, and (iii) the minimum obstacle
distance is at least the safety distance (default $3.0$\,m).  If any condition
fails the planner returns the raw TA* output, so TA*+Smooth always produces a
collision-safe trajectory.

\subsubsection{MuCO (one-shot multi-constraint gradient optimizer)}
\label{subsec:appendix_planner_muco}

\paragraph{Optimization variables and loss.}
MuCO treats every interior waypoint $\mathbf{p}_i$ as a free variable and
minimizes $L = \sum_i L_i(\mathbf{p}_i)$ by finite-difference gradient descent
($\varepsilon = 0.5$\,m).  $L_i$ is the weighted sum of:
\begin{itemize}[nosep]
\item \emph{Tracking} ($w_{\mathrm{tr}}=2.0$): $(\lVert\mathbf{p}_i - \mathbf{x}_{t(i)}\rVert - d_{\mathrm{opt}})^2$, with $d_{\mathrm{opt}}=28.0$\,m;
\item \emph{Smoothness} ($w_{\mathrm{sm}}=4.0$): $\lVert\mathbf{p}_{i+1} - 2\mathbf{p}_i + \mathbf{p}_{i-1}\rVert^2$;
\item \emph{Jerk} ($w_{\mathrm{je}}=3.0$): discrete third-difference norm squared;
\item \emph{Safety} ($w_{\mathrm{sa}}=2.0$): $\tfrac{1}{2}(d_{\mathrm{inf}} - d_{\min})^2$ when the minimum obstacle distance drops below the influence radius $d_{\mathrm{inf}}=8.0$\,m;
\item \emph{Visibility} ($w_{\mathrm{vi}}=2.0$): $(1 - V(\mathbf{p}_i, \mathbf{x}_{t(i)}))^2$, with $V$ estimated by a ray-fraction test (the implementation reuses Track A*'s 5-ray bundle smoothed by the $\varepsilon = 0.5$\,m FD window so the resulting gradient is well-behaved);
\item \emph{View angle} ($w_{\mathrm{va}}=1.0$): segmental deviation from a $45^\circ$ pitch target and the smoothed pedestrian heading;
\item \emph{Path length} ($w_{\mathrm{pl}}=2.0$): $0.1\,\lVert\mathbf{p}_i - \mathbf{p}_{i-1}\rVert$.
\end{itemize}
Fixed-coefficient regularisers additionally penalize altitude below
$z_{\min}=20$\,m (coefficient $50$), above the preferred altitude
$z_{\mathrm{pref}}=20$\,m (coefficient $20$), altitude oscillation (factor
$8$), and pitch deviation outside $[30^\circ, 60^\circ]$.

\paragraph{Outer loop and convergence.}
The outer loop runs for at most $1500$ iterations, with learning rate $0.05$
and a per-iteration per-waypoint displacement clip of $0.5$\,m.  Convergence
is declared when $|\Delta L| < 10^{-5}$.

\paragraph{Projection and the relaxed safety floor.}
After each gradient step every waypoint is projected to the feasible region:
$z$ is clamped to $[z_{\min}, z_{\max}]$, per-step displacement is capped by
$v_{\max}\,dt$, and at most $10$ iterations of obstacle push-out are
performed along the local outward normal.  The nominal safety distance is
$3.0$\,m and the relaxed floor $2.5$\,m; the relaxed floor is engaged only
when the projection cannot reach the nominal floor within the budget.  Hard
interpenetration (clearance $< 0$\,m) is always rejected regardless of
relaxation.

\paragraph{Building-circling mitigation.}
We observed that on long line-of-sight blockages MuCO can chase marginal
visibility gains and produce loops around buildings.  The mitigation is
implemented inside the optimizer rather than as ad-hoc post-processing:
runs of consecutive waypoints whose per-frame visibility $V$ stays below
$0.3$ for at least $20$ frames are treated as a \emph{low-visibility run};
inside such a run, the visibility and view-angle gradient updates are masked
to zero while smoothness, safety, tracking, and altitude terms continue to be
optimized, preventing the planner from chasing marginal visibility gains
while keeping the trajectory safe and smooth.

\paragraph{Hyperparameter selection.}
The seven main loss weights and the auxiliary regulariser coefficients reported
above were determined through a small number of manual tuning rounds on
$\sim 20$ pedestrian trajectories from Town10HD\_Opt, using mean visibility and
the smoothness score $S$ (Eq.~\ref{eq:smoothness_score}) as joint diagnostics,
rather than a systematic grid search.  A more rigorous sensitivity study is left
to future work.
}

\FloatBarrier
{\let\appendix\relax
\section{Weather and Time-of-Day Augmentation}
\label{app:weather}

To improve the visual diversity and domain robustness of the generated
dataset, we introduce a \emph{weather and time-of-day (ToD) injection}
module that systematically varies atmospheric conditions and solar
illumination across rendered trajectories.  This appendix details the
preset taxonomy, the parameter space, the selection strategy, and the
resulting metadata schema.

\subsection{Design Rationale}
\label{app:weather:rationale}

Real-world drone and pedestrian navigation must cope with varying
visibility caused by precipitation, fog, haze, and illumination
changes.  Rather than exhaustively sweeping the full CARLA weather
parameter space (13 continuous knobs), we define a compact set of
\textbf{6 minimal knobs} that capture the perceptually dominant axes
of variation while keeping the remaining knobs at CARLA defaults to
prevent implicit parameter drift.  These six fields map one-to-one to
CARLA \texttt{WeatherParameters} attributes:

\begin{enumerate}
  \item \emph{cloudiness} — sky overcast ratio ($0$--$100$).
  \item \emph{precipitation} — rain intensity ($0$--$100$).
  \item \emph{fog density} — volumetric fog concentration ($0$--$100$).
  \item \emph{fog distance} — distance (m) before fog begins attenuating.
  \item \emph{sun altitude angle} — solar elevation (degrees); negative values place the sun below the horizon, producing nighttime-like lighting.
  \item \emph{sun azimuth angle} — solar azimuth (degrees).
\end{enumerate}

\noindent
The first four knobs are governed by \emph{weather presets}, and the
last two by \emph{time-of-day (ToD) presets}.  These two preset pools
are defined and sampled independently—each weather preset specifies
only the four atmospheric knobs, and each ToD preset specifies only
the two solar geometry knobs—enabling combinatorial coverage of
$15 \times 4 = 60$ configurations without per-knob manual tuning.

\subsection{Weather Presets}
\label{app:weather:presets}

We define 15 weather presets organized into four semantic groups
(Table~\ref{tab:weather_presets}).

\begin{table}[!htbp]
\centering
\caption{Weather preset definitions.  Each preset specifies four
atmospheric knobs; remaining CARLA weather parameters stay at engine
defaults.}
\label{tab:weather_presets}
\scriptsize
\resizebox{0.98\textwidth}{!}{%
\begin{tabular}{llrrrr}
\toprule
\textbf{Group} & \textbf{Preset Name} & \textbf{Cloud.} & \textbf{Precip.} & \textbf{Fog Den.} & \textbf{Fog Dist.~(m)} \\
\midrule
\multirow{5}{*}{\textit{Sky}}
  & clear          &  5 &  0 &  0 & 60 \\
  & fair           & 20 &  0 &  0 & 60 \\
  & partly cloudy  & 40 &  0 &  0 & 60 \\
  & cloudy         & 70 &  0 &  0 & 60 \\
  & overcast       & 95 &  0 &  0 & 60 \\
\midrule
\multirow{4}{*}{\textit{Rain}}
  & drizzle        & 50 & 15 & 10 & 50 \\
  & light rain     & 60 & 30 &  5 & 50 \\
  & medium rain    & 80 & 60 & 15 & 30 \\
  & heavy rain     & 95 & 90 & 25 & 20 \\
\midrule
\multirow{3}{*}{\textit{Fog}}
  & thin fog       & 30 &  0 & 30 & 30 \\
  & mist           & 50 &  0 & 50 & 20 \\
  & dense fog      & 70 &  0 & 80 & 10 \\
\midrule
\multirow{3}{*}{\textit{Haze}}
  & smog           & 60 &  0 & 60 & 15 \\
  & dust haze      & 70 &  0 & 70 & 12 \\
  & snow haze      & 90 &  0 & 40 & 20 \\
\bottomrule
\end{tabular}
}
\end{table}

\noindent
These are \emph{visual/atmospheric} preset names describing the
intended perceptual effect; they do not activate full physical weather
simulation.  For instance, \emph{snow haze} approximates snow-fog
conditions via cloudiness and fog density without enabling
precipitation deposits or a dust-storm effect.  CARLA weather
parameters not listed in Table~\ref{tab:weather_presets} (e.g.,
wetness, wind intensity, scattering) remain at engine defaults.

\subsection{Time-of-Day Presets}
\label{app:weather:tod}

Four discrete time-of-day presets control solar geometry
(Table~\ref{tab:tod_presets}).  When the sun altitude angle is
negative, the sun is placed below the horizon, resulting in sub-horizon
illumination where the scene relies primarily on streetlights and
vehicle headlights for nighttime-like lighting conditions.

\begin{table}[!htbp]
\centering
\caption{Time-of-day preset definitions.}
\label{tab:tod_presets}
\small
\begin{tabular}{lrrl}
\toprule
\textbf{ToD Name} & \textbf{Altitude~($^\circ$)} & \textbf{Azimuth~($^\circ$)} & \textbf{Description} \\
\midrule
morning & $+15$ &  $90$ & Low-angle eastern sunlight \\
noon    & $+75$ & $180$ & Near-overhead sun, minimal shadows \\
dusk    & $  0$ & $270$ & Horizon sun, long shadows, warm tones \\
night   & $-30$ & $  0$ & Sub-horizon; streetlights \& headlights only \\
\bottomrule
\end{tabular}
\end{table}

\noindent
The combination of 15 weather presets $\times$ 4 ToD presets yields
\textbf{60 unique (weather, ToD) configurations}, providing broad
visual coverage without manual curation.

\subsection{Selection Modes}
\label{app:weather:modes}

Three injection modes govern how a (weather, ToD) pair is assigned to
each trajectory (Table~\ref{tab:weather_modes}):

\begin{table}[!htbp]
\centering
\caption{Weather injection modes.  In random-per-path mode,
the deterministic seed derivation ensures reproducibility.}
\label{tab:weather_modes}
\small
\begin{minipage}{0.94\linewidth}
\begin{enumerate}[leftmargin=*,itemsep=1pt,topsep=2pt]
\item \textbf{Off:} no weather modification; CARLA built-in defaults apply, and the weather record is omitted from output metadata.
\item \textbf{Fixed:} user-specified weather and ToD preset names are looked up in the preset pools and applied to every trajectory.
\item \textbf{Random-per-path:} a per-trajectory PRNG is seeded from the weather seed and trajectory index (Eq.~\ref{eq:weather_seed}); one weather and one ToD preset are sampled uniformly. A negative seed falls back to true randomness for exploratory renders only.
\end{enumerate}
\end{minipage}
\end{table}

\paragraph{Deterministic seed derivation.}
For random-per-path mode with a non-negative seed, the PRNG
state is initialized as:

\begin{equation}
  s = s_{\text{weather}} \times 1{,}000{,}003
    + i_{\text{path}} \times 7{,}919 + 11
\label{eq:weather_seed}
\end{equation}

\noindent
where $s_{\text{weather}}$ is the global weather seed and
$i_{\text{path}}$ is the trajectory index.
This linear mixing ensures that (a) the same (seed, index) pair
always produces the identical (weather, ToD) draw, enabling exact
reproduction, and (b) different trajectories within the same batch
receive distinct draws with high probability.

\subsection{Integration with Rendering Pipeline}
\label{app:weather:pipeline}

Figure~\ref{fig:weather_pipeline} illustrates the four-stage data
flow.  At the start of each trajectory replay, the renderer:

\begin{enumerate}
  \item \textbf{Load preset definitions} — the weather and ToD
        preset tables are read once per batch.
  \item \textbf{Resolve mode} — the active injection mode determines
        the (weather, ToD) pair: \emph{off} skips weather entirely;
        \emph{fixed} looks up the user-specified pair;
        \emph{random-per-path} samples uniformly via the
        deterministic seed (Eq.~\ref{eq:weather_seed}).
  \item \textbf{Apply weather} (if mode $\neq$ off) —
        a CARLA \texttt{WeatherParameters} object is constructed from
        the 6 resolved knobs and applied via the CARLA API.
  \item \textbf{Write metadata} (if mode $\neq$ off) —
        the resolved choice is serialized into per-frame and
        trajectory-level metadata files.  In off mode, the
        weather record is omitted from all outputs.
\end{enumerate}

\noindent
All preset definitions, rendering scripts, and metadata schemas
follow the conventions described above.

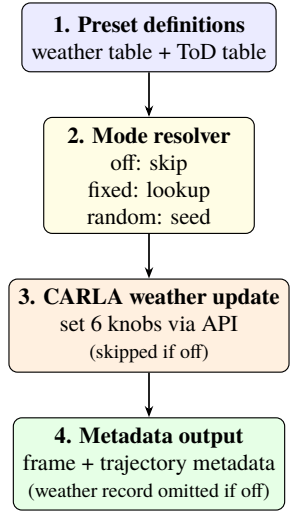
\begin{figure}[!htbp]
\centering
\scalebox{1.07}{%
\begin{tikzpicture}[
  node distance=0.55cm,
  stage/.style={draw, rounded corners=3pt, minimum height=0.85cm,
                minimum width=2.6cm, align=center, font=\footnotesize},
  arr/.style={-{Stealth[length=4pt]}, semithick},
]
\node[stage, fill=blue!8] (load)
  {\textbf{1. Preset definitions}\\weather table + ToD table};
\node[stage, fill=yellow!12, below=of load] (mode)
  {\textbf{2. Mode resolver}\\
   off: skip\\
   fixed: lookup\\
   random: seed};
\node[stage, fill=orange!12, below=of mode] (apply)
  {\textbf{3. CARLA weather update}\\
   set 6 knobs via API\\
   {\scriptsize(skipped if off)}};
\node[stage, fill=green!10, below=of apply] (out)
  {\textbf{4. Metadata output}\\
   frame + trajectory metadata\\
   {\scriptsize(weather record omitted if off)}};

\draw[arr] (load) -- (mode);
\draw[arr] (mode) -- (apply);
\draw[arr] (apply) -- (out);
\end{tikzpicture}}
\caption{Weather injection pipeline: preset definitions are loaded
once per batch; the mode resolver runs per trajectory.
In off mode, stages~3--4 are bypassed.}
\label{fig:weather_pipeline}
\end{figure}

\subsection{Output Metadata Schema}
\label{app:weather:meta}

When weather injection is active (mode $\neq$ off), each frame's
metadata and the trajectory-level record both contain a weather object
with the fields listed in Table~\ref{tab:weather_meta}.  In off mode,
this record is omitted entirely.

\begin{table}[!htbp]
\centering
\caption{Weather metadata fields recorded per frame and per
trajectory.}
\label{tab:weather_meta}
\small
\begin{minipage}{0.94\linewidth}
\begin{description}[leftmargin=2.9cm,style=nextline,itemsep=1pt,topsep=2pt]
\item[\texttt{mode} (str)] ``fixed'' or ``random\_per\_path''.
\item[\texttt{weather\_name} (str)] Matched weather preset name.
\item[\texttt{tod\_name} (str)] Matched ToD preset name.
\item[\texttt{params} (dict)] The six resolved numeric knobs listed in Table~\ref{tab:weather_params_fields}.
\end{description}
\end{minipage}
\end{table}

\begin{table}[!htbp]
\centering
\caption{Fields inside the weather parameter record.}
\label{tab:weather_params_fields}
\small
\begin{tabular}{llr}
\toprule
\textbf{Key} & \textbf{Range} & \textbf{Default} \\
\midrule
\texttt{cloudiness}           & $[0,\,100]$       & 0 \\
\texttt{precipitation}        & $[0,\,100]$       & 0 \\
\texttt{fog\_density}         & $[0,\,100]$       & 0 \\
\texttt{fog\_distance}        & $(0,\,\infty)$~m  & 60 \\
\texttt{sun\_altitude\_angle} & $[-90,\,90]$~deg  & 45 \\
\texttt{sun\_azimuth\_angle}  & $[0,\,360)$~deg   & 0 \\
\bottomrule
\end{tabular}
\end{table}

\subsection{CLI Configuration Reference}
\label{app:weather:cli}

Table~\ref{tab:weather_cli} lists the command-line arguments used to
control weather injection during rendering.

\begin{table}[!htbp]
\centering
\caption{Command-line flags for weather control.}
\label{tab:weather_cli}
\small
\begin{minipage}{0.94\linewidth}
\begin{description}[leftmargin=3.2cm,style=nextline,itemsep=1pt,topsep=2pt]
\item[\texttt{--weather-mode}] Config key \texttt{weather\_mode}; mode selector: off, fixed, or random-per-path (default: off).
\item[\texttt{--weather-name}] Config key \texttt{weather\_name}; weather preset name in fixed mode.
\item[\texttt{--tod-name}] Config key \texttt{tod\_name}; ToD preset name in fixed mode.
\item[\texttt{--weather-pool}] Config key \texttt{weather\_pool}; path to the weather preset definition file.
\item[\texttt{--tod-pool}] Config key \texttt{tod\_pool}; path to the ToD preset definition file.
\item[\texttt{--weather-seed}] Config key \texttt{weather\_seed}; PRNG seed, where a negative value means true random.
\end{description}
\end{minipage}
\end{table}

\subsection{Visual Examples}
\label{app:weather:examples}

Figure~\ref{fig:weather_gallery} presents rendered samples from the
same viewpoint under representative (weather, ToD) combinations,
demonstrating the breadth of visual variation the augmentation
achieves.  Accompanying debug renders (not shown) were used
internally to verify that target pedestrians, vehicles, and projection
bounding boxes remain plausible under each weather condition.

\begin{figure}[!htbp]
\centering
\setlength{\tabcolsep}{2pt}
\begin{tabular}{ccc}
  \includegraphics[width=0.31\textwidth]{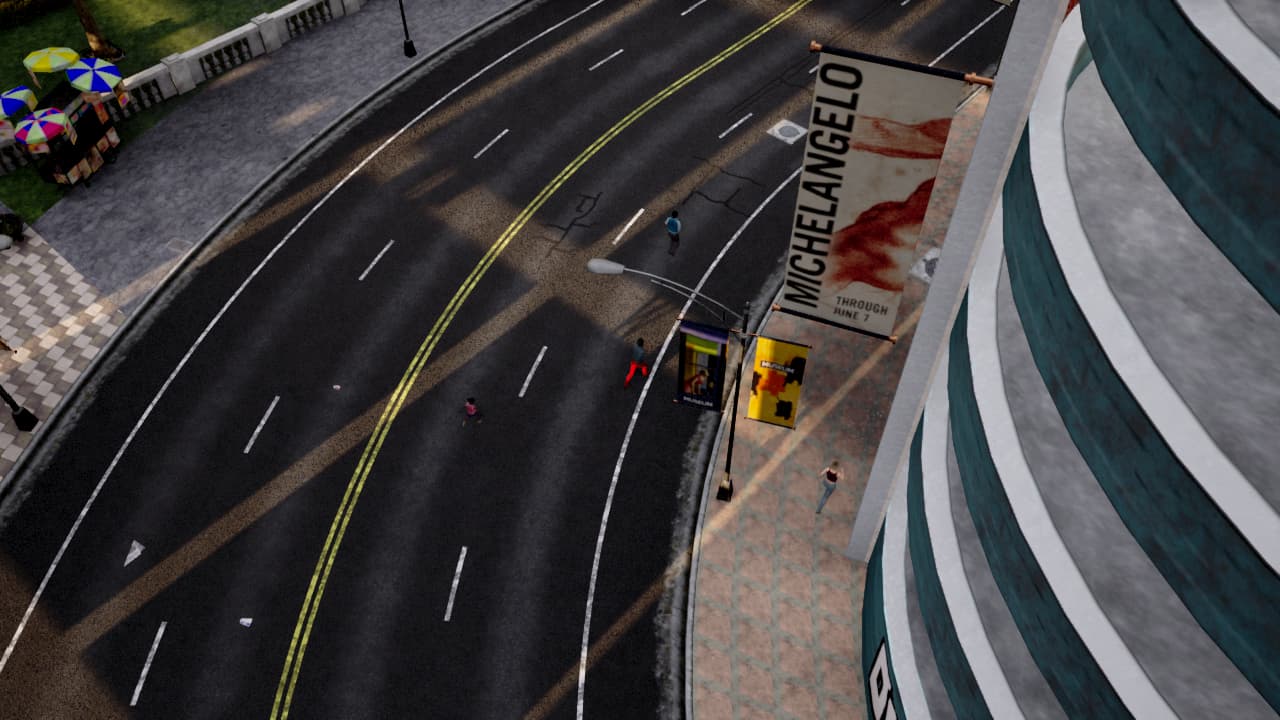} &
  \includegraphics[width=0.31\textwidth]{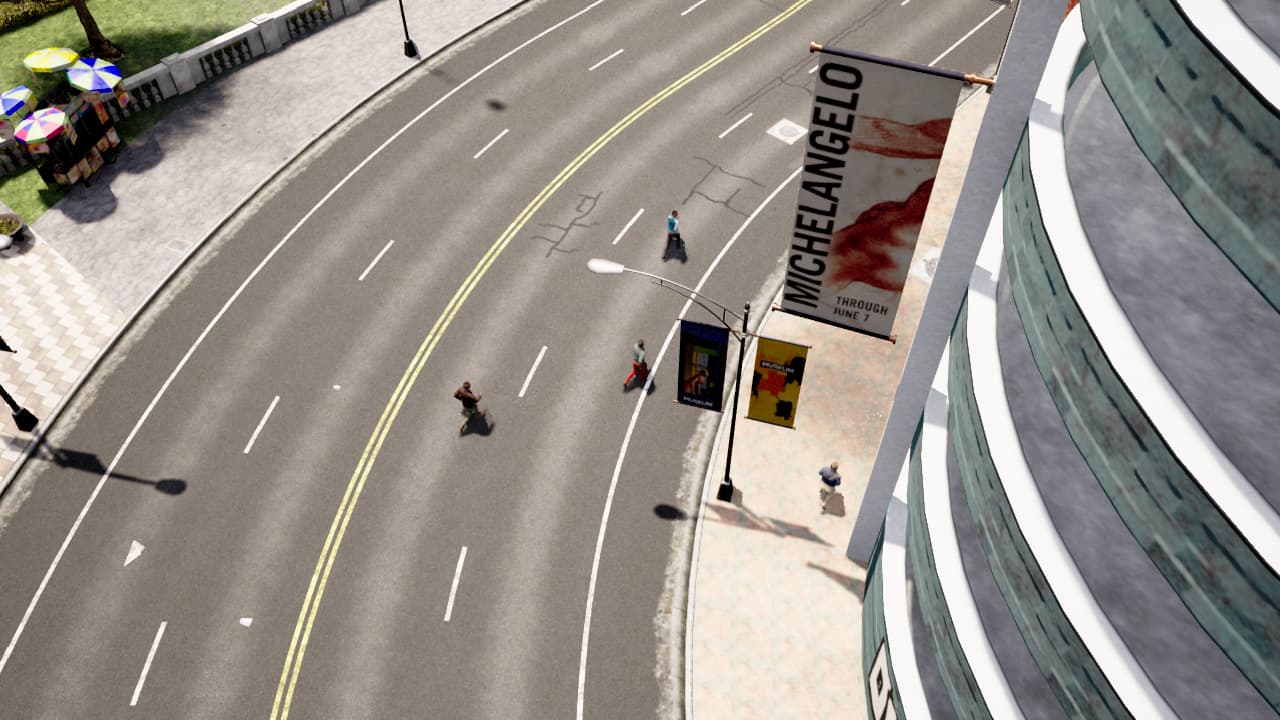} &
  \includegraphics[width=0.31\textwidth]{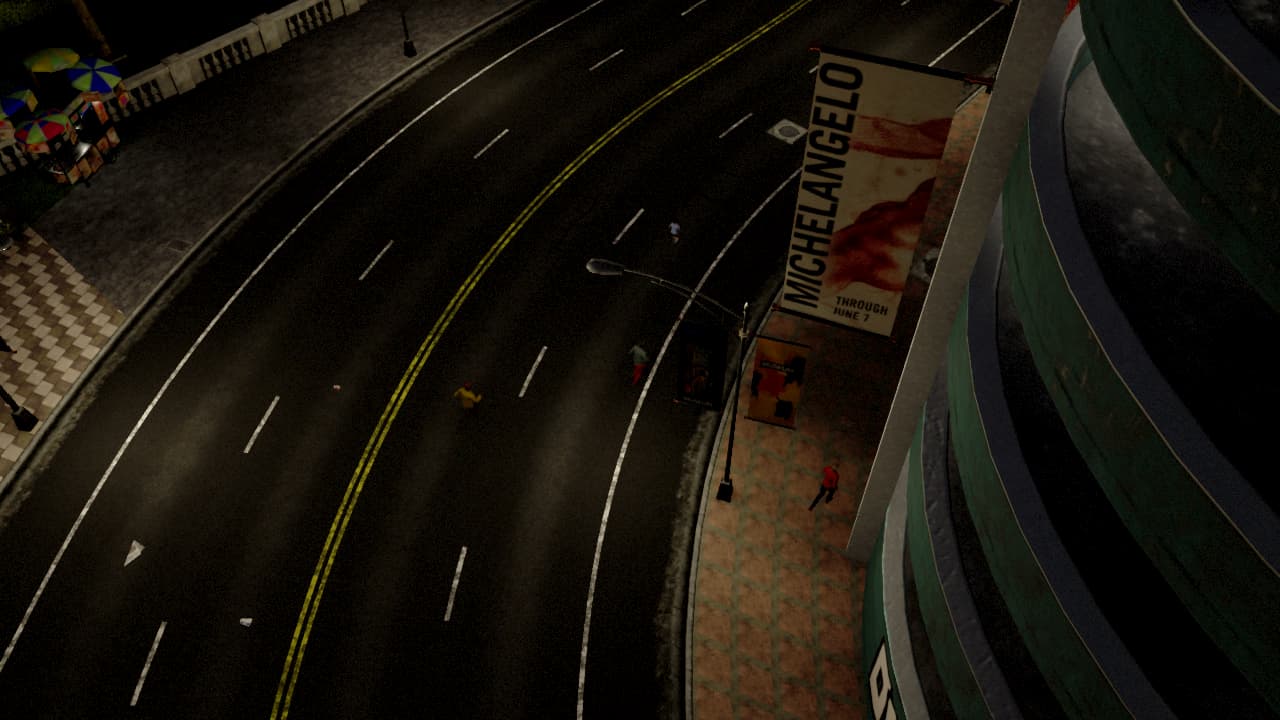} \\[-1pt]
  {\small (clear, morning)} &
  {\small (clear, noon)} &
  {\small (clear, dusk)} \\[4pt]
  \includegraphics[width=0.31\textwidth]{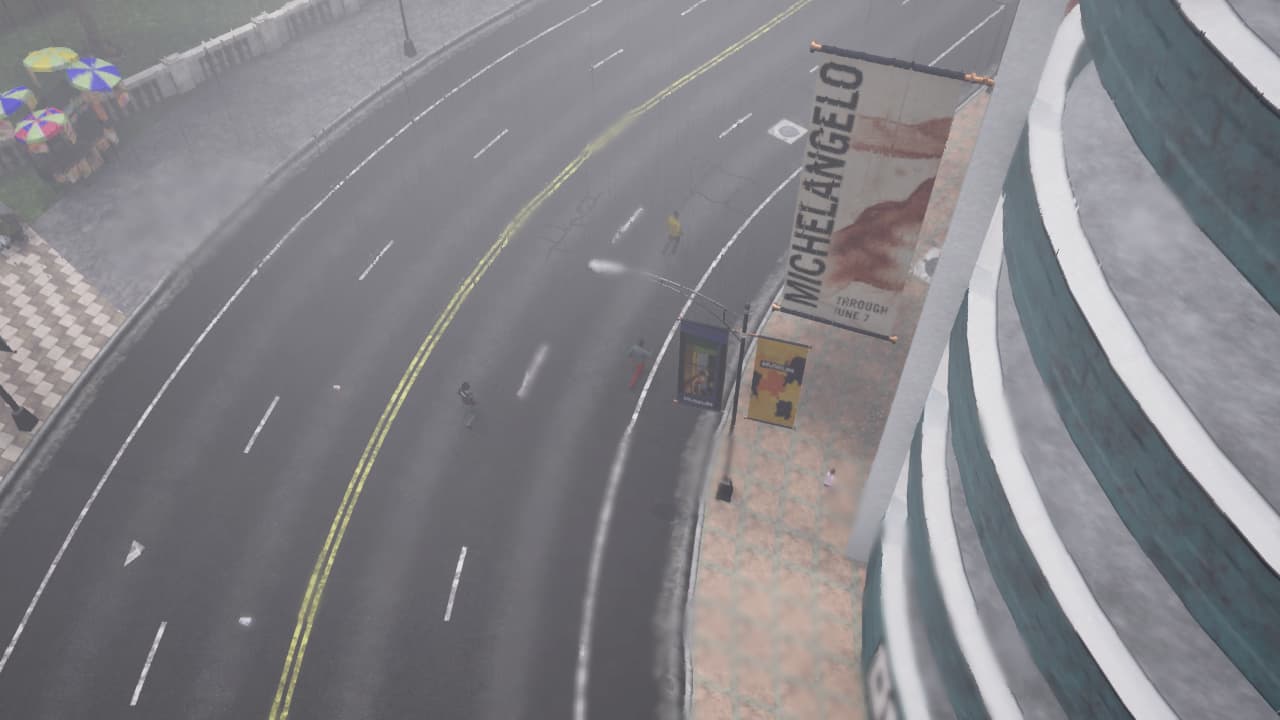} &
  \includegraphics[width=0.31\textwidth]{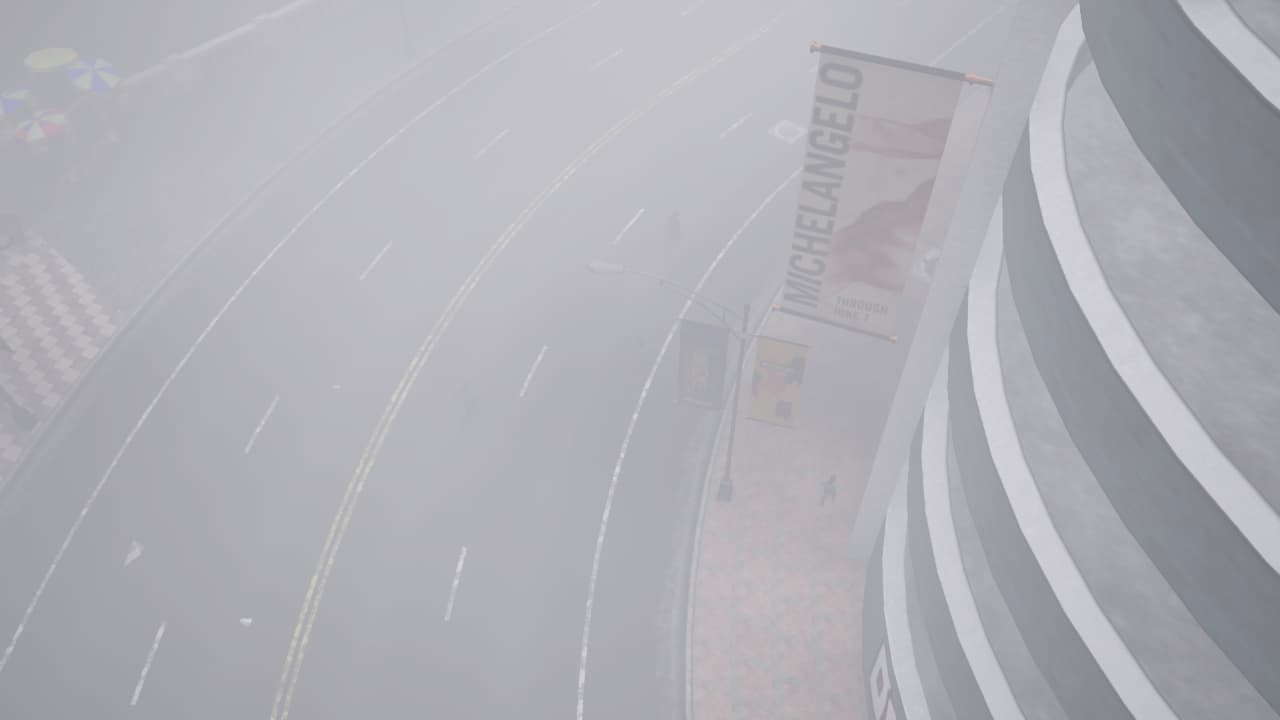} &
  \includegraphics[width=0.31\textwidth]{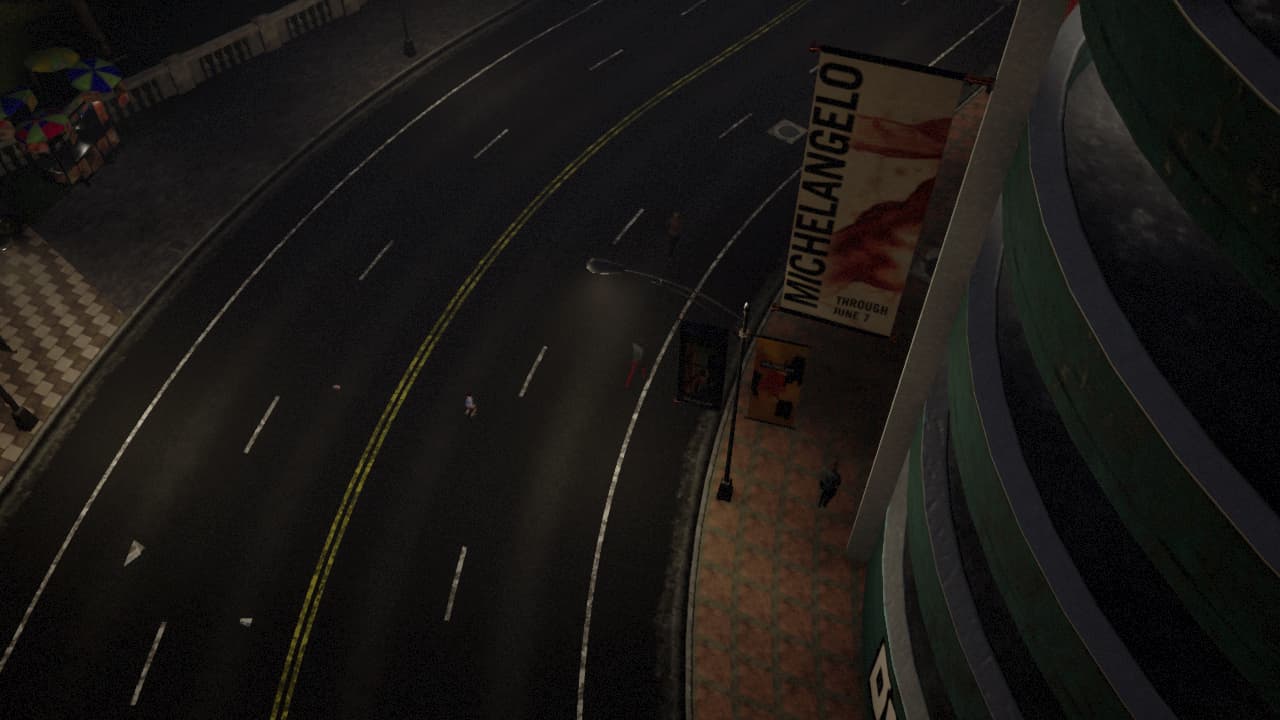} \\[-1pt]
  {\small (heavy rain, noon)} &
  {\small (dense fog, morning)} &
  {\small (smog, dusk)} \\[4pt]
  \includegraphics[width=0.31\textwidth]{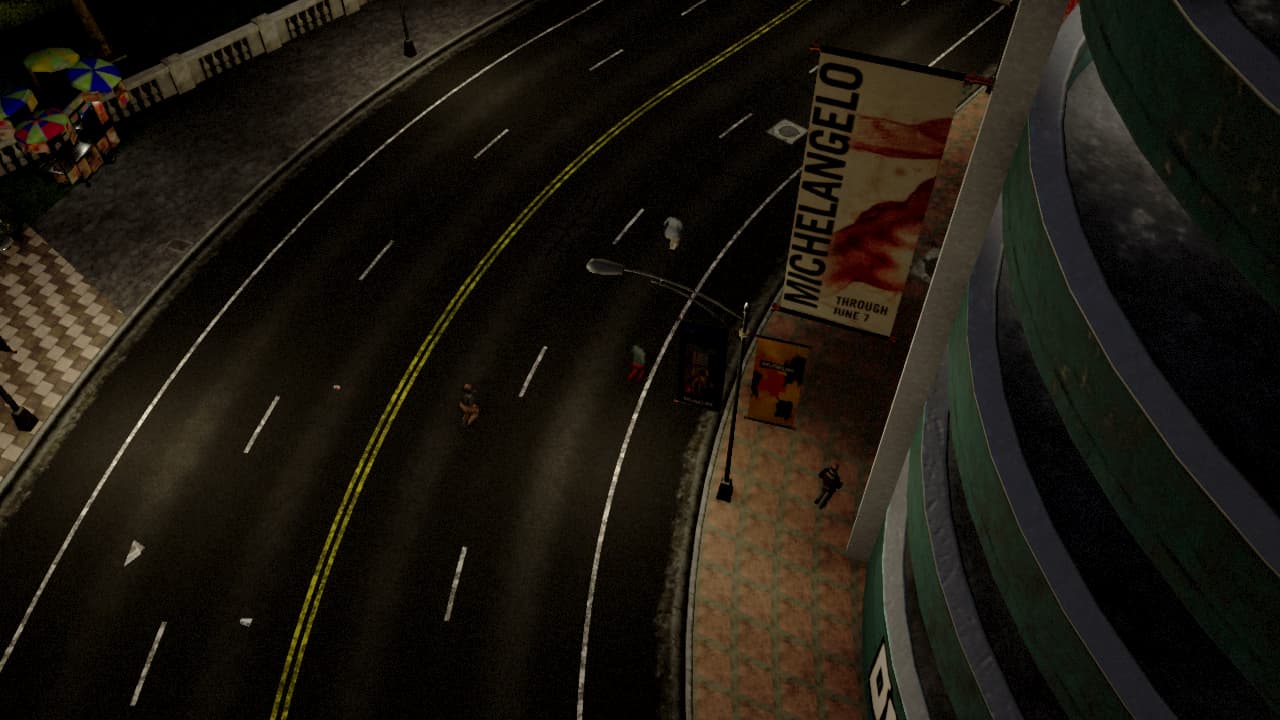} &
  \includegraphics[width=0.31\textwidth]{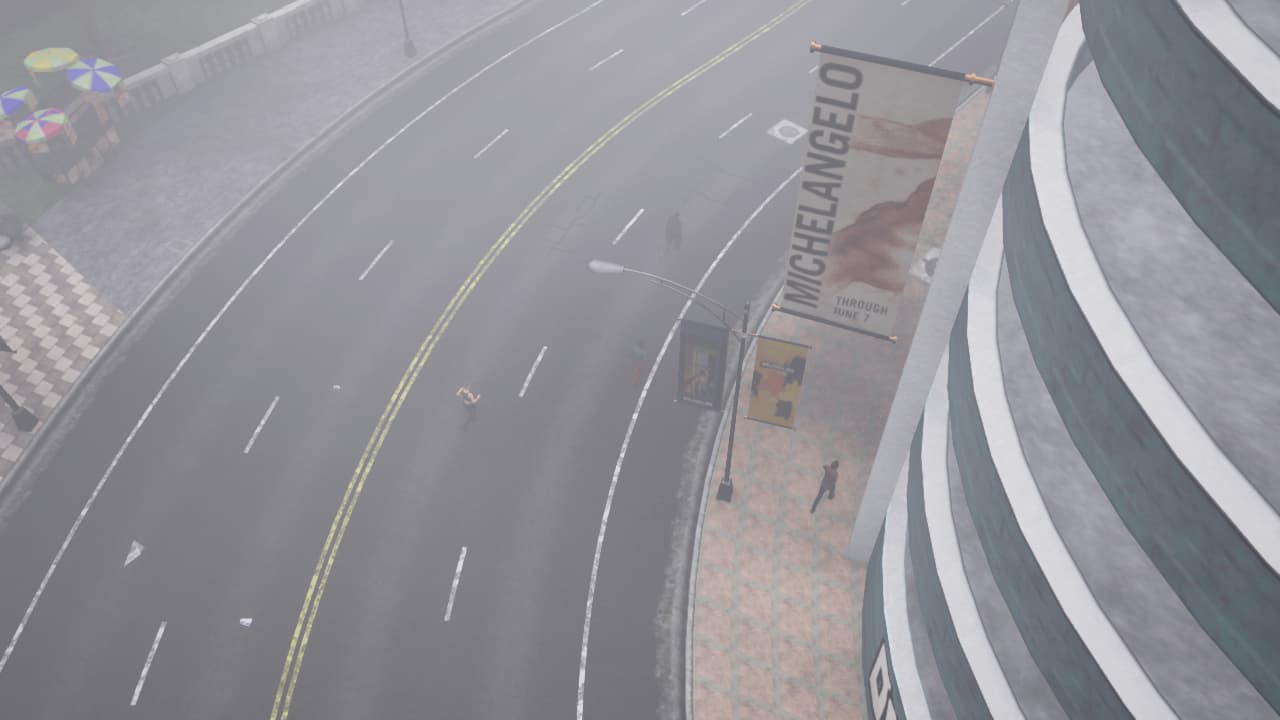} &
  \includegraphics[width=0.31\textwidth]{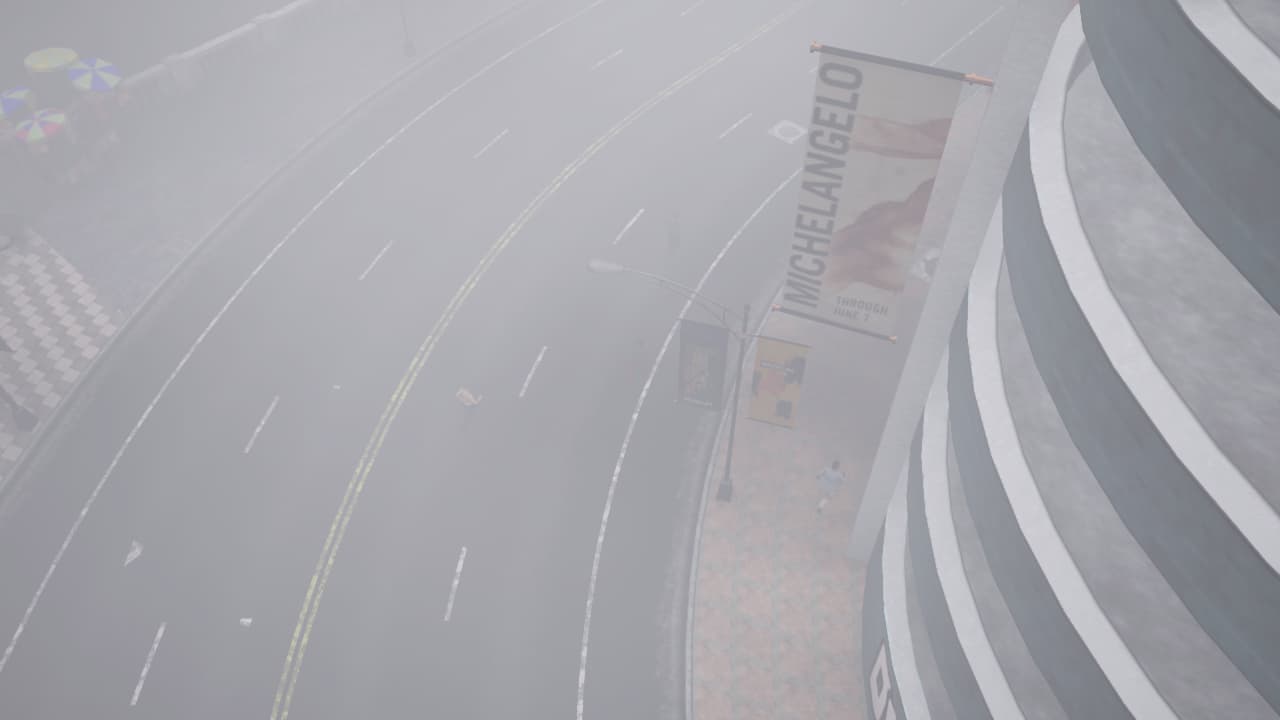} \\[-1pt]
  {\small (overcast, night)} &
  {\small (snow haze, noon)} &
  {\small (dust haze, morning)} \\
\end{tabular}
\caption{Nine representative (weather, ToD) configurations rendered
from the same trajectory (index 0) in fixed mode.
All images are $1280\!\times\!720$ at base FOV $90^\circ$ with
$1.5\times$ optical zoom (effective HFOV $\approx\!67.4^\circ$).
Full parameters are recorded in per-frame and trajectory-level
metadata.
The three high-fog presets (dense fog, dust haze, snow haze) appear
visually similar; their distinction lies in preset semantics rather
than perceptual difference (Table~\ref{tab:weather_presets}).
The (overcast, night) sample uses sun altitude $= -30^\circ$; the
engine's fallback ambient lighting produces a flat low-contrast
appearance rather than a visually dark scene—nighttime is identifiable
primarily through the metadata.
See also Table~\ref{tab:weather_coverage} for the full configuration
matrix.}
\label{fig:weather_gallery}
\end{figure}

\subsection{Statistical Coverage}
\label{app:weather:coverage}

In the default production configuration (random-per-path mode with
seed 20260423), each of the 15 weather presets and 4 ToD presets is
sampled uniformly.  For a batch of $N$ trajectories, the
\emph{expected} number of occurrences for any single preset is:
\[
  \mathbb{E}[\text{count}] = \frac{N}{|\mathcal{P}|}
\]
where $|\mathcal{P}| = 15$ for weather and $|\mathcal{P}| = 4$ for
ToD.  With $N \ge 100$ trajectories, each weather preset is expected
to appear $\approx\!6.7$ times and each ToD preset $\approx\!25$
times, providing non-trivial expected coverage for distribution-level
diversity.  We emphasize that these are \emph{expected values under
uniform sampling}, not guaranteed counts; actual coverage in any
single run depends on the specific seed.  Users can verify the
realized distribution by aggregating the weather-name and ToD-name
fields from the output trajectory metadata, and can reproduce an
identical distribution by reusing the same seed.
Table~\ref{tab:weather_coverage} shows the $15 \times 4$
\emph{configuration space}; note that this represents the set of all
possible combinations, not the realized frequency of any particular
production run.

\newcommand{\wGcell}{\cellcolor{red!20}\scriptsize\textbf{G}}
\newcommand{\wEcell}{\scriptsize$\cdot$}

\begin{table}[!htbp]
\centering
\caption{$15 \times 4$ weather--ToD configuration space.
\textbf{G}~= gallery sample (Figure~\ref{fig:weather_gallery});
$\cdot$~= available but not visualized.}
\label{tab:weather_coverage}
\small
\setlength{\tabcolsep}{6pt}
\begin{tabular}{cl|cccc}
\toprule
\textbf{Group} & \textbf{Preset} & \textbf{morning} & \textbf{noon} & \textbf{dusk} & \textbf{night} \\
\midrule
\multirow{5}{*}{\rotatebox{90}{\textit{Sky}}}
  & clear          & \wGcell & \wGcell & \wGcell & \wEcell \\
  & fair           & \wEcell & \wEcell & \wEcell & \wEcell \\
  & partly cloudy  & \wEcell & \wEcell & \wEcell & \wEcell \\
  & cloudy         & \wEcell & \wEcell & \wEcell & \wEcell \\
  & overcast       & \wEcell & \wEcell & \wEcell & \wGcell \\
\midrule
\multirow{4}{*}{\rotatebox{90}{\textit{Rain}}}
  & drizzle        & \wEcell & \wEcell & \wEcell & \wEcell \\
  & light rain     & \wEcell & \wEcell & \wEcell & \wEcell \\
  & medium rain    & \wEcell & \wEcell & \wEcell & \wEcell \\
  & heavy rain     & \wEcell & \wGcell & \wEcell & \wEcell \\
\midrule
\multirow{3}{*}{\rotatebox{90}{\textit{Fog}}}
  & thin fog       & \wEcell & \wEcell & \wEcell & \wEcell \\
  & mist           & \wEcell & \wEcell & \wEcell & \wEcell \\
  & dense fog      & \wGcell & \wEcell & \wEcell & \wEcell \\
\midrule
\multirow{3}{*}{\rotatebox{90}{\textit{Haze}}}
  & smog           & \wEcell & \wEcell & \wGcell & \wEcell \\
  & dust haze      & \wGcell & \wEcell & \wEcell & \wEcell \\
  & snow haze      & \wEcell & \wGcell & \wEcell & \wEcell \\
\bottomrule
\end{tabular}
\end{table}

\newcommand{\wHcell}[1]{%
  \ifdim #1pt > 74pt \cellcolor{red!50}\else
  \ifdim #1pt > 49pt \cellcolor{orange!40}\else
  \ifdim #1pt > 24pt \cellcolor{yellow!35}\else
  \ifdim #1pt > 0pt  \cellcolor{green!15}\else
                      \cellcolor{white}\fi\fi\fi\fi #1}

\begin{table}[H]
\centering
\caption{Preset parameter summary with shading.  Cell shading is
applied to the three percentage knobs ($0$--$100$; darker = higher);
the fog-distance column shows raw meter values without shading
($10$--$60\,$m), where \emph{smaller} distances produce stronger
near-field fog attenuation.}
\label{tab:weather_heatmap}
\footnotesize
\setlength{\tabcolsep}{3pt}
\begin{tabular}{l|rrrr}
\toprule
\textbf{Preset} & \textbf{Cloud.} & \textbf{Precip.} & \textbf{Fog D.} & \textbf{Fog Dist.~(m)} \\
\midrule
clear          & \wHcell{5}  & \wHcell{0}  & \wHcell{0}  & 60 \\
fair           & \wHcell{20} & \wHcell{0}  & \wHcell{0}  & 60 \\
partly cloudy  & \wHcell{40} & \wHcell{0}  & \wHcell{0}  & 60 \\
cloudy         & \wHcell{70} & \wHcell{0}  & \wHcell{0}  & 60 \\
overcast       & \wHcell{95} & \wHcell{0}  & \wHcell{0}  & 60 \\
\midrule
drizzle        & \wHcell{50} & \wHcell{15} & \wHcell{10} & 50 \\
light rain     & \wHcell{60} & \wHcell{30} & \wHcell{5}  & 50 \\
medium rain    & \wHcell{80} & \wHcell{60} & \wHcell{15} & 30 \\
heavy rain     & \wHcell{95} & \wHcell{90} & \wHcell{25} & 20 \\
\midrule
thin fog       & \wHcell{30} & \wHcell{0}  & \wHcell{30} & 30 \\
mist           & \wHcell{50} & \wHcell{0}  & \wHcell{50} & 20 \\
dense fog      & \wHcell{70} & \wHcell{0}  & \wHcell{80} & 10 \\
\midrule
smog           & \wHcell{60} & \wHcell{0}  & \wHcell{60} & 15 \\
dust haze      & \wHcell{70} & \wHcell{0}  & \wHcell{70} & 12 \\
snow haze      & \wHcell{90} & \wHcell{0}  & \wHcell{40} & 20 \\
\bottomrule
\end{tabular}
\end{table}
}

\FloatBarrier
{\let\appendix\relax
\providecommand{\IfRefDefined}[3]{%
  \expandafter\ifx\csname r@#1\endcsname\relax #3\else #2\fi}
\appendix
\section{SimWorld Infinite Generation Pipeline}
\label{app:simworld}

To scale beyond the fixed map inventory shipped with CARLA, we
integrate the SimWorld procedural scene engine (Unreal Engine~5)
with our existing CARLA-side trajectory sampler.  We refer to the
resulting end-to-end loop as \textbf{Random Procedural Scene
Synthesis} (RPSS): a text prompt is converted into a UE world, the
world is exported as a structured 3D box map, that map drives
pedestrian and drone trajectory planning, and the optimized
trajectories are replayed inside SimWorld to harvest RGB, depth, and
metadata frames.  Because both the scene and the trajectories are
generated on demand, the supply of training data is, in principle,
unbounded.

This appendix documents the six pipeline stages, their input/output
contracts, the migration notes that arose from porting our CARLA
(UE\,4) scripts to SimWorld (UE\,5), and the verification artifacts
produced by a representative production run.

\subsection{Pipeline Overview}
\label{app:simworld:overview}

Figure~\ref{fig:simworld_pipeline} summarises the data flow.  Each
stage owns one CLI entry point and one canonical JSON product; all
downstream stages consume only those JSON products, which keeps the
pipeline auditable and trivially resumable from any checkpoint.
For navigation, \S\,\ref{app:simworld:step1}--\ref{app:simworld:step6}
below correspond one-to-one with Steps~1--6 in the diagram.

\medskip\noindent\textbf{Notation convention.}
Throughout this appendix we identify each stage by the role of its
inputs and outputs rather than by the concrete file paths or
scripts.  The canonical JSON product of each stage is shown in the
green box of Figure~\ref{fig:simworld_pipeline}; the corresponding
CLI flags are tabulated in
Table~\ref{tab:simworld_step6_cli}, and we do not repeat them in
the prose.

\begin{figure}[!htbp]
\centering
\begin{tikzpicture}[
  node distance=0.45cm and 0.5cm,
  box/.style={draw, rounded corners=3pt, minimum height=0.95cm,
              minimum width=3.05cm, align=center, font=\small},
  stage/.style={box, fill=orange!12},
  file/.style={box, fill=blue!8, minimum height=0.65cm,
               minimum width=2.6cm, font=\scriptsize},
  outfile/.style={box, fill=green!10, minimum height=0.65cm,
                  minimum width=2.6cm, font=\scriptsize},
  arr/.style={-{Stealth[length=5pt]}, thick},
  mapdep/.style={-{Stealth[length=5pt]}, thick, dashed, gray!70},
  livedep/.style={-{Stealth[length=5pt]}, thick, dashed, orange!70!red},
]
\node[stage] (s1) {Step~1\\Text-to-World (Gemini)};
\node[stage, below=of s1] (s2) {Step~2\\3D Box Export (UnrealCV)};
\node[stage, below=of s2] (s3) {Step~3\\Box Simplification};
\node[stage, below=of s3] (s4) {Step~4\\Sampling + A* Path};
\node[stage, below=of s4] (s5) {Step~5\\OneShot Trajectory Opt.};
\node[stage, below=of s5] (s6) {Step~6\\Replay \& Frame Capture};

\node[outfile, right=0.8cm of s1] (p1) {\texttt{combined\_world.json}};
\node[outfile, right=0.8cm of s2] (p2) {\texttt{boxes\_3d.json}};
\node[outfile, right=0.8cm of s3] (p3) {\texttt{step2\_simplified.json}};
\node[outfile, right=0.8cm of s4] (p4) {\texttt{path.json}};
\node[outfile, right=0.8cm of s5] (p5) {\texttt{drone\_trace.json}};
\node[outfile, right=0.8cm of s6] (p6) {\texttt{frames\_playback/}\\\texttt{rgb,depth,meta}};

\node[file, left=0.8cm of s1] (in1) {prompt\\\textit{LLM API key}};
\node[file, left=0.8cm of s4] (in4) {\texttt{roi\_polygon.json}\\(manual ROI)};

\draw[arr] (s1) -- (p1);
\draw[arr] (s2) -- (p2);
\draw[arr] (s3) -- (p3);
\draw[arr] (s4) -- (p4);
\draw[arr] (s5) -- (p5);
\draw[arr] (s6) -- (p6);

\draw[arr] (in1) -- (s1);
\draw[arr] (in4) -- (s4);

\draw[arr] (s1) -- (s2);
\draw[arr] (s2) -- (s3);
\draw[arr] (s3) -- (s4);
\draw[arr] (s4) -- (s5);
\draw[arr] (s5) -- (s6);

\draw[mapdep] (p3.east) .. controls +(1.8,0) and +(1.8,0) ..
  node[right=10pt, font=\scriptsize, gray]{\textit{simplified map}}
  (s5.east);
\draw[livedep] (s1.west) .. controls +(-2.0,0) and +(-2.4,1.2) ..
  node[left=4pt, font=\scriptsize, orange!70!red]{\textit{live UE world}}
  (s6.west);
\end{tikzpicture}
\caption{RPSS pipeline.  Solid arrows mark the linear data
dependency.  The two dashed arrows encode distinct cross-stage
couplings: the gray arrow (\textit{simplified map}) carries the
Step~3 map back into Step~5, while the orange arrow
(\textit{live UE world}) marks that Step~6 shares the same live
UE process as Step~1.  Every stage exposes exactly one canonical
JSON artifact (green), which keeps checkpointing straightforward.}
\label{fig:simworld_pipeline}
\end{figure}
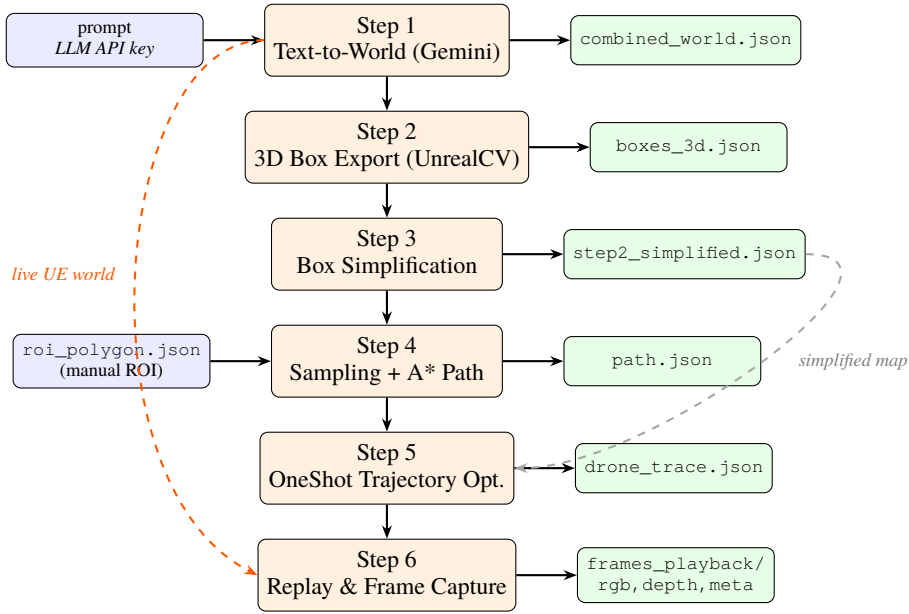

\subsection{Step~1 — Text-to-World via Gemini}
\label{app:simworld:step1}

The first stage converts a free-form natural-language prompt into a
loadable UE world.  A Gemini-backed scene planner emits three
intermediate artefacts: a structured high-level plan, a
procedurally generated city layout, and a text-driven incremental
placement layer.  The latter two are then merged into the
canonical world description (Figure~\ref{fig:simworld_pipeline},
green box at Step~1), which a separate loader hands to a running
UE service to materialize the scene.

\begin{figure}[!htbp]
\centering
\begin{tabular}{cc}
  \includegraphics[width=0.46\textwidth]{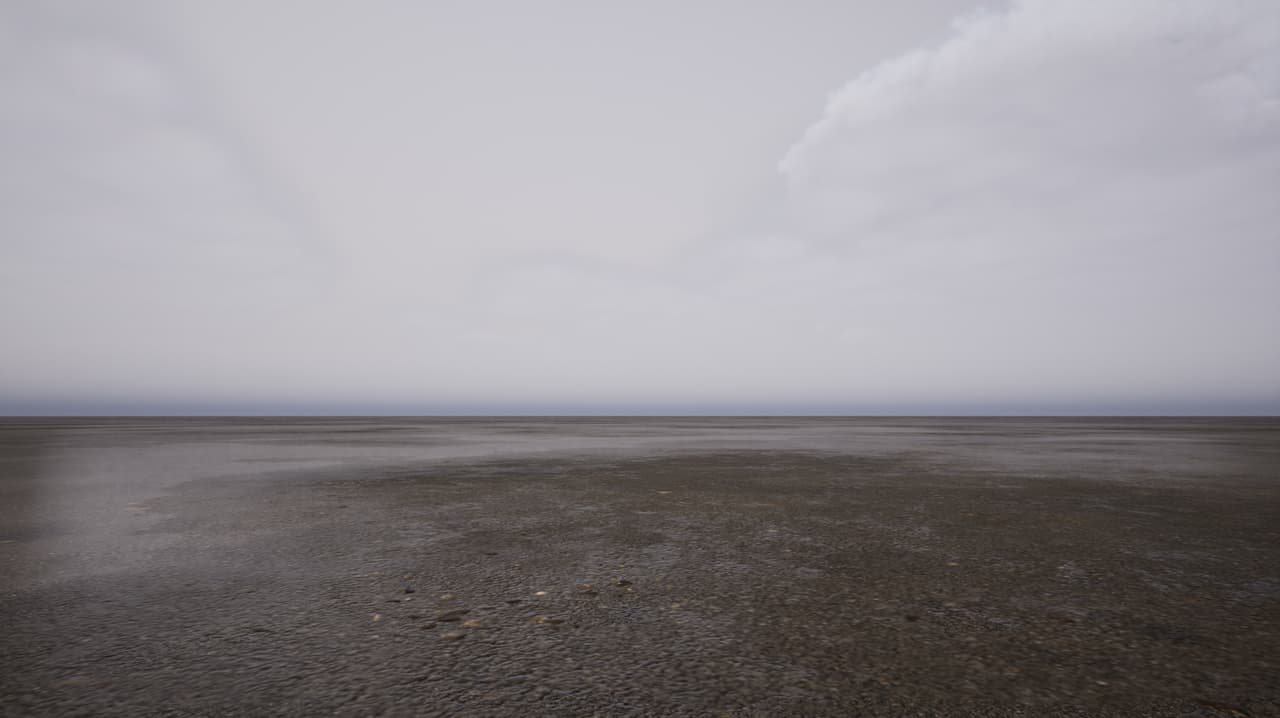} &
  \includegraphics[width=0.46\textwidth]{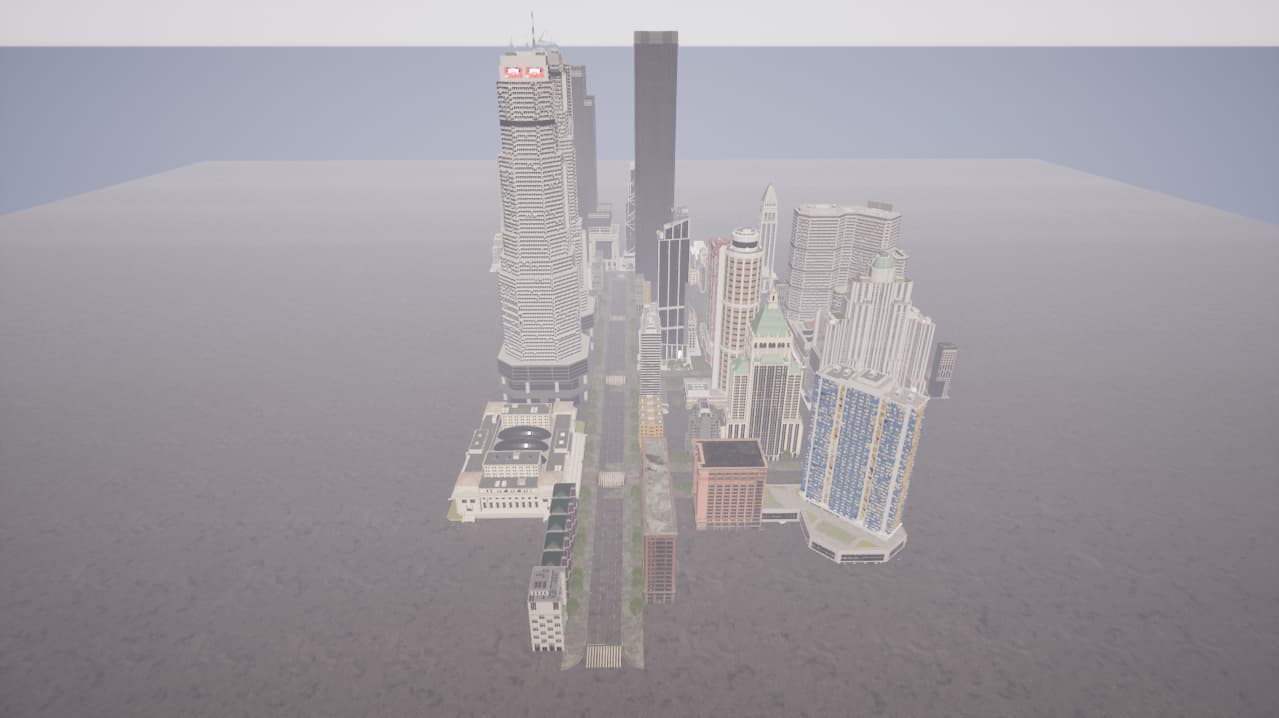} \\[-1pt]
  {\small (a) Empty UE world before loading} &
  {\small (b) Scene materialized from prompt} \\
\end{tabular}
\caption{Step~1 visual outcome.  (a)~The UE service exposes only the
default ground and sky.  (b)~After loading the canonical world JSON
generated from the prompt
\emph{``a medium-scale city with modern high-rises, street-side
vegetation, and varied building heights''}, an explorable mid-density
downtown grid is instantiated in a single call.}
\label{fig:simworld_step1}
\end{figure}

\begin{samepage}
\begin{description}\setlength{\itemsep}{2pt}
  \item[Inputs.] A natural-language prompt and the LLM API
        credentials.
  \item[Outputs.] A timestamped run directory holding the LLM raw
        trace, the three intermediate artefacts, and the canonical
        merged world description recommended for downstream
        loading.
  \item[Reproducibility.] The only state that leaves this stage is
        the canonical world JSON on disk; restarting the UE
        service and re-loading the same file yields a
        deterministic scene without re-querying the LLM
        \emph{provided the SimWorld build and the underlying asset
        catalogue are unchanged}.
\end{description}
\end{samepage}

\subsection{Step~2 — Offline 3D Box Export}
\label{app:simworld:step2}

With the UE world live, the second stage harvests a standardized
3D bounding-box description of every static actor through the
UnrealCV bridge.  The UnrealCV box exporter emits the canonical
box map (Figure~\ref{fig:simworld_pipeline}, Step~2 product), a
metadata sidecar, and a self-contained HTML viewer for interactive
inspection.

\begin{figure}[!htbp]
\centering
\includegraphics[width=0.56\textwidth]{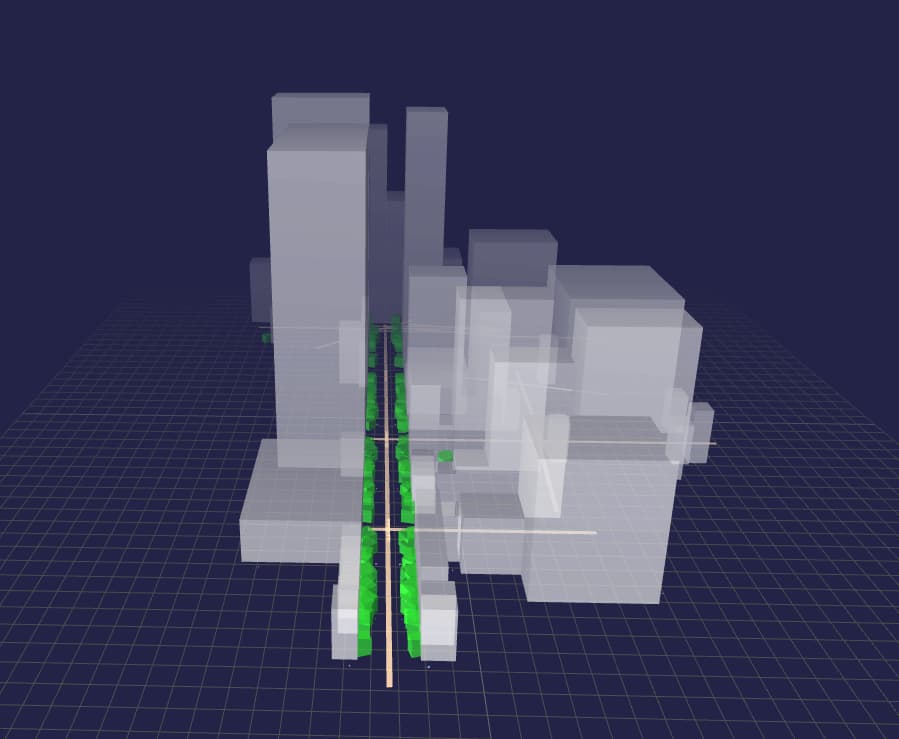}
\caption{Step~2 interactive viewer rendering of the canonical box
map.  Buildings are translucent gray prisms, vegetation are small
green markers, and the road network is the yellow strip down the
centre spine.  The run reported here exported $1{,}704$ objects,
distributed across the four extent-source classes as
$\texttt{road}{=}86$, $\texttt{bbox}{=}0$,
$\texttt{engine}{=}1{,}617$, and $\texttt{default}{=}1$.}
\label{fig:simworld_step2}
\end{figure}

\paragraph{I/O contract.}
The exporter expects a live UE service reachable on the default
UnrealCV port and emits a CARLA-compatible box JSON, so downstream
tooling does not need to branch on engine origin.  An interactive
HTML viewer is produced alongside it for visual QA before paying
the cost of the simplification stage.

The exporter classifies each box's extent source as one of
\{\texttt{road}, \texttt{bbox}, \texttt{engine}, \texttt{default}\}.
On the reference run the \texttt{bbox} bucket is empty
(see Figure~\ref{fig:simworld_step2}); we still monitor this
histogram across all runs because a sudden growth in the
\texttt{default} bucket indicates that a new asset class is missing
its UnrealCV metadata and would otherwise propagate as a unit-sized
collider into later stages.

\subsection{Step~3 — Box Simplification (merge $\to$ crop $\to$ prune)}
\label{app:simworld:step3}

The raw box dump is too noisy for path planning: large vegetation
clusters are split across many overlapping leaves, building shells
contain nested decorative volumes, and a small fraction of boxes
extends below the ground plane.  The simplification pipeline
applies four configurable passes in sequence, summarised in
Table~\ref{tab:simworld_simplify}.

\begin{table}[!htbp]
\centering
\caption{Step~3 simplification passes.  Default values reflect the
production configuration shipped with the pipeline.}
\label{tab:simworld_simplify}
\small
\begin{minipage}{0.94\linewidth}
\begin{enumerate}[leftmargin=*,itemsep=1pt,topsep=2pt]
\item \textbf{\texttt{merge}:} fuses vegetation/building boxes whose centres lie within an L$_\infty$ neighbourhood. Defaults: vegetation tolerance \texttt{v}{=}2.0\,m and building tolerance \texttt{b}{=}5.0\,m.
\item \textbf{\texttt{crop\_tree}:} caps excessive tree heights using an adaptive split so distant trunks do not occlude planning grids. Defaults: \texttt{adaptive} strategy, \texttt{split\_z}{=}5.0\,m, \texttt{adaptive\_height}{=}5.0\,m, and \texttt{cell\_size}{=}0.5\,m.
\item \textbf{\texttt{crop\_below\_ground}:} discards sub-ground extent that A* would later treat as a spurious obstacle. Default: \texttt{ground\_z}{=}0.0\,m for all box types.
\item \textbf{\texttt{prune\_nested}:} removes boxes geometrically contained inside a larger box of the same class. Defaults: all types, $\epsilon{=}10^{-6}$, leaf size 16.
\end{enumerate}
\end{minipage}
\end{table}

\paragraph{Count budget.}
On the reference run, the input contained $1{,}704$ boxes (Vegetation
$871$, Buildings $247$, other $586$).  After all four passes the
output contained $2{,}559$ boxes: the count \emph{increases} because
adaptive tree cropping fragments tall trees into shorter,
planner-friendly cells, and that gain outweighs the reduction from
nested-box pruning.  The simplified-map JSON
(Figure~\ref{fig:simworld_pipeline}, Step~3 product), together
with a metadata sidecar and a human-readable simplification
report, becomes the canonical map for Steps~4 and~5.

\subsection{Step~4 — ROI Annotation, Sampling, and A* Planning}
\label{app:simworld:step4}

Pedestrian-feasible regions must be filtered out of the otherwise
all-purpose box map.  The sampling-and-planning stage is therefore
split into two sub-stages joined by a lightweight web annotator.

\paragraph{(i) ROI mask.}
The simplified map is first rasterised into a planning grid (cell
pitch $0.5$\,m), optionally suppressing the terrain class so the
walkable ground is not mistakenly counted as obstacle.  Each raw
obstacle cell is then dilated with a circular structuring element
of radius $r_{\text{infl}}{=}2$ cells (\,$\approx 1.0$\,m of
pedestrian clearance), and the dilated mask is dumped to PNG.  On
the reference run the grid was $3{,}784 \times 2{,}357$ cells with
$1{,}888{,}811$ raw obstacle cells, inflated to $1{,}967{,}820$
(\,$+4.2\%$) for safety margin.

\paragraph{(ii) Manual polygon annotation.}
Operators open a lightweight web annotator, import the inflated
mask and its metadata, trace the walkable region with a polygon,
and export it as the ROI polygon JSON consumed by the sampler
(Figure~\ref{fig:simworld_pipeline}, Step~4 input).  This
single-time human step keeps the rest of the loop fully automated.

\paragraph{(iii) Pipeline run.}
With the polygon committed, the full sequence
\textit{grid $\to$ project $\to$ inflate $\to$ connectivity $\to$
sample $\to$ astar $\to$ report} is invoked end-to-end.
Table~\ref{tab:simworld_step4} reports the production run, which
finished in $41.6$\,s and produced $20/20$ valid sample pairs and
$20/20$ planned paths.

\begin{table}[!htbp]
\centering
\caption{Step~4 reference run summary.}
\label{tab:simworld_step4}
\small
\begin{tabular}{lr}
\toprule
\textbf{Metric} & \textbf{Value} \\
\midrule
Grid resolution & $3{,}784 \times 2{,}357$ \\
Total boxes after Step~3 & $2{,}559$ \\
Height-overlapping boxes processed & $1{,}606$ \\
Raw obstacle cells & $1{,}888{,}811$ \\
Inflated obstacle cells & $1{,}967{,}820$ \\
Sampled $(s,g)$ pairs & $20/20$ unique \\
Planned A* paths & $20/20$ \\
End-to-end wall-clock time & $41.6$\,s \\
\bottomrule
\end{tabular}
\end{table}

The canonical deliverable handed to Step~5 is the path JSON
(Figure~\ref{fig:simworld_pipeline}, Step~4 product); a companion
PNG overlays all planned paths on the inflated mask for quick
visual QA.

\subsection{Step~5 — OneShot Drone Trajectory Optimization}
\label{app:simworld:step5}

Pedestrian paths from Step~4 anchor the ground truth, but each path
also needs a companion drone trajectory that smoothly tracks the
pedestrian while respecting the static obstacle field.  The
OneShot solver batches the multi-path problem; on a 16-worker
configuration the reference run produced all $20$ drone traces in
$\approx 12$\,min wall-clock total (median per-scenario solve
$\approx 36$\,s).  An example scenario reports $121$ trajectory
points, $2{,}559$ obstacles, $35{,}696$\,ms planning time, $52$
iterations, and $47.9$\,m total length.

\begin{figure}[!htbp]
\centering
\includegraphics[width=0.70\textwidth]{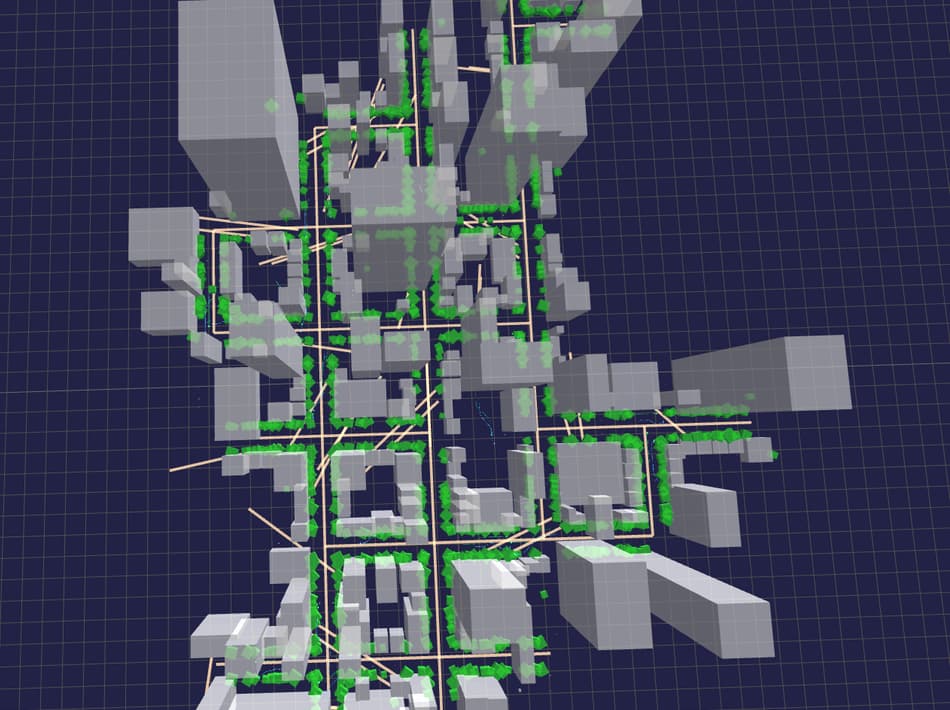}
\caption{Step~5 interactive viewer rendering of all $20$ drone
traces on the simplified box map.  Buildings appear as translucent
gray prisms, vegetation as small green markers, and the planned
pedestrian/drone tracks as the tan polylines threaded along the
street grid.  The viewer's side panel (omitted) exposes per-path
pair/human/drone toggles so any subset of the planned scenarios
can be replayed or isolated for inspection.}
\label{fig:simworld_step5}
\end{figure}

The solver consumes the Step~3 simplified map and the Step~4
pedestrian paths and emits the canonical drone-trace JSON
(Figure~\ref{fig:simworld_pipeline}, Step~5 product), together
with an optional self-contained HTML viewer for qualitative
inspection.

\subsection{Step~6 — SimWorld Replay and Frame Capture}
\label{app:simworld:step6}

The final stage replays each drone trace inside the live SimWorld
process to capture training frames.  Critically, the executor was
ported from the CARLA Python API to the UnrealCV~+~Communicator
stack while keeping both the upstream input contract (the
canonical drone-trace JSON of Step~5) and the downstream per-frame
output contract \emph{unchanged}: the same trajectory-level JSON
schema and the same playback directory tree (one subdirectory per
frame, each holding an RGB frame, a depth frame, and a per-frame
metadata file) that CARLA-side consumers already expect.

\begin{figure}[!htbp]
\centering
\includegraphics[width=0.78\textwidth]{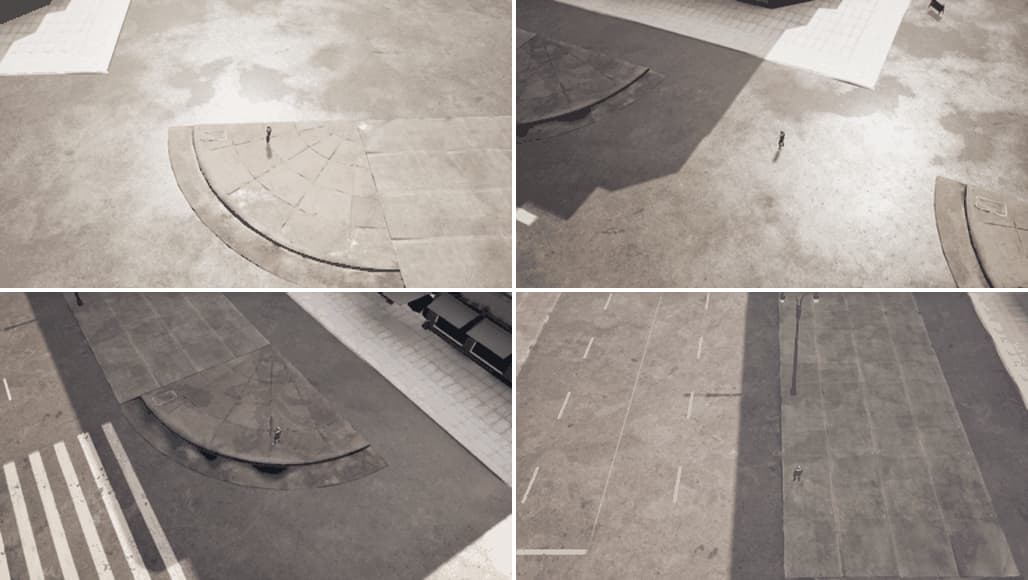}
\caption{Step~6 replay preview (frames sampled at
$t\!\approx\!10\%$, $40\%$, $70\%$, and $100\%$ of the $121$-frame
run, top-left to bottom-right, extracted from the auto-generated
preview GIF).  A
scripted humanoid follows the Step~4 pedestrian path while the
camera tracks the Step~5 drone trace; the four panels visibly
traverse the plaza, an arterial sidewalk, and a zebra-crossing,
demonstrating that the recorded motion actually covers the planned
trajectory.  Per-frame RGB, depth, and metadata are written to disk
at every step so downstream learners see exactly the same file
layout the CARLA-based pipeline produced.}
\label{fig:simworld_step6}
\end{figure}

\paragraph{Contract preservation across the engine swap.}
We treat the per-frame directory layout as the regression contract
between CARLA and SimWorld replays: a byte-level recursive diff
on the same drone trace reports an identical directory tree,
identical JSON schemas at both the trajectory and per-frame level,
and an identical total frame count.  Only the RGB pixel content
and the renderer-dependent metadata fields differ; the
engine-agnostic pose fields match byte-for-byte, which is how
downstream learners remained unchanged across the port.

\paragraph{Reference run and CLI.}
Replaying the first of the $20$ planned scenarios produced
$121/121$ frames into a timestamped output directory, alongside an
auto-generated preview GIF for visual QA.
Table~\ref{tab:simworld_step6_cli} lists the CLI flags that govern
the executor; defaults are tuned so that re-running with only the
configuration file and a path index reproduces the reference
frames.

\begin{table}[!htbp]
\centering
\caption{Step~6 command-line interface.}
\label{tab:simworld_step6_cli}
\small
\begin{minipage}{0.94\linewidth}
\begin{description}[leftmargin=3.4cm,style=nextline,itemsep=1pt,topsep=2pt]
\item[\texttt{--trace}] Drone trace produced by Step~5 (\texttt{drone\_trace.json}).
\item[\texttt{--path-index}] Single index, or \texttt{all} to replay every planned path.
\item[\texttt{--cvip} / \texttt{--cvport}] UnrealCV endpoint; default \texttt{127.0.0.1:9000}, with 9001 sometimes used when 9000 is occupied.
\item[\texttt{--output-root}] Root directory for the run; a timestamped subdirectory is created per invocation.
\item[\texttt{--image-width} / \texttt{--image-height} / \texttt{--fov}] Camera intrinsics for the captured RGB/Depth pair.
\item[\texttt{--require-rgb} / \texttt{--require-depth}] Per-modality export toggles.
\item[\texttt{--settle-delay}] Wait time after each pose change before reading the framebuffer.
\item[\texttt{--coord-scale}] Unit conversion factor; default $100.0$, i.e.\ $1$\,m $\to 100$\,UE\,cm.
\item[\texttt{--coord-z-offset}] Optional uniform $Z$ offset; tune only if the recorded altitude looks systematically off.
\item[\texttt{--gif} / \texttt{--gif-fps} / \texttt{--gif-name}] Post-run GIF compilation for fast visual acceptance.
\end{description}
\end{minipage}
\end{table}

\enlargethispage{5\baselineskip}

\subsection{Migration Notes: CARLA (UE\,4) $\to$ SimWorld (UE\,5)}
\label{app:simworld:migration}

Although the upstream pipeline up to Step~5 is engine-agnostic
(it only consumes JSON), the live execution stages (Step~1, Step~2,
Step~6) crossed an engine boundary.  Three classes of adaptation
proved unavoidable.

\begin{samepage}
\begin{enumerate}
  \item \textbf{Coordinate units.}
        Step~4 produces coordinates with metric semantics (m),
        whereas SimWorld and UnrealCV consume Unreal centimetres.
        Skipping the conversion silently parks the camera at ground
        level so the recorded RGB only shows the humanoid's legs.
        Step~6 therefore exposes \texttt{--coord-scale} (default
        $100.0$, i.e.\ $1$\,m $\to 100$\,UE\,cm) and an optional
        \texttt{--coord-z-offset} microtuning knob.

  \item \textbf{Actor and camera API surface.}
        The CARLA-style actor-spawn, transform, and sensor-attach
        calls were replaced with UnrealCV humanoid commands
        (rotation plus forward step) and the SimWorld camera-pose
        API.  This kept the \emph{interface} of Step~6 identical
        -- it still consumes the canonical Step~5 product and
        writes the same playback directory tree -- which is what
        shielded downstream learners from the engine swap.

  \item \textbf{Box-extent semantics.}
        UE\,5 reports actor extents through a different metadata
        path; Step~2 therefore tracks the
        \texttt{road/bbox/engine/default} histogram so that any
        regression in the engine-side metadata is caught at export
        time, before it can poison the simplified map.
\end{enumerate}
\end{samepage}

\subsection{Discussion}
\label{app:simworld:discussion}

By decoupling \emph{scene supply} (Steps~1--3, driven by SimWorld
and UnrealCV) from \emph{trajectory supply} (Steps~4--5, driven by
the CARLA-derived planner), RPSS turns each text prompt into a fresh
data factory.  Two properties make the loop attractive at scale.

\paragraph{Visual diversity is bounded only by the language model.}
Because Step~1 is parameterised by a free-form prompt, the
diversity ceiling is set by the prompt distribution rather than by a
hand-curated map inventory.

\paragraph{Trajectory diversity is bounded only by the sampler
budget.}
For any fixed scene, Step~4 will draw a fresh batch of
\texttt{(start, goal)} pairs under the same ROI polygon; combined
with the OneShot optimiser of Step~5 and the
\IfRefDefined{app:weather}%
  {weather/ToD injection of Appendix~\ref{app:weather}}%
  {weather/ToD injection module described elsewhere in this paper},
the same world can yield arbitrarily many independent
trajectory$+$lighting$+$weather realisations without any further
human intervention.

\paragraph{Limitations.}
We highlight three practical caveats that future work should
address.
\begin{itemize}[nosep,leftmargin=*]
  \item \textbf{Manual ROI polygon.} The Step~4 ROI polygon is
        currently traced by hand; automating it with a walkability
        segmenter is the most obvious next step.
  \item \textbf{External LLM dependency.} Step~1's reliance on a
        third-party LLM API adds an external dependency and per-call
        cost, which batch users can amortise via prompt-level caching.
  \item \textbf{Operational caveats.} The four-bucket extent
        histogram of Step~2 should move from manual monitoring to CI;
        Step~6 also assumes the same UE service from Steps~1--2 remains alive.
\end{itemize}
}

\end{document}